\documentclass[
10pt, 		
a4paper, 	
oneside, 	
headinclude,footinclude, 
BCOR5mm, 	
]{scrartcl}

\usepackage[
nochapters, 
beramono, 
eulermath,
pdfspacing, 
dottedtoc 
]{classicthesis} 

\usepackage{arsclassica} 

\usepackage[T1]{fontenc} 

\usepackage[utf8]{inputenc} 

\usepackage{graphicx} 

\graphicspath{{Figs/}} 

\usepackage{epsfig}

\usepackage{ifpdf}

\usepackage[labelfont=bf]{subcaption} 

\usepackage{enumitem} 

\usepackage{lipsum} 

\usepackage{amsmath,amssymb,amsthm,bm} 

\usepackage{varioref} 

\usepackage{scalerel} 

\usepackage{dutchcal} 

\usepackage{layout} 

\theoremstyle{definition}

\theoremstyle{plain}

\theoremstyle{remark} 

\hypersetup{
colorlinks=true, breaklinks=true, bookmarks=true,bookmarksnumbered,
urlcolor=webbrown, linkcolor=RoyalBlue, citecolor=webgreen, 
pdftitle={}, 
pdfauthor={\textcopyright}, 
pdfsubject={}, 
pdfkeywords={}, 
pdfcreator={pdfLaTeX}, 
pdfproducer={LaTeX with hyperref and ClassicThesis} 
}

\newcommand{\trans}{^{\ensuremath{\intercal}}}

\newcommand{\vr}[1]{\ensuremath{\mathbf{#1}}}

\newcommand{\gvr}[1]{\ensuremath{\bm{#1}}}

\newcommand{\mt}[1]{\ensuremath{\mathbf{#1}}}

\title{\normalfont\spacedallcaps{The Otbot Project}}

\subtitle{Dynamic modelling, parameter identification and motion control of \\an omnidirectional tire-wheeled robot}

\author{\spacedlowsmallcaps{
		Pere Giró\textsuperscript{1}, 
		Enric Celaya\textsuperscript{1} \& 
		Llu\'{\i}s Ros\textsuperscript{1}}}

\date{\large May 2021} 

\begin{document}


\renewcommand{\sectionmark}[1]{\markright{\spacedlowsmallcaps{#1}}} 

\lehead{\mbox{\llap{\small\thepage\kern1em\color{halfgray} \vline}\color{halfgray}\hspace{0.5em}\rightmark\hfil}} 

\pagestyle{scrheadings} 


\maketitle
\thispagestyle{empty}


	


\let\thefootnote\relax\footnotetext{\textsuperscript{1} \textit{Institut de Rob\`otica i Inform\`atica Industrial (CSIC-UPC), Barcelona. Emails: \texttt{pere.giro5@gmail.com},
\texttt{enric.celaya@gmail.com}, and \texttt{lluis.ros@upc.edu}}}



\vspace{-5mm}
 
\section*{Abstract} 
In recent years, autonomous mobile platforms are finding an increasing range of applications in inspection or surveillance tasks, or to the transport of objects, in places such as smart warehouses, factories or hospitals. In these environments it is useful for the robot to have omnidirectional capability in the plane, so it can navigate through narrow or cluttered areas, or make position and orientation changes without having to maneuver. While this capability is usually achieved with directional sliding wheels, this work studies a particular robot that achieves omnidirectionality using conventional wheels, which are easier to manufacture and maintain, and support larger loads in general. This robot, which we call ``Otbot'' (for Omnidirectional tire-wheeled robot), was already conceived in the late 1990s, but all the controllers that have been proposed for it are based on purely kinematic models so far. These controllers may be sufficient if the robot is light, or if its motors are powerful, but on platforms that have to carry large loads, or that have more limited motors, it is necessary to resort to control laws based on dynamic models if the full acceleration capacities are to be exploited. This work develops a dynamic model of Otbot, proposes a plausible methodology to identify its parameters, and designs a control law that, using this model, is able to track prescribed trajectories in an accurate and robust manner.



\setcounter{tocdepth}{2} 

\tableofcontents 




\setlength{\voffset}{9mm}
\setlength{\hoffset}{-2.5mm}


\section{Introduction}
\label{ch:introduction}

\subsection{Motivation}
\label{sec:motivation}
 
\begin{figure}[b!]
	\centering
	\includegraphics[width=\linewidth]{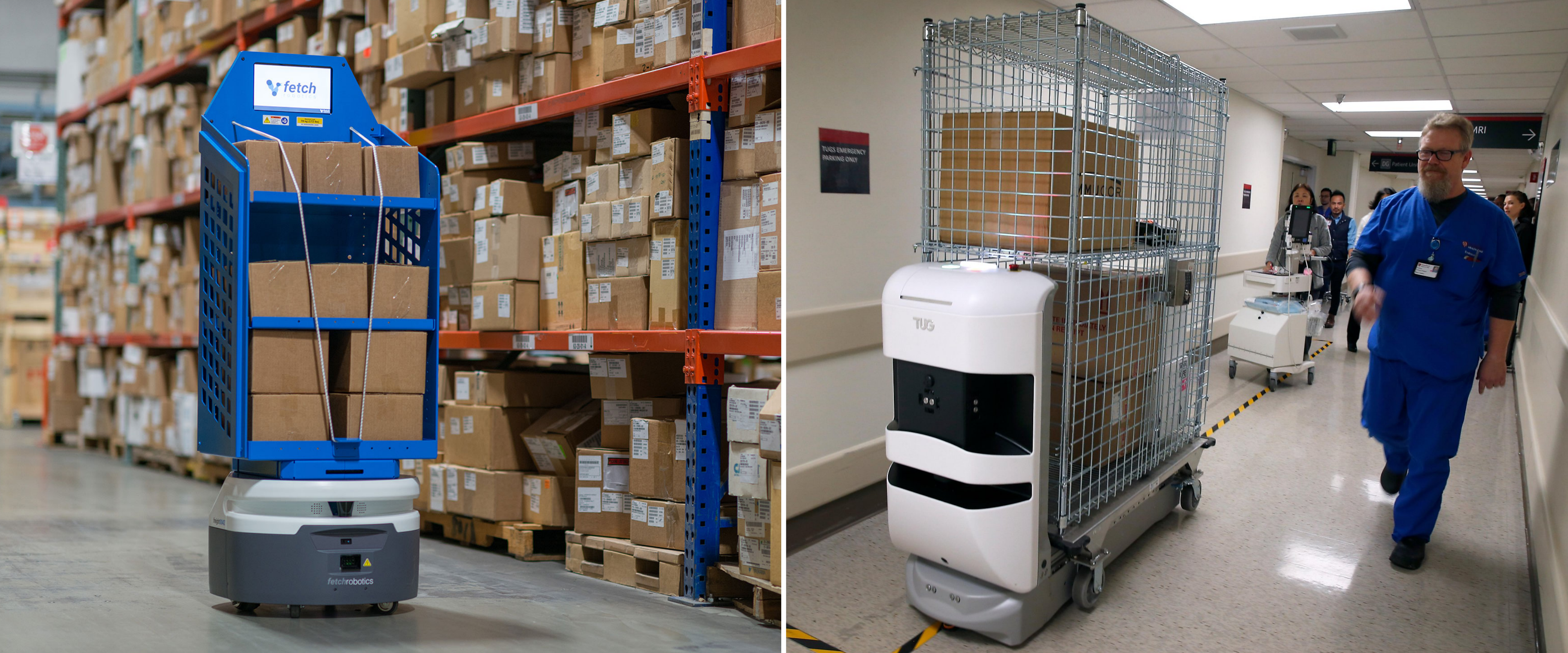}
	\caption{\label{fig:figstart} Automated material transport in a smart warehouse (left) and a hospital (right). Whereas the left robot is not omnidirectional, the one on the right achieves omnidirectionality using Mecanum wheels. Pictures courtesy of Fetch Robotics and Aethon. See \url{https://youtu.be/KyH3xXiqbUk} and \url{https://youtu.be/MLZMAW9lqXE} for details.} 
\end{figure}

In recent years, autonomous mobile platforms are finding an increasing range of applications in inspection or surveillance tasks, or to the transport of objects, in places such as smart warehouses, factories or hospitals (Fig.~\ref{fig:figstart}). In these environments it is useful for the robot to have omnidirectional capability in the plane, so it can navigate through narrow corridors or cluttered areas, or make position or orientation changes without having to maneuver. While this ability is typically achieved with Mecanum wheels~\cite{Agullo_1987,Agullo_1989,Lynch_2017}, in this paper we study a particular robot that attains omnidirectionality using conventional wheels, which are easier to manufacture and maintain, and support larger loads in general.

\begin{figure}[t!]
	\centering
	\includegraphics[width=.85\linewidth]{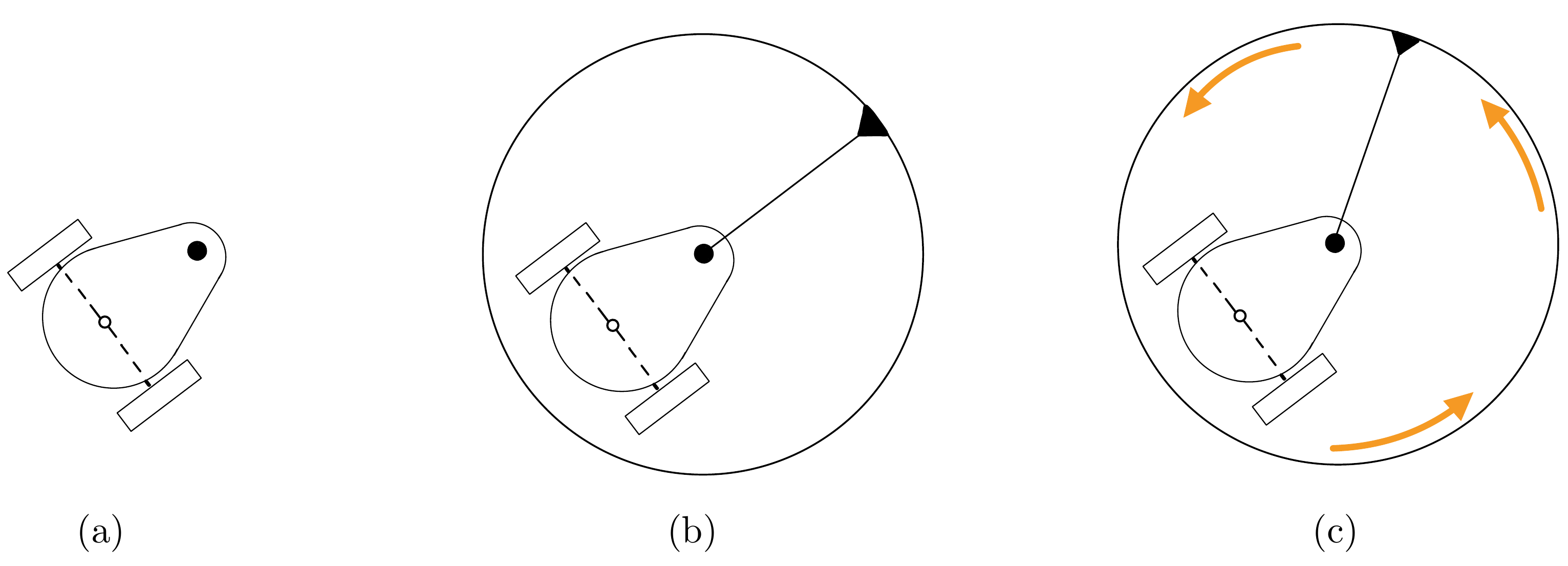}
	\caption{\label{fig:kinstrucOtbot} Otbot's kinematic structure. The black dot represents the pivot joint.} 
\end{figure}

This robot, which we call ``Otbot'' (for ``Omnidirectional tire-wheeled robot''), is composed of a rolling chassis and a circular platform mounted on top (Fig.~\ref{fig:kinstrucOtbot}). The chassis is a classic differential drive base with two wheels actuated by DC motors [Fig.~\ref{fig:kinstrucOtbot} \textbf{(a)}], and passive caster wheels for horizontal stability (not drawn). The circular platform [Fig.~\ref{fig:kinstrucOtbot} \textbf{(b)}] can rotate with respect to the chassis through a pivot joint in its center, which is controlled by another DC motor [Fig.~\ref{fig:kinstrucOtbot} \textbf{(c)}]. The offset of the pivot joint from the wheels axis is key: it allows the platform to undergo arbitrary 2D translations and rotations, and so to become omnidirectional.

Such a platform was already conceived in the 1990s~\cite{Jung:1999, Jung:2000, Jung:2002}, but all the controllers that have been proposed for it are based only on kinematic models. These controllers may be sufficient if the robot is light, or if its motors are powerful, but on platforms that have to carry large loads, or that employ more limited motors, it is necessary to resort to control laws based on dynamic models if Otbot's full acceleration capacities are to be exploited. The aim of this work is to provide new tools for the analysis and use of such capacities, in situations in which Otbot has to follow prescribed trajectories in an autonomous and accurate manner.

\subsection{Objectives}
\label{sec:purpose}

Specifically, this work pursues three objectives: 
\begin{enumerate}
\item To develop a dynamic model of Otbot that accounts for the rolling contact constraints of its chassis and the viscous friction at its motor shafts.
\item To propose a plausible methodology to identify the dynamic parameters of this model using on-board sensors of the robot.
\item To design a control law that, using the previous model, is able to track prescribed trajectories in a robust manner, under the presence of unmodelled disturbances.
\end{enumerate}

\vspace{-3mm}
\subsection{Scope of this work}
\label{sec:scope}

Usually, the development of system models and control methods is approached in two stages. In a first stage, the methods are designed and programmed, and their performance is tested in simulation. In a second stage, the methods are implemented and validated in a real system. In this work we cover the tasks of the first stage, leaving those of the second for the future. In particular, we

\begin{enumerate}
\item Obtain Otbot's dynamic model using the laws of Mechanics.
\item Propose a parameter estimation method to identify its parameters.
\item Design a control law to stabilize the system along a desired trajectory. 
\item Validate the identification and control methods in simulation.
\end{enumerate}
Therefore, the following aspects are left for future work:
\begin{enumerate}
\item The construction of a physical protoype of the robot.	
\item The development of control interfaces for all sensors and actuators.
\item The experimental validation of the methods we propose.
\end{enumerate}

\subsection{Approach}
\label{sec:approach}

Otbot is an omnidirectional vehicle whose motion control is not affected by actuation singularities. However, it is a constrained system, as the rolling contacts with the ground impose dependency constraints on the state variables. Since these constraints involve velocities and some of them cannot be integrated into a constraint involving only configuration coordinates, the robot is subject to non-holonomic constraints. This calls for the use of Lagrange's equation with multipliers to obtain the dynamical model, so further manipulations have to be applied to leave this model in a form suitable for simulation and control design purposes.

Once we have the model, we will start with the identification of its dynamical parameters. We will approach this topic by means of grey-box model identification. In this part, nonlinear optimization tools will be used to account for the full dynamic model. The estimation process will combine knowledge of the model written in terms of the unknown parameters, with information obtained from the on-board sensors of the robot. Since experimentation with a real prototype is out of our scope, we will validate the parameter identification process using a simulated model of reality. Our goal will be to see if we are able to retrieve the assumed values for the real robot parameters, under various excitation patterns and noisy sensory data.

The final goal of this work will be to design a controller able to eliminate position or velocity errors when following a trajectory. To achieve this goal we will design a computed-torque controller, which consists of a feedback linearization law combined with a standard law for the linearized system. The first law requires the inverse dynamics of the robot, which in Otbot has a particularly simple closed-form expression. The second law is then adjusted via simple pole-placement techniques, and we apply a final verification to ensure that motor torque bounds are not surpassed under the assumed tracking errors.

\subsection{Working hypotheses}
\label{sec:workhypoteses}

\subsubsection{State sensors}
\label{subsec:statesensors}

Our parameter identification and trajectory control algorithms require feedback of the pivot joint angle and the absolute pose of the platform over time. While the pivot angle can be easily obtained from an angular encoder attached to the pivot motor, the estimation of the platform pose is a bit more complex. Three common approaches to obtain it involve:
\begin{enumerate}
\item The use of a global positioning system (GPS): the platform pose is directly provided by a GPS signal, either from an indoor or an outdoor one depending on the context.	
\item The use of odometry: from the motor velocities we infer the platform velocities via forward kinematics, and then integrate such velocities to obtain the required pose.
\item The use of an inertial measurement unit (IMU): from platform acceleration and angular velocity readings we obtain the platform pose via integration.
\end{enumerate}

In this work we prefer the latter approach because the use of GPS signals makes the robot too dependent on external sensors, and resorting to odometry leads to poor pose estimations when wheel slippings occur relative to the ground. In the rest of the document, therefore, we assume the robot has an IMU sensor attached to the platform, and an angular encoder mounted at the pivot joint, which are the main sensors to be exploited by our algorithms. The only exception will be in the initial phase of our system identification procedure, which requires angular encoders on the wheels to identify the wheel moments of inertia and the viscous friction coefficients of the robot shafts.

\subsubsection{Dynamic model}
\label{subsec:dynamicmodelintro}

When deriving Otbot's dynamic model we have made the following assumptions: 

\begin{itemize}
	\setlength{\itemsep}{0\baselineskip}
	\item All servomotors of the robot accept and can follow torque commands.
	\item The dynamic effects of the passive caster wheels can all be neglected.
	\item The wheels establish a perfect rolling contact with the ground, so slidings do not occur in principle.
	\item The robot is considered to move on flat terrain, which implies that the gravitational term is null in the dynamic model. 
	\item The only frictional forces that will be considered are those due to viscous friction of the motors. The rest of friction forces are supposed to be negligible. This includes the viscous friction of the robot against the air, the static friction at the motors, and the rolling resistance of the wheels on the ground. 
\end{itemize}

\subsubsection{Control system}
\label{subsec:controlsystem}

Finally, we have assumed that the frequency at which the control loop operates is fast enough. This means that the state measurements and the evaluation of the control law are performed in a very short time, so the global system formed by the robot and its controller can be assimilated to a continuous-time system.

\subsection{Project context}
\label{sec:context}

The work we report was carried out from July 2020 to May 2021 in the context of the MSc thesis of the first author, advised by the 2nd and 3rd authors. The content of this paper corresponds to the one in \cite{giro2021mscthesis} with minor modifications. In the same period of time, Maxime Gautier from École Polytechnique Fédérale de Lausanne (EPFL) also teamed with the authors and carried out his MSc thesis, which developed collocation-based methods for the planning of optimal trajectories for Otbot~\cite{gautier2021mscthesis}. Mr. Gautier's thesis provides tools complementary to those we present.

\subsection{Notation}
\label{sec:Notconvent}

Throughout the document we write scalar magnitudes using normal font, and vector ones using bold font. Vectors and matrices are denoted with lowercase and uppercase letters, respectively. Thus, $x$ denotes a scalar, $\vr{x}$ a vector, and $\mt{X}$ a matrix. When $\vr{x}$ is a vector that appears in a mathematical operation, it is assumed to be a column vector.

When we have to make the components of a vector explicit, we normally display them in column form using square brackets:
\begin{equation}
\vr{x} = \left[
\begin{array}{c}
x_1\\
\vdots\\
x_n\\
\end{array}
\right].
\label{eq:example}
\end{equation}
In some places, we also write $\vr{x} = (x_1, \ldots, x_n)$ for convenience, which we assume to be equivalent to Eq.~\eqref{eq:example}. Similarly, we use $\vr{x} = (\vr{x}_1, \ldots, \vr{x}_n)$ to refer to the linear concatenation of the vectors $\vr{x}_1, \ldots, \vr{x}_n$. This expression is meant to be equivalent to $\vr{x} = \left[ \vr{x}_1 \trans, \ldots, \vr{x}_n \trans \right]\trans$, where $\vr{x}_1, \ldots, \vr{x}_n$ are column vectors. 

We sometimes use 
\begin{equation*}
\underbrace{\text{expression}}_{\text{symbol}}
\label{eq:example2}
\end{equation*}
to mean that, from this point onwards, the shown symbol will denote the underbraced expression. Occasionally, we also use $s_{\alpha}$ and $c_{\alpha}$ to refer to $\sin \alpha$ and $\cos \alpha$.

\subsection{Document structure}
\label{sec:docstructure}

\noindent The rest of this document is structured as follows: 

\begin{itemize}
	\setlength{\itemsep}{0\baselineskip}
	\item In Section~\ref{ch:kinematicmodel} we explain how the kinematic model of Otbot has been obtained. This model describes the feasible motions of the robot without considering their generating forces. The resulting equations are necessary to later develop the robot's dynamic model. 
	\item In Section~\ref{ch:dyn_model} we obtain the dynamic model of Otbot. By this model we mean the equations of motion that relate the accelerations of the robot to its motor torques. Solutions to both the forward and inverse dynamical problems are provided to later use them for simulation and control design purposes.
	\item In Section~\ref{ch:chap4ParamIdent} we propose a grey-box model identification process to estimate the dynamical parameters of Otbot. We explain how this process can be implemented using nonlinear regression on noisy readings from the IMU and pivot joint sensors, and validate the entire process in simulation. 
	\item In Section~\ref{ch:trackingcontrl} we deal with the topic of tracking control. We show how, by using feedback linearization, it is possible to design a computed-torque controller that robustly keeps the robot along a desired trajectory. We also explain how to tune this controller to stabilize the error signal of the robot in a given time.
	\item Finally, in Section~\ref{ch:conclusions} we provide the main conclusions of this work, and list several points for further attention. 
\end{itemize}
\section{Kinematic model}
\label{ch:kinematicmodel}

A kinematic model describes the feasible motions of a robot regardless of their generating forces and torques. This section starts with a description of the reference frames and state coordinates used to describe Otbot, and then formulates the kinematic constraints imposed by the wheel contacts with the ground, as well as the solutions to the forward and inverse instantaneous kinematic problems. The resulting expressions will be used in Section~\ref{ch:dyn_model} to obtain the dynamic model for the robot. The analysis of the inverse instantaneous kinematic Jacobian will also reveal that the platform can undergo arbitrary velocities in the plane, which proves its omnidirectional capability. 

\subsection{Reference frames and state coordinates}
\label{sec:Refframesstatecoords}

\begin{figure}[b!]
	\centering
	\includegraphics[width=0.65\linewidth]{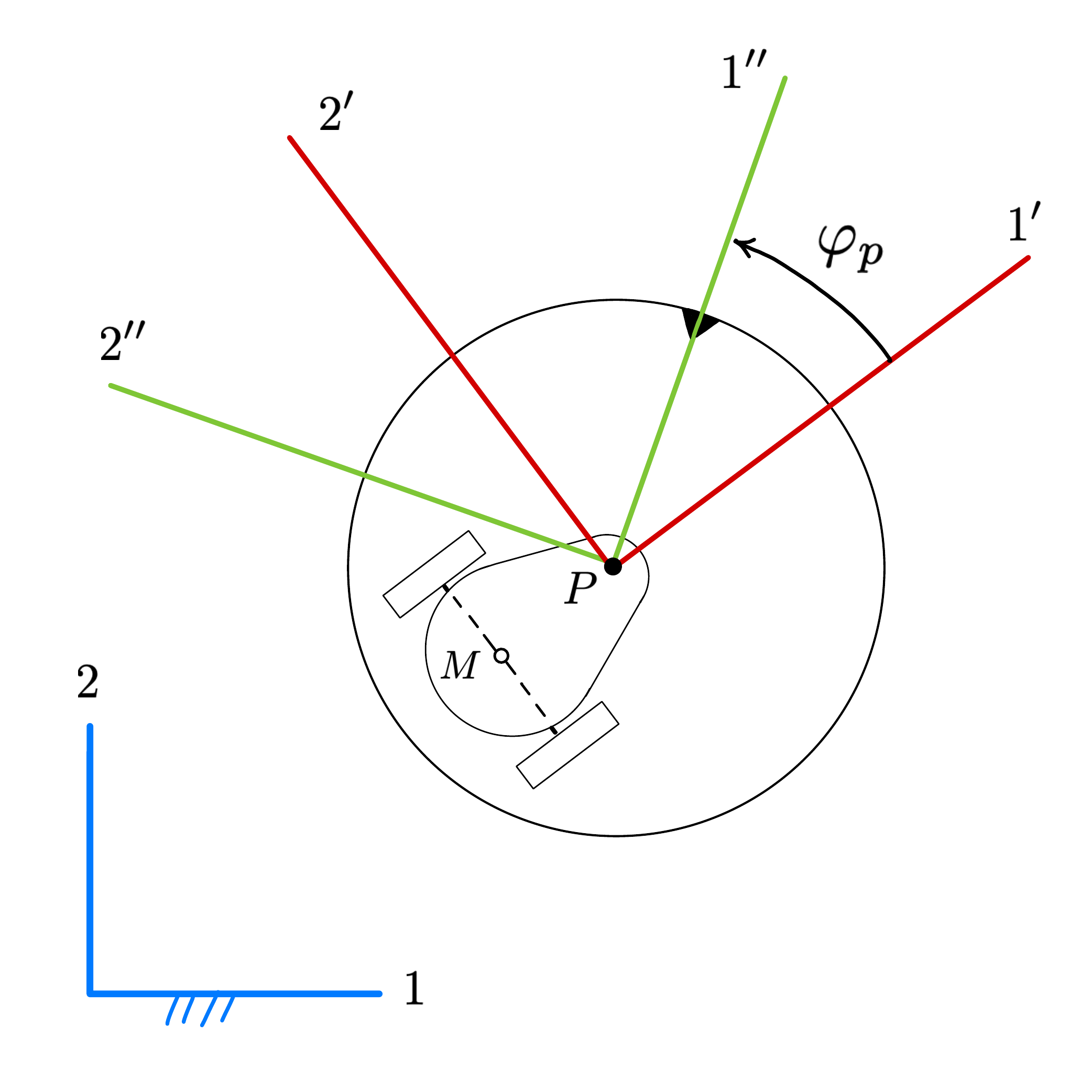}
	\caption{\label{fig:Refsframes} Reference frames used in the models.} 
\end{figure}

To obtain the kinematic and dynamic models of the robot, we will use the reference frames shown in Fig.~\ref{fig:Refsframes}. The blue frame is the absolute frame fixed to the ground. The red and green frames are fixed to the chassis and the platform, respectively. Point $M$ is the midpoint of the wheels axis and point $P$ is the location of the pivot joint. The red axis $1'$ is always aligned with $M$ and $P$. The angle between the green and red frames coincides with the pivot joint angle $\varphi_p$.

Each frame has a vector basis attached to it, whose unit vectors are directed along the frame axes. The three bases will be referred to as $B = \{1, 2, 3\}$, $B' = \{1', 2', 3'\}$ and $B'' = \{1'', 2'', 3''\}$ respectively.

\begin{figure}[b!]
	\centering
	\includegraphics[width=0.80\linewidth]{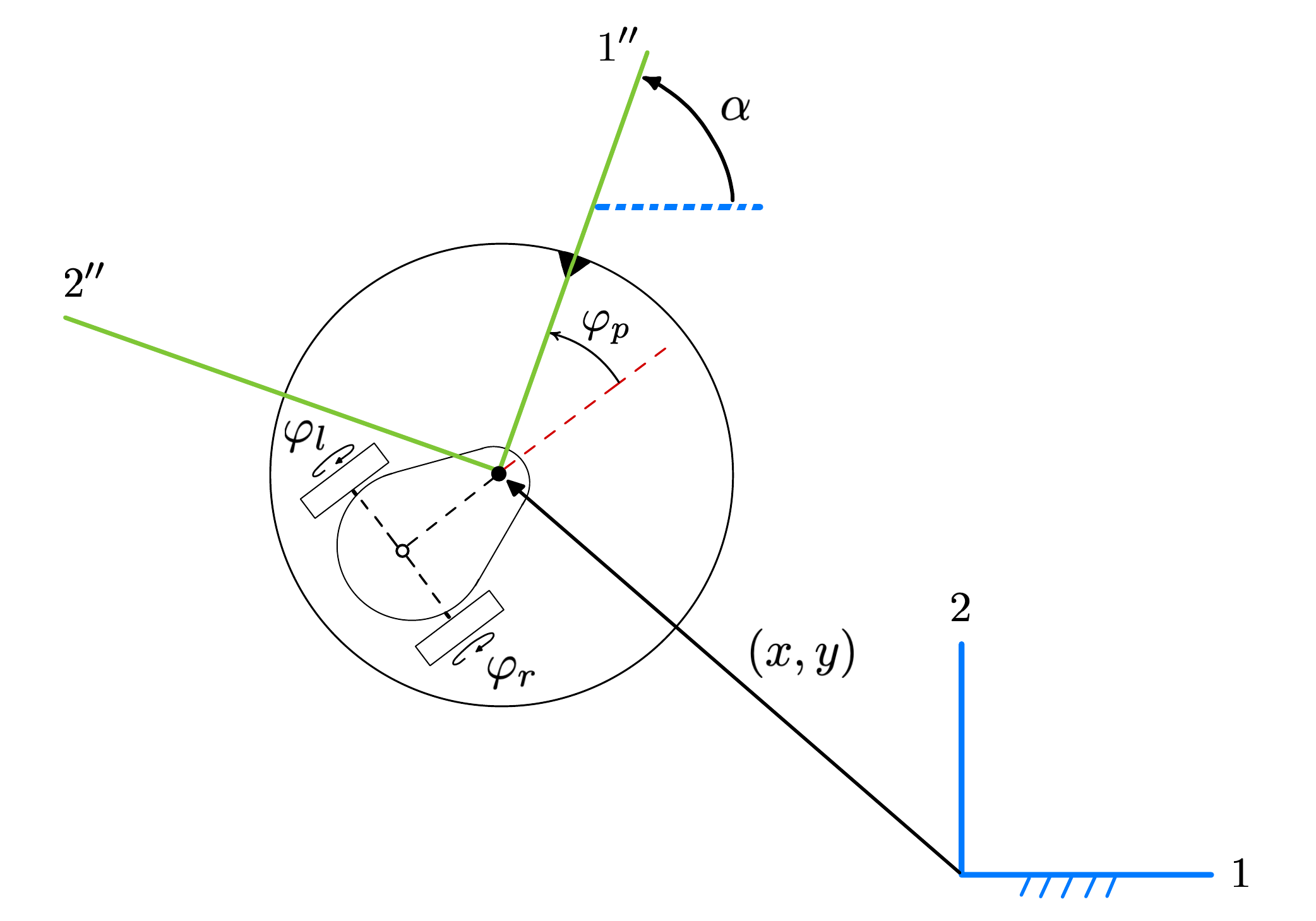}
	\caption{\label{fig:confignstatecoords} Otbot's configuration and state coordinates.} 
\end{figure}

The robot configuration can be described by the following six coordinates (Fig.~\ref{fig:confignstatecoords}):
\begin{itemize}
	\setlength{\itemsep}{0\baselineskip}
	\item The absolute position  of the pivot joint $(x,y)$.
	\item The absolute angle $\alpha$ of the platform.
	\item The pivot angle $\varphi_p$ between the platform and the chassis. 
	\item The angles of the right and left wheels relative to the chassis, $\varphi_r$ and $\varphi_l$.
\end{itemize}
The robot configuration is thus given by
\begin{equation}
\vr{q} = (x,y,\alpha, \varphi_r,\varphi_l,\varphi_p),
\label{eq:chap2eqQ}
\end{equation}
and its time derivative provides the robot velocity: 
$$\vr{\dot{q}} = (\dot{x},\dot{y},\dot{\alpha}, \dot{\varphi}_r,\dot{\varphi}_l,\dot{\varphi}_p).$$
Therefore, the robot state is given by 
$$\vr{x} = (\vr{q},\vr{\dot{q}}) = (x,y,\alpha, \varphi_r,\varphi_l,\varphi_p,\dot{x},\dot{y},\dot{\alpha}, \dot{\varphi}_r,\dot{\varphi}_l,\dot{\varphi}_p).$$

\subsection{Kinematic constraints}
\label{sec:Kinconst}

The rolling contacts of the wheels with the ground impose a kinematic constraint of the form
\begin{equation}
\mt{J}(\vr{q}) \; \vr{\dot{q}} = \vr{0},
\label{eq:eq_jq=0}      
\end{equation}
where $\mt{J}(\vr{q}) $ is a $3 \times 6$ Jacobian matrix. We shall derive this constraint in two steps. First we will obtain the kinematic constraints imposed by the wheels, and then we will rewrite these constraints using the $\vr{x}$ variables only. We next see these two steps in detail. In our derivations, we will use $\vr{v}(Q)$ to denote the velocity of a point $Q$ of the robot. The vector basis in which $\vr{v}(Q)$ is expressed will be mentioned explicitly, or understood by context.

\subsubsection{Kinematic constraints of the differential drive}
\label{subsec:kinconsdifdrive}

\begin{figure}[t!]
	\centering
	\includegraphics[width=\linewidth]{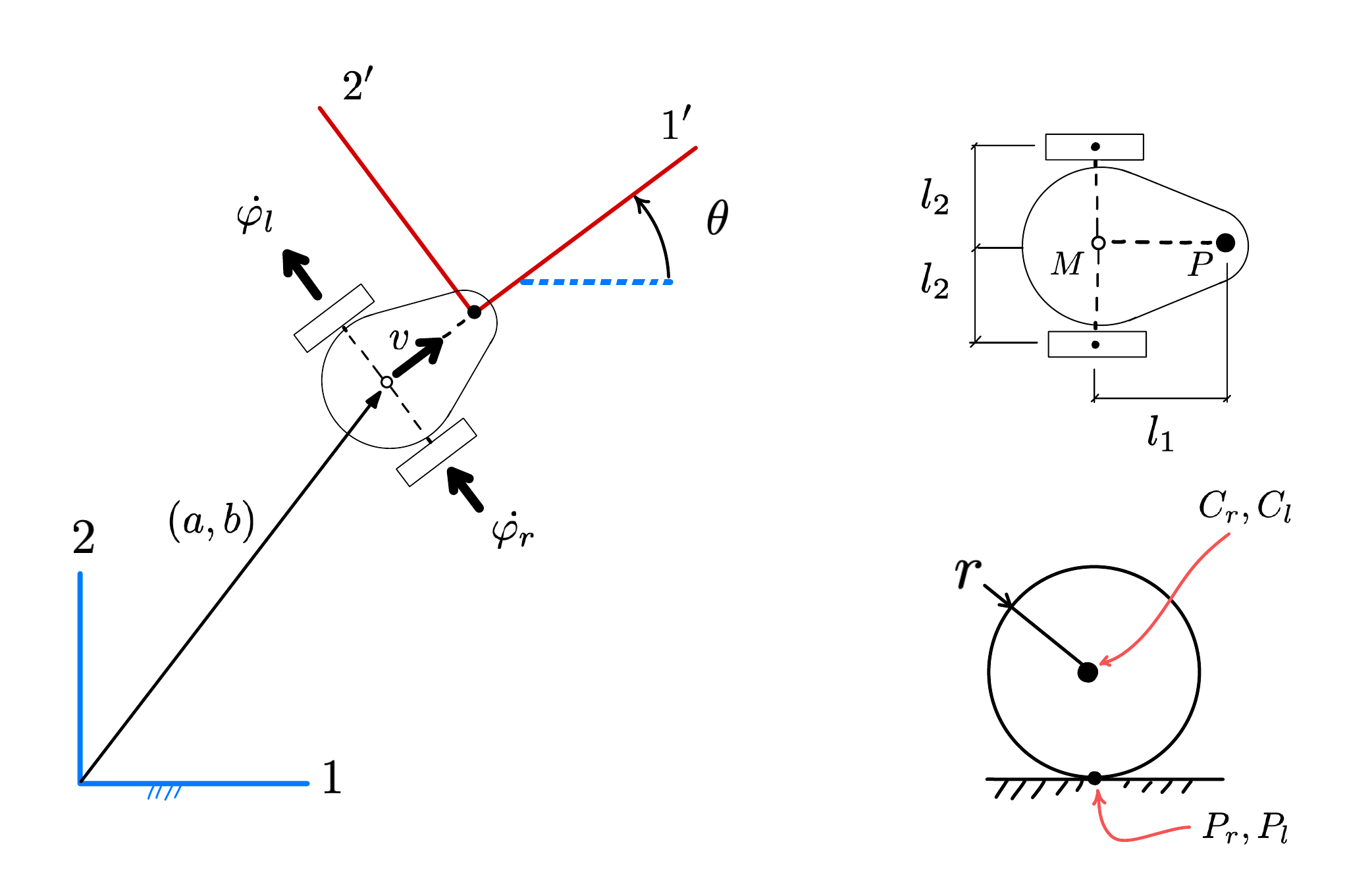}
	\caption{\label{fig:Kinconstrdifdrive} Notation used in the kinematic modelling of Otbot's chassis.} 
\end{figure}
For the moment, let us neglect the platform and focus our attention on the chassis, as seen in Fig.~\ref{fig:Kinconstrdifdrive}. The chassis pose is given by the position vector $(a,b)$ of point M, and by the orientation angle $\theta$. We use $l_1$ and $l_2$ to refer to the pivot offset from M and the half-length of the wheels axis, respectively. Also, $C_r$ and $C_l$ denote the centers of the right and left wheels, and $P_r$ and $P_l$ are the contact points of the wheels with the ground. The two wheels have the same radius $r$. 

The rolling contact constraints of the chassis can be found by computing the velocities $\vr{v}(P_r)$ and $\vr{v}(P_l)$ in terms of $\dot{a}$, $\dot{b}$, $\dot{\theta}$, $\dot{\varphi}_r$, and $\dot{\varphi}_l$ (i.e., as if the robot were a floating kinematic tree) and forcing these velocities to be zero (as the wheels do not slip when placed on the ground).
 
To obtain $\vr{v}(P_r)$ and $\vr{v}(P_l)$, note first that the velocity of $M$ can only be directed along axis $1'$, since lateral slipping is forbidden under perfect rolling. Therefore in $B' = \{1',2',3'\}$ we have
\begin{equation*}
\vr{v}(M) = \left[
\begin{array}{c}
v\\
0\\
0
\end{array}
\right].
\label{eq:vofM}
\end{equation*}
Also note that, if $\gvr{\omega}_{\mathrm{chassis}}$ is the angular velocity of the chassis, we have
\begin{equation*}
\vr{v}(C_r) = \vr{v}(M) + \gvr{\omega}_{\mathrm{chassis}} \times \vr{MC}_r = 
\left\lbrack 
\begin{array}{c}
v \\
0 \\
0
\end{array}
\right\rbrack
+
\left\lbrack 
\begin{array}{c}
0 \\
0 \\
\dot{\theta} 
\end{array}
\right\rbrack 
\times 
\left\lbrack 
\begin{array}{c}
0 \\
-l_2 \\
0
\end{array}
\right\rbrack
=
\left\lbrack 
\begin{array}{c}
v + l_2\;\dot{\theta} \\
0 \\
0
\end{array}
\right\rbrack
\label{eq:vofCr}
\end{equation*}
\begin{equation*}
\vr{v}(C_l) = 
\vr{v}(M) + \gvr{\omega}_{\mathrm{chassis}} \times \vr{MC}_l = 
\left\lbrack 
\begin{array}{c}
v \\
0 \\
0
\end{array}
\right\rbrack
+
\left\lbrack 
\begin{array}{c}
0 \\
0 \\
\dot{\theta} 
\end{array}
\right\rbrack 
\times 
\left\lbrack 
\begin{array}{c}
0 \\
l_2 \\
0
\end{array}
\right\rbrack
=
\left\lbrack 
\begin{array}{c}
v-l_2\;\dot{\theta} \\
0 \\
0
\end{array}
\right\rbrack.
\label{eq:vofCl}
\end{equation*}
The velocities of the ground contact points are thus given by
\begin{align*}
\begin{split}
\vr{v}(P_r) 
&=
\vr{v}(C_r) +
\gvr{\omega}_{\text{wheel}} 
\times 
\vr{C}_r \vr{P}_r \\
&= 
\left\lbrack 
\begin{array}{c}
v + l_2 \; \dot{\theta} \\
0 \\
0
\end{array}
\right\rbrack
+
\left\lbrack 
\begin{array}{c}
0 \\
\dot{\varphi}_r \\
\dot{\theta} 
\end{array}
\right\rbrack 
\times 
\left\lbrack 
\begin{array}{c}
0 \\
0 \\
-r
\end{array}
\right\rbrack
=
\left\lbrack 
\begin{array}{c}
v + l_2 \; \dot{\theta} - r \; \dot{\varphi}_r \\
0 \\
0
\end{array}
\right\rbrack
\end{split}
\end{align*}

\begin{align*}
\begin{split}
\vr{v}(P_l) 
&=
\vr{v}(C_l) 
+ 
\gvr{\omega}_{\mathrm{wheel}}  
\times  
\vr{C}_l \vr{P}_l \\
&= 
\left\lbrack 
\begin{array}{c}
v - l_2 \; \dot{\theta} \\
0 \\
0
\end{array}
\right\rbrack
+
\left\lbrack 
\begin{array}{c}
0 \\
\dot{\varphi}_l \\
\dot{\theta} 
\end{array}
\right\rbrack 
\times 
\left\lbrack 
\begin{array}{c}
0 \\
0 \\
-r
\end{array}
\right\rbrack
=
\left\lbrack 
\begin{array}{c}
v - l_2 \; \dot{\theta} - r \; \dot{\varphi}_l \\
0 \\
0
\end{array}
\right\rbrack.
\end{split}
\end{align*}

Since $P_r$ and $P_l$ do not slip, $\vr{v}(P_r)$ and $\vr{v}(P_l)$ must be zero, which gives us the two fundamental constraints of the robot:
\begin{equation*}
v + l_2 \dot{\theta} -r\dot{\varphi}_r =0
\label{eq:nonslip1}
\end{equation*}
\begin{equation*}
v - l_2 \dot{\theta} -r\dot{\varphi}_l =0.
\label{eq:nonslip2}
\end{equation*}
It is now easy to solve for $v$ and $\dot{\theta}$ in these two equations:
\begin{align}
&v=\frac{r\left( \dot{\varphi}_l + \dot{\varphi}_r\right)}{2}
\label{eq:solutionv} \\
&\dot{\theta} = - \frac{r\left( \dot{\varphi}_l - \dot{\varphi}_r\right)}{2l_2}
\label{eq:solutionthetadot}
\end{align}

Note from Fig.~\ref{fig:Kinconstrdifdrive} that in basis $B = \{1, 2, 3\}$ we have
\begin{equation*}
\vr{v}(M) 
= 
\left\lbrack 
\begin{array}{c}
\dot{a} \\
\dot{b} \\
0
\end{array}
\right\rbrack,
\label{eq:vinbasisB}
\end{equation*}
and that it must be
\begin{equation*}
\left\lbrace
\begin{array}{l}
\dot{a} = v \cdot \cos{\theta} \\
\dot{b} = v \cdot \sin{\theta}
\end{array}
\right.
\label{eq:projections}
\end{equation*}
By substituting Eq.~\eqref{eq:solutionv} in these two equations, and adding Eq.~\eqref{eq:solutionthetadot}, we obtain
\begin{equation}
\left\lbrace
\begin{array}{l}
\dot{a} =\mathrm{cos}\left(\theta \right)\left(\dfrac{r\dot{\varphi}_l }{2}+\dfrac{r\dot{\varphi}_r }{2}\right)\vspace{1mm}\\
\dot{b} =\mathrm{sin}\left(\theta \right)\left(\dfrac{r\dot{\varphi}_l }{2}+\dfrac{r\dot{\varphi}_r }{2}\right)\vspace{1mm}\\
\dot{\theta} =-\dfrac{r\left(\dot{\varphi}_l -\dot{\varphi}_r \right)}{2l_2} 
\end{array}
\right.
\label{eq:systemofequations}
\end{equation}
This system provides the instantaneous forward kinematic solution for the differential drive, but it can also be viewed as a set of equations that expresses the rolling contact constraints of the chassis.

\subsubsection{Kinematic constraints of the whole robot}
\label{subsec:kinconswholerobot}

To obtain the kinematic constraints of Otbot itself, we just need to rewrite the previous system using the $\vr{x}$ variables only. This entails substituting $\theta$, $\dot{\theta}$, $\dot{a}$, and $\dot{b}$ by their expressions in terms of $\vr{x}$ coordinates. 

\begin{figure}[t!]
	\centering
	\includegraphics[width=0.75\linewidth]{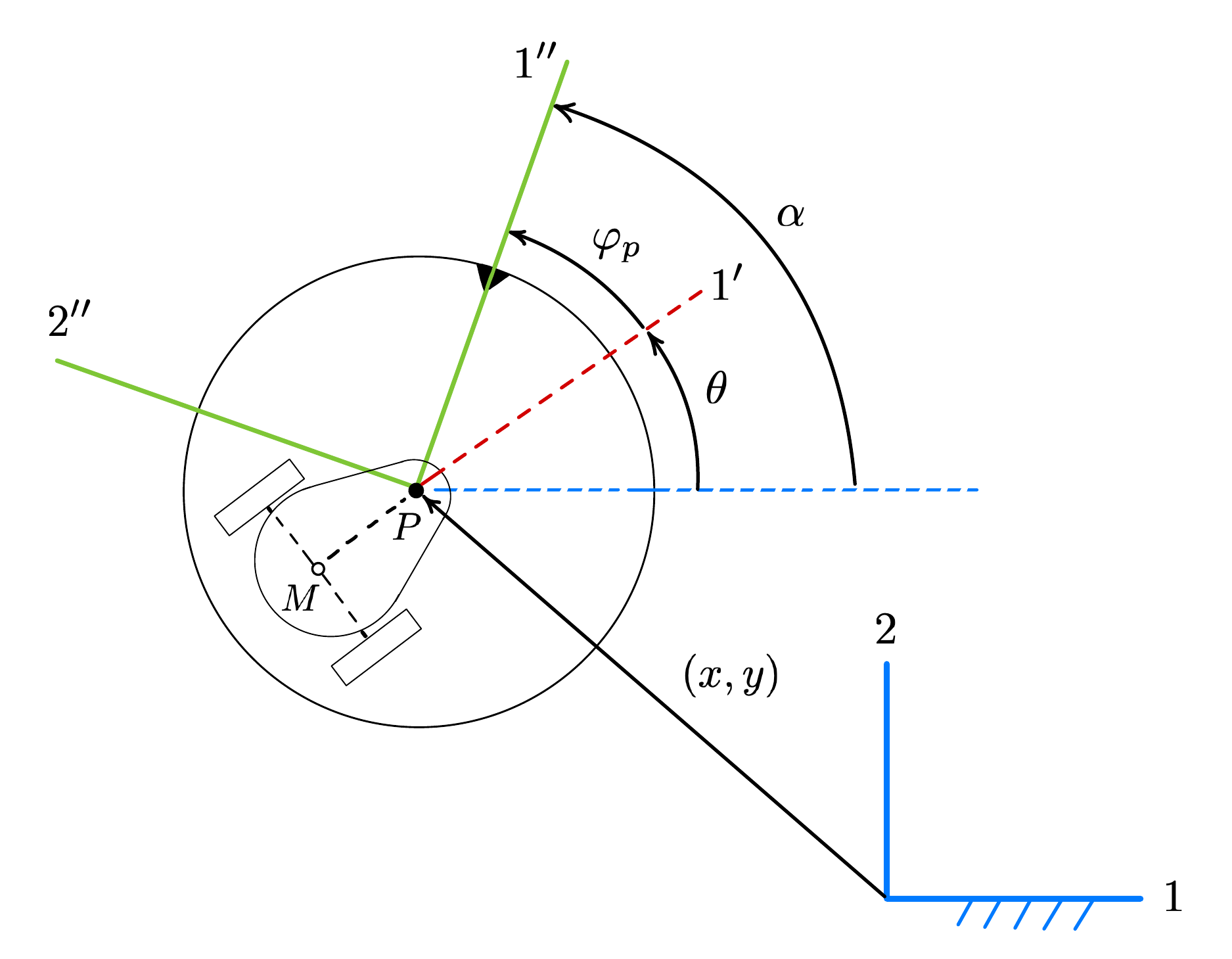}
	\caption{\label{fig:anglesOtbot} Relationship between $\alpha$, $\varphi_p$, and $\theta$.} 
\end{figure} 

From Fig.~\ref{fig:anglesOtbot} it is clear that
\begin{equation*}
\begin{array}{l}
\theta =\alpha -\varphi_p\\
\dot{\theta} = \dot{\alpha} -\dot{\varphi}_p
\end{array}
\label{eq:thetanthetadot}
\end{equation*}
and we also see that
\begin{equation*}
\vr{v}(M) = \vr{v}(P) + \omega_{\mathrm{chassis}} \times \vr{PM},
\label{eq:vofMnP}
\end{equation*}
so using $B = \{1, 2, 3\}$ we can write
 
\begin{align*}
\left\lbrack 
\begin{array}{c}
\dot{a} \\
\dot{b} \\
0
\end{array}
\right\rbrack
&
=
\left\lbrack 
\begin{array}{c}
\dot{x} \\
\dot{y} \\
0
\end{array}
\right\rbrack
+
\left\lbrack 
\begin{array}{c}
0 \\
0 \\
\dot{\theta} 
\end{array}
\right\rbrack 
\times 
\left\lbrack 
\begin{array}{c}
-l_1 \cdot \cos{\theta} \\
-l_1 \cdot \sin{\theta} \\
0
\end{array}
\right\rbrack
\nonumber \\
&
=
\left\lbrack 
\begin{array}{c}
\dot{x} + \dot{\theta} \; l_1 \sin{\theta} \\
\dot{y} - \dot{\theta} \; l_1 \cos{\theta} \\
0
\end{array}
\right\rbrack
=
\left\lbrack 
\begin{array}{c}
\dot{x} + (\dot{\alpha}-\dot{\varphi}_p) \; l_1 \; \sin{(\alpha-\varphi_p)} \\
\dot{y} - (\dot{\alpha}-\dot{\varphi}_p) \; l_1 \; \cos{(\alpha-\varphi_p)} \\
0
\end{array}
\right\rbrack.
\label{eq:vectorialrelationships}
\end{align*}
In sum we have the relationships
\begin{equation*}
\left\lbrace
\begin{array}{l}
\theta = \alpha - \varphi_p \\
\dot{\theta} = \dot{\alpha} - \dot{\varphi}_p \\
\dot{a} = \dot{x} + (\dot{\alpha}-\dot{\varphi}_p) \; l_1 \; \sin{(\alpha-\varphi_p)} \\
\dot{b} = \dot{y} - (\dot{\alpha}-\dot{\varphi}_p) \; l_1 \; \cos{(\alpha-\varphi_p)} 
\end{array}
\right.
\label{eq:equationrelationships}
\end{equation*}
which give us the desired values of  $\theta$, $\dot{\theta}$, $\dot{a}$, and $\dot{b}$ in terms of $\vr{q}$ and $\vr{\dot{q}}$. We thus can substitute these expressions in Eq.~\eqref{eq:systemofequations} to obtain:
\begin{equation}
\dot{x} +l_1 \,\mathrm{sin}\left(\alpha -\varphi_p \right)\,{\left(\dot{\alpha} -\dot{\varphi}_p \right)}=\mathrm{cos}\left(\alpha -\varphi_p \right)\,{\left(\frac{r\dot{\varphi}_l }{2}+\frac{r\dot{\varphi}_r }{2}\right)},
\label{eq:Kineq8}
\end{equation}
\begin{equation}
\dot{y} -l_1 \,\mathrm{cos}\left(\alpha -\varphi_p \right)\,{\left(\dot{\alpha} -\dot{\varphi}_p \right)}=\mathrm{sin}\left(\alpha -\varphi_p \right)\,{\left(\frac{r\dot{\varphi}_l }{2}+\frac{r\dot{\varphi}_r }{2}\right)},
\label{eq:Kineq9}
\end{equation}
\begin{equation}
\dot{\alpha} -\dot{\varphi}_p =-\frac{r\,{\left(\dot{\varphi}_l -\dot{\varphi}_r \right)}}{2\,l_2 }.
\label{eq:Kineq10}
\end{equation}
These are Otbot's \textit{kinematic constraints} expressed in $\vr{x}$ coordinates. Additionally, by re-ordering and integrating the last equation of this system, we can obtain an holonomic constraint of our system
\begin{equation}
\alpha-\frac{r}{2 \ell_{2}} \varphi_{r}+\frac{r}{2 \ell_{2}} \varphi_{l}-\varphi_{p} - K_{0}=0,
\label{eq:eq_holo_cstr}
\end{equation}
where 
$$K_{0} = \alpha(0)-\frac{r}{2 \ell_{2}} \varphi_{r}(0)+\frac{r}{2 \ell_{2}} \varphi_{e}(0)-\varphi_{p}(0).$$ 

In matrix form, Eqs.~\eqref{eq:Kineq8} -~\eqref{eq:Kineq10} can be written as
\begin{equation*}
\mt{J}\left(\vr{q}\right) \cdot \vr{\dot{q}} = \vr{0},
\label{eq:Kinconsrtvec}
\end{equation*}
where
\begin{equation}
\mt{J}(\vr{q}) = \left[
\begin{array}{cccccc}
1 & 0 & l_1s_{\alpha - \varphi_p} & - \frac{1}{2}rc_{\alpha - \varphi_p} & - \frac{1}{2}rc_{\alpha - \varphi_p} & -l_1s_{\alpha - \varphi_p}\\
0 & 1 & -l_1c_{\alpha - \varphi_p} & - \frac{1}{2}rs_{\alpha - \varphi_p} & - \frac{1}{2}rs_{\alpha - \varphi_p} & l_1c_{\alpha - \varphi_p}\\
0 & 0 & 1 & -\frac{r}{2l_2} & \frac{r}{2l_2} & -1
\end{array}
\right]
\label{eq:chap2JacobianJ}
\end{equation}

\subsection{Solution of the instantaneous kinematic problems}
\label{sec:IKP}

The configuration $\vr{q}$ in Eq.~\eqref{eq:chap2eqQ} can be partitioned into two vectors, $\vr{p}=(x,y,\alpha)$ and $\gvr{\varphi} = (\varphi_r,\varphi_l,\varphi_p)$, which provide the task and joint space coordinates of the robot. Their time derivatives  $\gvr{\dot{\varphi}}$ and $\vr{\dot{p}}$ are called the motor speeds and the platform twist, respectively. We next solve the following two problems:
\begin{itemize}
\item Forward instantaneous kinematic problem (FIKP): Given $\gvr{\dot{\varphi}}$, obtain $\vr{\dot{p}}$.
\item Inverse instantaneous kinematic problem (IIKP): Given $\vr{\dot{p}}$, obtain $\gvr{\dot{\varphi}}$.
\end{itemize}
We shall see that the IIKP is always solvable regardless of the robot configuration, which means that the platform can move omnidirectionally in the plane.

\subsubsection{Forward problem}
\label{subsec:FKP}

We only have to isolate $\dot{x}$, $\dot{y}$, and $\dot{\alpha}$ from Eqs.~\eqref{eq:Kineq8} -~\eqref{eq:Kineq10}:
\begin{equation*}
\dot{x} =\frac{l_2 \, r \, (\dot{\varphi}_l+\dot{\varphi}_r) \, \cos(\alpha-\varphi_p) + l_1 \, r \, (\dot{\varphi}_l-\dot{\varphi}_r) \, \sin(\alpha-\varphi_p)}{2\,l_2}	
\label{eq:Kineq11}
\end{equation*}
\begin{equation*}
\dot{y} =\frac{l_1 \, r \, (\dot{\varphi}_r-\dot{\varphi}_l) \, \cos(\alpha-\varphi_p) + l_2 \, r \, (\dot{\varphi}_l+\dot{\varphi}_r) \, \sin(\alpha-\varphi_p)}{2\,l_2}
\label{eq:Kineq12}
\end{equation*}
\begin{equation*}
\dot{\alpha} =\frac{2\,l_2 \dot{\varphi}_p -r\dot{\varphi}_l +r\dot{\varphi}_r }{2\,l_2 }
\label{eq:Kineq13}
\end{equation*}
These equations directly provide $\vr{\dot{p}}$ as a function of $\boldsymbol{\dot{\varphi}}$. This function can be expressed as
\begin{equation*}
\left\lbrack \begin{array}{c}
\dot{x} \\
\dot{y} \\
\dot{\alpha} 
\end{array}\right\rbrack = 
\mt{M}_{FIK}(\vr{q})
\;
\left\lbrack 
\begin{array}{c}
\dot{\varphi}_r \\
\dot{\varphi}_l \\
\dot{\varphi}_p 
\end{array}
\right\rbrack,
\label{eq:MaricialFIKP}
\end{equation*}
where $\mt{M}_{FIK}(\vr{q})$ has the expression
\begin{equation*}
\frac{r}{2\,l_2}
\left\lbrack \begin{array}{ccc}
	l_2 \, c_{\alpha-\varphi_p} - l_1 \, s_{\alpha-\varphi_p} &
	l_2 \, c_{\alpha-\varphi_p} + l_1 \, s_{\alpha-\varphi_p} &
	0
	\\
	l_1 \, c_{\alpha-\varphi_p} + l_2 \, s_{\alpha-\varphi_p} &
   -l_1 \, c_{\alpha-\varphi_p} + l_2 \, s_{\alpha-\varphi_p} &
	0
	\\
	1 & 
   -1 &
   \frac{2\,l_2}{r}
\end{array}\right\rbrack
\label{eq:MFIKsym}
\end{equation*}

\subsubsection{Inverse problem}
\label{subsec:IKP}

We now wish to find $\boldsymbol{\dot{\varphi}}$ as a function of $\vr{\dot{p}}$.  Clearly, this function is given by
\begin{equation*}
\left\lbrack 
\begin{array}{c}
\dot{\varphi}_r \\
\dot{\varphi}_l \\
\dot{\varphi}_p 
\end{array}
\right\rbrack
= 
\mt{M}_{IIK}(\vr{q})
\;
\left\lbrack \begin{array}{c}
\dot{x} \\
\dot{y} \\
\dot{\alpha} 
\end{array}\right\rbrack, 
\label{eq:MaricialIIKP}
\end{equation*}
where $\mt{M}_{IIK}(\vr{q}) = \mt{M}_{FIK}^{-1}(\vr{q})$ takes the form
\vspace{2mm}
\begin{equation*}
\frac{1}{r \, l_1}
\left\lbrack \begin{array}{ccc}
	l_1 \, c_{\alpha-\varphi_p} - l_2 \, s_{\alpha-\varphi_p} &
	l_2 \, c_{\alpha-\varphi_p} + l_1 \, s_{\alpha-\varphi_p} &
	0 
	\\
	l_1 \, c_{\alpha-\varphi_p} + l_2 \, s_{\alpha-\varphi_p} &
   -l_2 \, c_{\alpha-\varphi_p} + l_1 \, s_{\alpha-\varphi_p} &
	0
	\\
	r \, s_{\alpha-\varphi_p} &
   -r \, c_{\alpha-\varphi_p} &
	r \, l_1
\end{array}\right\rbrack
\label{eq:MIIKsym}
\end{equation*}
\vspace{2mm}

Note at this point that the determinant of $\mt{M}_{FIK}(\vr{q})$ is
\begin{equation*}
\det(\mt{M}_{FIK}(\vr{q})) = -\frac{l_1 \,r^2 }{2\,l_2 }.
\label{eq:MFIKdet}
\end{equation*}
Since $r$, $l_1$, and $l_2$ are all positive in Otbot, we see that $\mt{M}_{IIK}(\vr{q}) = \mt{M}_{FIK}^{-1}(\vr{q})$ always exists, so the IIKP is solvable in all configurations of the robot. An important consequence of this fact is that the platform is able to move under any velocity $\vr{\dot{p}}$ in the plane. In other words, it is omnidirectional.\\ 

\subsection{The kinematic model in control form}
\label{sec:KMincntrlform}

The following equations
\begin{equation*}
\left\lbrace 
\begin{array}{l}
\vr{\dot{p}} = 
\mt{M}_{FIK}(\vr{q}) 
\;
\boldsymbol{\dot{\varphi}} 
\\
\boldsymbol{\dot{\varphi}} = \boldsymbol{\dot{\varphi}} 
\end{array}
\right.
\label{eq:KinematicModel}
\end{equation*}
can be written in the usual form assumed in control engineering,
\begin{equation}
\vr{\dot{q}} = \vr{f}_{\mathrm{kin}}(\vr{q},\gvr{\omega}),
\label{eq:cntrleng}
\end{equation}
where $\gvr{\omega} = \boldsymbol{\dot{\varphi}}$ and
\begin{equation*}
\vr{f}_{\mathrm{kin}}(\vr{q},\gvr{\omega}) = \left\lbrack \begin{array}{c}
\mt{M}_{FIK}(\vr{q}) \\
\mt{I}_{3}
\end{array}\right\rbrack \; \gvr{\omega}.
\label{eq:kinincntrlform}
\end{equation*}
Notice that in this model $\gvr{\omega}$ plays the role of the control actions, which are motor velocities in this case.\\ 

If necessary, we could use Eq.~\eqref{eq:cntrleng} for controlling the motions of Otbot. In doing so, we would neglect the system dynamics, but a strong point of this model is its simplicity, which leads to higher frequency controllers. Moreover, the model would be sufficient if the commanded velocities $\gvr{\omega}(t)$ where easy to control (e.g., if the components of $\gvr{\omega}(t)$ were smooth enough for the motors at hand). However, our interest is in obtaining a more complete and accurate model so we can simulate and control our robot as realistically as possible, taking into account its dynamics. We develop such a model in the next section.
\section{Dynamic model}
\label{ch:dyn_model}

We next obtain a dynamic model of Otbot. Since the robot is a non-holonomic system, the model will be derived using Lagrange's equations with multipliers~\cite{Murray:1994}. After obtaining all terms in this equation, we will manipulate it to solve the inverse and forward dynamics problems. The resulting equations will be needed for simulation and control design purposes in Sections \ref{ch:chap4ParamIdent} and \ref{ch:trackingcontrl}. The forward dynamics solution, in particular, yields the dynamic model in control form, \mbox{$\vr{\dot{x}} = \vr{f}(\vr{x}, \vr{u})$}, which is the form typically assumed by many of the algorithms we use.

\subsection{Lagrange's equation with multipliers}
\label{sec:LageqwMulti}

It is well known that, in an inertial reference frame $\mathcal{F}$, the equation of motion of a non-holonomic system can be written as
\begin{equation}
    \mt{M}(\vr{q}) 
\; 
\vr{\ddot{q}} 
+ 
\mt{C}(\vr{q},\vr{\dot{q}}) 
\; 
\vr{\dot{q}} 
+ 
\mt{G}(\vr{q}) 
+ 
\mt{J}(\vr{q}) \trans
\; 
\gvr{\lambda} 
= 
\vr{Q}_a
+
\vr{Q}_f,
\label{eq:eq_lagrange}
\end{equation}
where
\begin{itemize}
    \item $\vr{q}$ is the configuration vector of the robot.
    \item $\mt{M}\left(\vr{q}\right)$ is the mass matrix characterizing the kinetic energy of the system in $\mathcal{F}$.
    \item $\mt{C}\left(\vr{q},\vr{\dot{q}} \right)\;$ is the generalized Coriolis and centrifugal force matrix. 
    \item $\mt{G}(\vr{q}) = \partial U / \partial \vr{q}$, where $U(\vr{q})$ is the potential energy function of the robot.
    \item $\mt{J}(\vr{q})$ is the constraint Jacobian of the robot.
    \item $\gvr{\lambda}$ is a vector of Lagrange multipliers.
    \item $\vr{Q}_a$ is the generalized force of actuation, which models the motor forces.
    \item $\vr{Q}_f$ is the generalized force modelling all friction forces in the system.
\end{itemize}
To particularize this equation for Otbot we let $\mathcal{F}$ be the absolute frame fixed to the ground (Section~\ref{sec:Refframesstatecoords}), which is assumed to be inertial at all effects. Recall that in this frame the robot configuration is given by
\begin{equation*}
\vr{q} = \left( x, y, \alpha, \varphi_r, \varphi_l, \varphi_p  \right).
\label{eq:chap3eq1Nonum}
\end{equation*}
We also define $\vr{u}$ as 
\begin{equation*}
\vr{u} = \left( \tau_r, \tau_l, \tau_p  \right),
\label{eq:chap3eq2Nonum}
\end{equation*}
where $\tau_r$, $\tau_l$, and $\tau_p$ are the torques applied by the motors to the left and right wheels and to the pivot joint, respectively. Note that $\mt{G}(\vr{q}) = \vr{0}$ in our case, as the Otbot moves on flat terrain, and that $\mt{J}(\vr{q})$ is the Jacobian in Eq.~\eqref{eq:chap2JacobianJ}. We next see how to obtain the remaining terms in Eq.~\eqref{eq:eq_lagrange}: 
$$\mt{M}(\vr{q}), \mt{C}(\vr{q},\vr{\dot{q}}), \vr{Q}_a, \vr{Q}_f.$$
\begin{figure}[b!]
	\centering
	\includegraphics[width=0.70\linewidth]{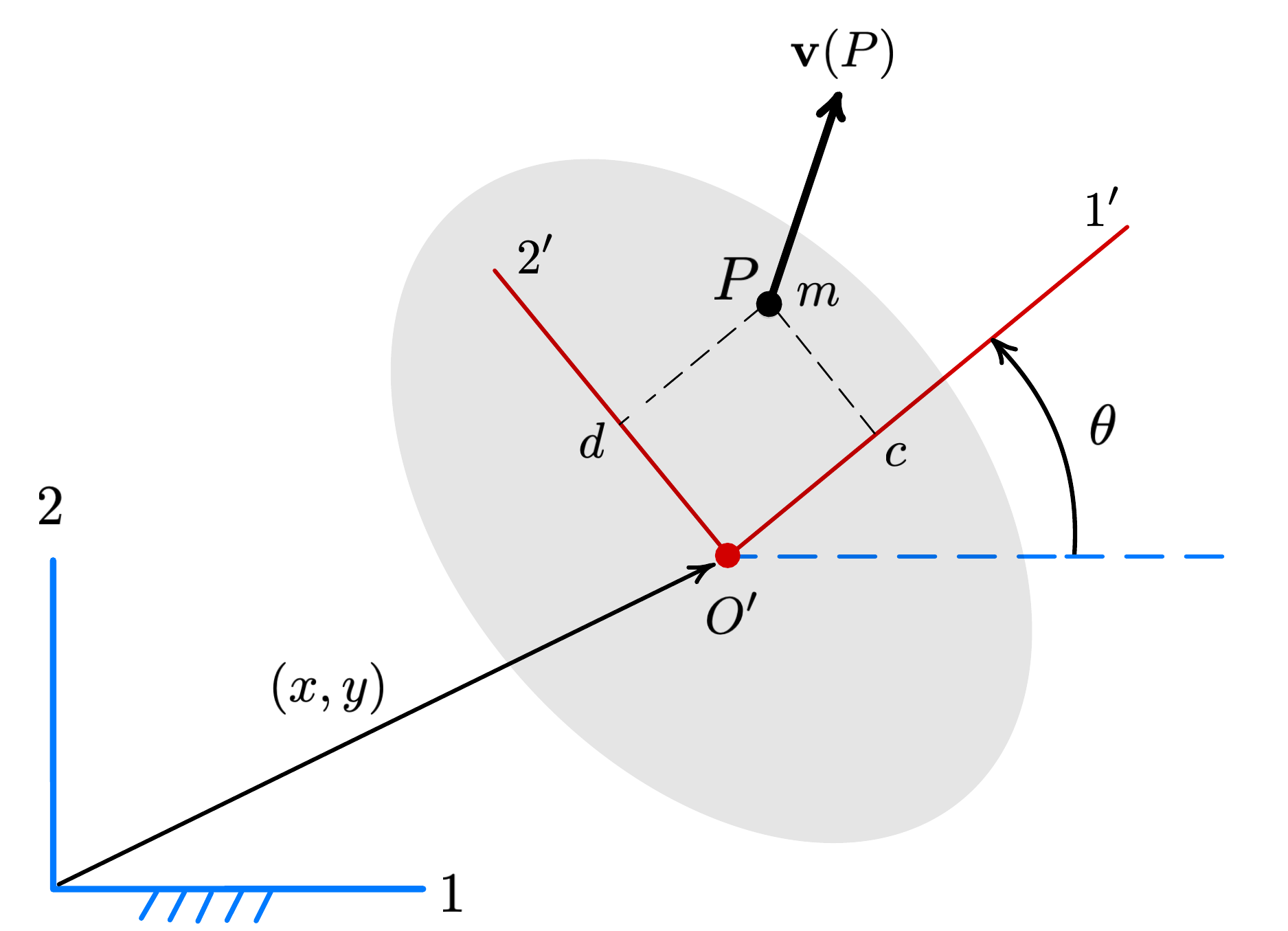}
	\caption{\label{fig:fig_kin_en_trans} Notation used to derive a general expression for the translational kinetic energy of a point of mass $m$ located at a point $P$ of a body moving under planar motion (shown in grey). The velocity of the body is given by $\dot{x}$, $\dot{y}$ and $\dot{\theta}$. The coordinates of $P$ in the body-fixed frame $\{1', 2'\}$ are $(c,d)$.} 
\end{figure}

\subsubsection{Mass matrix}
\label{subsec:subkinenergysystem}

The mass matrix $\mt{M}(\vr{q})$ describes the mass of our entire dynamical system in terms of the configuration variables. This matrix can be computed by writing the total kinetic energy $T$ of the robot as a function of $\vr{q}$ and $\vr{\dot{q}}$, and expressing it in the form
\begin{equation*}
T(\vr{q},\vr{\dot{q}}) = 
\frac{1}{2} \; 
\vr{\dot{q}} \trans
\; 
\mt{M}(\vr{q}) 
\; 
\vr{\dot{q}}.    
\end{equation*}

The robot has four bodies: the main body of the chassis, the left wheel, the right wheel, and the platform. We have to compute the translational and rotational kinetic energies of each body and add them all to obtain the expression of $T$ for the complete system.

To compute the translational kinetic energy, we first derive a general expression for the kinetic energy $T_t$ of a point mass $m$ located at some point $P$ of a body under planar motion (see Fig.~\ref{fig:fig_kin_en_trans}). We wish to express $T_t$ in terms of the velocity coordinates of the body, \mbox{$\vr{w} = (\dot{x}, \dot{y}, \dot{\theta})$}, the coordinates of $P$ in the body-fixed frame $\{O', 1', 2'\}$, the body angle $\theta$, and $m$. To obtain $T_t$, we first express the velocity of $P$ as
\begin{equation*}
\vr{v}(P) = \vr{v}(O') + \gvr{\omega}_{\mathrm{body}} \times \vr{O'P},
\label{eq:chap3eq3Nonum}
\end{equation*}
where $\gvr{\omega}_{\mathrm{body}} = (0,0,\dot{\theta})$. This yields
\begin{equation*}
\left\lbrack 
\begin{array}{c}
v_1 \\
v_2 \\
0
\end{array}
\right\rbrack = 
\left\lbrack 
\begin{array}{c}
\dot{x} \\
\dot{y} \\
0
\end{array}
\right\rbrack
+ 
\left\lbrack 
\begin{array}{c}
0\\
0 \\
\dot{\theta}
\end{array}
\right\rbrack
\times 
\left\lbrack 
\begin{array}{c}
c \cos{\theta}  - d \sin{\theta}\\
c \sin{\theta}  + d \cos{\theta} \\
0
\end{array}
\right\rbrack
= 
\left\lbrack 
\begin{array}{c}
\dot{x} - c \dot{\theta} \sin{\theta}  - d \dot{\theta} \cos{\theta} \\
\dot{y} + c \dot{\theta} \cos{\theta}  - d \dot{\theta} \sin{\theta}  \\
0
\end{array}
\right\rbrack.
\end{equation*}
By adding $\dot{\theta}$ in the third row, we then obtain
\begin{equation*}
\underbrace{\left\lbrack 
\begin{array}{c}
v_1 \\
v_2 \\
\dot{\theta}
\end{array}
\right\rbrack}_{\textstyle \vr{t}} = 
\underbrace{\left\lbrack
\begin{array}{ccc}
1 & 0 & - c \sin{\theta} - d \cos{\theta} \\
0 & 1 & c \cos{\theta} - d \sin{\theta}  \\
0 & 0 & 1
\end{array}
\right\rbrack}_{\textstyle \mt{A}}
\underbrace{
\left\lbrack 
\begin{array}{c}
\dot{x} \\
\dot{y} \\
\dot{\theta}
\end{array}
\right\rbrack.}_{\textstyle \vr{w}}
\end{equation*}
Now note that $T_t$ can be expressed as follows in terms of $\vr{t}$:
\begin{equation*}
T_t = \frac{1}{2} m (v_1^2 + v_2^2) = \frac{1}{2} 
\underbrace{\left[
v_1 \; v_2 \; \dot{\theta}
\right]}_{\vr{t}\trans}
\underbrace{\left\lbrack
\begin{array}{ccc}
m & 0 & 0 \\
0 & m & 0 \\
0 & 0 & 0
\end{array}
\right\rbrack}_{\textstyle \mt{M}'}
\underbrace{\left\lbrack
\begin{array}{c}
v_1 \\
v_2 \\
\dot{\theta}
\end{array}
\right\rbrack}_{\vr{t}}.
\end{equation*}
Using $\vr{t} = \mt{A} \; \vr{w}$, we can now express $T_t$ as
\begin{equation*}
T_t= \frac{1}{2} \; \vr{w}\trans \underbrace{\mt{A}\trans \mt{M}' \mt{A}}_{\mt{M}_t}  \vr{w} 
= \frac{1}{2} \; \vr{w}\trans \; \mt{M}_t \; \vr{w},   
\end{equation*}
where
\begin{align*}
&\mt{M}_t(c,d,m)  = \mt{A}\trans \mt{M}' \; \mt{A} = \\
&= 
\left\lbrack
\begin{array}{ccc}
m & 0 & - c \; m \sin{\theta} - d \; m \cos{\theta}\\
0 & m & c \; m \cos{\theta} - d \; m \sin{\theta}  \\
- c \; m \sin{\theta} - d \; m \cos{\theta} & c \; m \cos{\theta} - d \; m \sin{\theta} & m(c^2 + d^2)
\end{array}
\right\rbrack.
\end{align*}
Using the relations above, we thus can express the translational kinetic energy of a point mass $m$ located at some point $P$ of a body under planar motion as:
\begin{equation*}
T_t(c, d, m, \dot{x}, \dot{y}, \theta, \dot{\theta}) = 
\frac{1}{2} \; 
\vr{w}\trans
\; 
\mt{M}_t(c,d,m) 
\; 
\vr{w}.    
\end{equation*}
Using this formula and the dynamic parameters defined in Table~\ref{tab:tableotbotsymbols}, it is easy to see that the translational kinetic energy of the rolling chassis (including the wheels) and the platform are given by
\begin{align*}
T_{\mathrm{trans,chass}} &= T_t(x_B, y_B, m_c, \dot{x}, \dot{y}, \theta, \dot{\theta})\\
T_{\mathrm{trans,plat}}  &= T_t(x_F, y_F, m_p, \dot{x}, \dot{y}, \alpha, \dot{\alpha})
\end{align*}
Again with the same parameters, the rotational kinetic energies of the chassis, the right and left wheels, and the platform can be expressed as
\begin{align*}
T_{\mathrm{rot,chass}} &= \frac{1}{2} \cdot I_c \cdot \dot{\theta}^2,\\
T_{\mathrm{rot,r}} &= \frac{1}{2} \cdot I_a \cdot \dot{\varphi}_r^2,\\
T_{\mathrm{rot,l}} &= \frac{1}{2} \cdot I_a \cdot \dot{\varphi}_l^2,\\
T_{\mathrm{rot,plat}} &= \frac{1}{2} \cdot I_p \cdot \dot{\alpha}^2,
\end{align*}
where we emphasize that $I_c$ is the vertical moment of inertia of the rolling chassis about its center of mass $B$ (i.e., assuming that the chassis body and the wheels form a single rigid body). The total kinetic energy of the robot is thus given by
\begin{equation*}
T(\vr{q}, \vr{\dot{q}}) = T_{\mathrm{trans,chass}} + T_{\mathrm{trans,plat}} + T_{\mathrm{rot,chass}} + T_{\mathrm{rot,r}} + T_{\mathrm{rot,l}} + T_{\mathrm{rot,plat}},
\end{equation*}
Since $T(\vr{q}, \vr{\dot{q}})$ is a quadratic form, the Hessian of $T$ will provide the desired mass matrix $\mt{M}(\vr{q})$.
\begin{table}[t!]

\begin{center}

\includegraphics[width=0.7\textwidth]{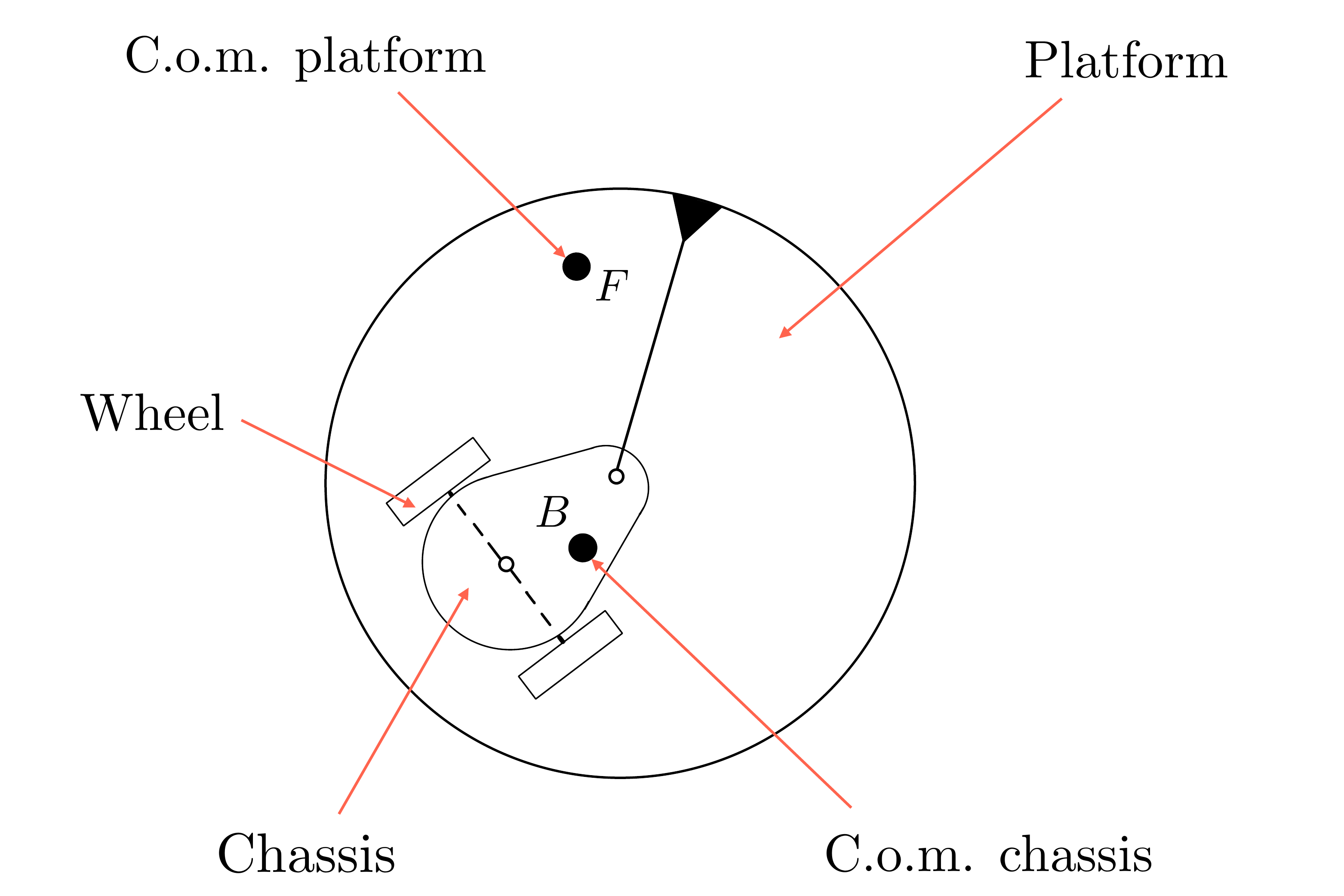}

\vspace{5mm}
\small
\begin{tabular}{ll}
\toprule
\textbf{Symbol} & \textbf{Meaning} \\
\toprule
$(x_B, y_B)$ & Coordinates of the center of mass of the chassis in the chassis frame 
\\ 
$(x_F, y_F)$ & Coordinates of the center of mass of the platform in the platform frame 
\\ 
$m_c$ & Total mass of the rolling chassis
\\
$m_p$ & Mass of the platform
\\
$I_c$ & Vertical moment of inertia of the rolling chassis at point $B$
\\ 
$I_p$ & Vertical moment of inertia of the platform at point $F$
\\ 
$I_a$ &  Axial moment of inertia of a wheel
\\ 
\bottomrule
\end{tabular}
\caption{Robot parameters and their meaning.}
\label{tab:tableotbotsymbols}
\end{center}
\end{table}
With the help of a symbolic manipulation tool we obtain the expression shown in Fig.~\ref{fig:massmatrixexpr}. 
\begin{figure}[t!]
\centering
$\displaystyle
\left\lbrack 
\begin{array}{cccccc}
m_c + m_p  
& 
0 
& 
\begin{array}{c}
- m_p \, y_F \, c_{\alpha} \\
- m_p \, x_F \, s_{\alpha} \\
- m_c \, y_B \, c_{\theta} \\ 
- m_c \, x_B \, s_{\theta}
\end{array}

& 
0 
& 
0 
& 
\begin{array}{cc}
m_c \, y_B \, c_{\theta} \\
+ m_c \, x_B \, s_{\theta} 
\end{array}
\\ [12mm]
0 
& 
m_c + m_p 
& 
\begin{array}{c}
m_p \,x_F \,c_{\alpha} \\
- m_p \, y_F \, s_{\alpha} \\
+ m_c \, x_B \, c_{\theta} \\
- m_c \, y_B \, s_{\theta} 
\end{array}
& 
0 
& 
0 
& 
\begin{array}{c}
m_c \, y_B \, s_{\theta} \\
- m_c \, x_B \, c_{\theta}
\end{array}
\\ [12mm]
\begin{array}{c}
-m_p \, y_F \, c_{\alpha} \\
- m_p \, x_F \, s_{\alpha} \\
- m_c \, y_B \, c_{\theta} \\
- m_c \, x_B \, s_{\theta} 
\end{array}
& 
\begin{array}{c}
m_p \, x_F \, c_{\alpha} \\
- m_p \, y_F \, s_{\alpha} \\
+ m_c \, x_B \, c_{\theta} \\
- m_c \, y_B \, s_{\theta}
\end{array}
& 
\begin{array}{c}
m_c \, {x_B}^2 + m_p \, {x_F}^2 \\
+ m_c \, {y_B}^2 +m_p \, {y_F}^2 \\
+ I_c + I_p  
\end{array}
& 
0 
& 
0
& 
\begin{array}{c}
-m_c \, {x_B}^2 \\
- m_c \, {y_B}^2 \\
- I_c 
\end{array}
\\ [12mm]
0 & 0 & 0 & I_a & 0 & 0
\\ [12mm]
0 & 0 & 0 & 0 & I_a & 0
\\ [12mm]
\begin{array}{c}
m_c \, y_B \, c_{\theta} \\
+ m_c \, x_B \, s_{\theta} 
\end{array}
& 
\begin{array}{c}
m_c \, y_B \, s_{\theta} \\
- m_c \, x_B \, c_{\theta} 
\end{array}
& 
\begin{array}{c}
-m_c \,{x_B }^2 \\
-m_c \,{y_B }^2 \\
-I_c  
\end{array}
& 
0 
& 
0 
& 
\begin{array}{c}
m_c \, {x_B}^2 \\
+ m_c \, {y_B}^2 \\
+ I_c 
\end{array}
\end{array}
\right\rbrack
$
\caption{\label{fig:massmatrixexpr} Symbolic expression of the mass matrix $\mt{M}(\vr{q})$}
\end{figure}

\subsubsection{Coriolis matrix}
\label{subsec:cormatrix}

To compute the Coriolis matrix $\mt{C}(\vr{q}, \vr{\dot{q}})$ we have used the expression
\begin{equation}
    {\mt{C}}_{\mathrm{ij}} =\frac{1}{2}\sum_{\mathrm{k}=1}^{\mathrm{n}} \left(\frac{\partial {\mt{M}}_{\mathrm{ij}} }{\partial {q}_{\mathrm{k}} }+\frac{\partial {\mt{M}}_{\mathrm{ik}} }{\partial {q}_{\mathrm{j}} }-\frac{\partial {\mt{M}}_{\mathrm{kj}} }{\partial {q}_{\mathrm{i}} }\right)\vr{\dot{q}}_{\mathrm{k}} 
    \label{eq:eq_coriolis}
\end{equation} 
given in~\cite{Murray:1994}, where $\mt{M}_{ij}$ and $\mt{C}_{ij}$ refer to the $(i,j)$ entry of $\mt{C}(\vr{q}, \vr{\dot{q}})$ and $\mt{M}(\vr{q})$ respectively. This yields the expression for $\mt{C}(\vr{q}, \vr{\dot{q}})$ shown in Fig.~\ref{fig:cormatrixexpr}.
\begin{figure}[t!]
\centering
$\displaystyle
\left\lbrack 
\begin{array}{cccccc}
0 
& 
0 
& 
\begin{array}{c}
-\dot{\theta} \, m_c (x_B \, c_\theta - y_B \, s_\theta) \\
-\dot{\alpha} \, m_p (x_F \, c_\alpha - y_F \, s_\alpha) 
\end{array}
& 
0 
& 
0 
& 
\dot{\theta} \, m_c  \, {(x_B \, c_\theta-y_B \, s_\theta)}
\\ [7mm]
0 
& 
0 
& 
\begin{array}{c}
-\dot{\theta} \, m_c (y_B \, c_\theta + x_B \, s_\theta) \\
-\dot{\alpha} \, m_p (y_F \, c_\alpha + x_F \, s_\alpha) 
\end{array}
& 
0 
& 
0 
& 
\dot{\theta} \, m_c \, (y_B \, c_\theta+x_B \, s_\theta)
\\ [7mm]
0 & 0 & 0 & 0 & 0 & 0
\\ [7mm]
0 & 0 & 0 & 0 & 0 & 0
\\ [7mm]
0 & 0 & 0 & 0 & 0 & 0
\\ [7mm]
0 & 0 & 0 & 0 & 0 & 0
\end{array}
\right\rbrack
$
\caption{\label{fig:cormatrixexpr} Symbolic expression of the Coriolis matrix $\mt{C}(\vr{q},\vr{\dot{q}})$}
\end{figure}

\subsubsection{Generalized force of actuation}
\label{subsec:GenFofaction}

To write the expression for the generalized actuation force $\vr{Q}_a$ we will use the method of virtual power. This requires calculating the power $P_a$ generated by the actuation forces $\vr{u} = (\tau_r, \tau_l, \tau_p)$ under a virtual velocity $\vr{\dot{q}}$ at which the mechanism is moving, and writing it in the form
\begin{equation}
P_a = \vr{Q}_a\trans \; \vr{\dot{q}}.
\label{eq:ch3Powereq1}
\end{equation}
Notice that, clearly,
\begin{equation*}
P_a = \underbrace{\left[
\begin{array}{ccc}
\tau_r & \tau_l & \tau_p
\end{array}
\right]}_{\vr{u}\trans} \; \underbrace{\left[
\begin{array}{c}
\dot{\varphi}_r\\
\dot{\varphi}_l\\
\dot{\varphi}_p
\end{array}
\right]}_{\gvr{\dot{\varphi}}},
\label{eq:Powereq2}
\end{equation*}
which we can rewrite in terms of $\vr{\dot{q}}$ as follows
\begin{equation*}
P_a = \underbrace{\left[
\begin{array}{cc}
\mt{0} & \vr{u}\trans
\end{array}
\right]}_{\vr{Q}_a\trans} \; \underbrace{\left[
\begin{array}{c}
\vr{w}\\
\gvr{\dot{\varphi}}
\end{array}
\right]}_{\vr{\dot{q}}}.
\label{eq:Powereq3withQa}
\end{equation*}
By identifying the vectors of this equation with those of Eq.~\eqref{eq:ch3Powereq1} it is clear that
\begin{equation*}
\vr{Q}_a = (0, 0, 0, \tau_r, \tau_l, \tau_p).
\label{eq:Qaexpression}
\end{equation*}
For later developments, note also that \mbox{$\vr{Q}_a = \mt{E} \cdot \vr{u}$}, where
\begin{equation*}
\mathbf{E} =\left\lbrack \begin{array}{ccc}
0 & 0 & 0\\
0 & 0 & 0\\
0 & 0 & 0\\
1 & 0 & 0\\
0 & 1 & 0\\
0 & 0 & 1
\end{array}\right\rbrack.
\end{equation*}

\subsubsection{Generalized forces of friction}
\label{subsec:GenFoffriction}

As explained in Section~\ref{subsec:dynamicmodelintro}, we assume that all friction forces can be neglected except those due to viscous friction at the motor shafts. Therefore, the generalized force of friction will be
\begin{equation*}
\vr{Q}_f = - (0, 0, 0, b_w \cdot \dot{\varphi}_r, b_w \cdot \dot{\varphi}_l, b_p \cdot \dot{\varphi}_p),
\label{eq:viscfriction2}
\end{equation*} 
where $b_w$, and $b_p$ are the viscous friction coefficients for the right and left wheels, and for the pivot joint, respectively. For later use, we shall express $\vr{Q}_f$ as
\begin{equation*}
\vr{Q}_f = \mt{E}_f \cdot \vr{\dot{q}}
\label{eq:viscfriction2nonumber}
\end{equation*} 
where
\begin{equation*}
\mt{E}_f =
\left[
\begin{array}{cccccc}
0 & 0 & 0 & 0 & 0 & 0 \\
0 & 0 & 0 & 0 & 0 & 0 \\
0 & 0 & 0 & 0 & 0 & 0 \\
0 & 0 & 0 & -b_w & 0 & 0 \\
0 & 0 & 0 & 0 & -b_w & 0 \\
0 & 0 & 0 & 0 & 0 & -b_p \\
\end{array}
\right].
\label{eq:viscfrictionmatrix}
\end{equation*}   

\subsection{Conventional dynamics solutions}
\label{sec:convdynmethods}

We now wish to solve the inverse and forward dynamics problems, and obtain the dynamic model in control form $\vr{\dot{x}}=\vr{f}(\vr{x}, \vr{u})$. To do so, we will be using Lagrange's equation of motion cited in Eq.~\eqref{eq:eq_lagrange} and we will augment it with an acceleration constraint when needed. Solving for the robot model using this conventional method yields an equation for the system dynamics that requires the inversion of a $9\times9$ matrix. In Section~\ref{sec:CancelLagrange} we will provide an alternative method that results in more compact expressions for the model and the inverse dynamics.

\subsubsection{Inverse dynamics}
\label{subsec:InvDyn}

We now wish to compute the inverse dynamics of our system, which consists in finding the torques $\vr{u} = (\tau_r, \tau_l, \tau_p)$ that correspond to a given $\vr{\ddot{q}}$. These torques are obtained by solving
\begin{equation}
\mt{M} \; \vr{\ddot{q}}
+
\mt{C} \; \vr{\dot{q}} 
+ 
{\mt{J}}\trans 
\;
\gvr{\lambda} 
=
\mt{E} \; \vr{u} + \mt{E}_f \; \vr{\dot{q}}
\label{eq:eq_simple_lagrange} 
\end{equation}
for $\vr{u}$ and $\gvr{\lambda}$ (6 equations and 6 unknowns). This is the Lagragian equation adapted to our system, with the dependencies on $\vr{q}$ and $\vr{\dot{q}}$ omitted for simplicity, and with $\vr{Q}_a$ and $\vr{Q}_f$ replaced by $\mt{E} \, \vr{u}$ and $\mt{E}_f \, \vr{\dot{q}}$. To solve the problem we define 
$$\gvr{\tau}_{\mathrm{\scaleto{ID}{4pt}}} = \mt{M} \; \vr{\ddot{q}} + \left( \mt{C} - \mt{E}_f \right) \; \vr{\dot{q}}$$ 
and rewrite Eq.~\eqref{eq:eq_simple_lagrange} as
\begin{equation*}
\mt{E} \; \vr{u}
-
{\mt{J}}\trans 
\;
\gvr{\lambda} 
=
\gvr{\tau}_{\mathrm{\scaleto{ID}{4pt}}}, 
\end{equation*}
or equivalently, as
\begin{equation*}
\left\lbrack 
\begin{array}{cc}
\mt{E}  & -\mt{J}\trans
\end{array}
\right\rbrack 
\;
\left\lbrack 
\begin{array}{c}
\vr{u} \\
\gvr{\lambda}
\end{array}
\right\rbrack
=
\gvr{\tau}_{\mathrm{\scaleto{ID}{4pt}}}.
\end{equation*}

It is easy to see that $[\mt{E} \; -\mt{J}\trans]$ is a full rank matrix irrespective of $\vr{q}$. This follows directly from the expressions of $\mt{E}$ and $\mt{J}$ in the Otbot. Thus, we can write
\begin{equation*}
\left\lbrack 
\begin{array}{c}
\vr{u} \\
\gvr{\lambda}
\end{array}
\right\rbrack
=
{
\left\lbrack 
\begin{array}{cc}
\mt{E}  & -\mt{J}\trans
\end{array}
\right\rbrack
}^{-1}
\;
\gvr{\tau}_{\mathrm{\scaleto{ID}{4pt}}},
\end{equation*}
which gives the solution to the inverse dynamics:
\begin{equation}
\vr{u}
=
\left\lbrack 
\begin{array}{cc}
\mt{I}_{\mathrm{3}}  & \vr{0}
\end{array}
\right\rbrack
\;
{
\left\lbrack 
\begin{array}{cc}
\mt{E}  & -\mt{J}\trans
\end{array}
\right\rbrack
}^{-1}
\;
\gvr{\tau}_{\mathrm{\scaleto{ID}{4pt}}}.
\label{eq:eq_inverse_dyn}    
\end{equation}

\subsubsection{Forward dynamics}
\label{subsec:FwdDyn}

The forward dynamics problem consists in finding the acceleration $\vr{\ddot{q}}$ that corresponds to a given action $\vr{u}$. Similarly to the inverse problem, we need to solve Eq.~\eqref{eq:eq_simple_lagrange}, but this time for $\vr{\ddot{q}}$, which gives us a system of 6 equations in 9 unknowns (6 coordinates in $\vr{\ddot{q}}$ and 3 in $\gvr{\lambda}$). Thus, we need three more equations, which we can obtain by taking the time derivative of $\mt{J} \; \vr{\dot{q}} = \vr{0}$, which gives 
\begin{equation*}
\mt{J}
\;
\vr{\ddot{q}} 
+
\mt{\dot{J}} 
\;
\vr{\dot{q}} = \vr{0},    
\end{equation*}
or equivalently
\begin{equation*}
\mt{J}
\;
\vr{\ddot{q}} 
=
-
\mt{\dot{J}} 
\;
\vr{\dot{q}}.
\label{eq:eq_acc_cstr}    
\end{equation*}
The latter equation is called the acceleration constraint of the robot. By combining it with Eq.~\eqref{eq:eq_simple_lagrange}, we obtain a solvable linear system of 9 equations with 9 unknowns with the form:
\begin{equation*}
\left\lbrack
\begin{array}{cc}
\mt{M}     &  {\mt{J}}\trans \\
\mt{J} & \vr{0}
\end{array}\right\rbrack 
\left\lbrack
\begin{array}{c}
\vr{\ddot{q}} \\
\gvr{\lambda} 
\end{array}\right\rbrack 
=
\left\lbrack 
\begin{array}{c}
\mt{E} \; \vr{u} + \left( \mt{E}_f - \mt{C} \right) \; \vr{\dot{q}} \\
-\mt{\dot{J}} \; \vr{\dot{q}} 
\end{array}
\right\rbrack.
\end{equation*}

Since $\mt{M}$ is positive-definite and $\mt{J}$ is full row rank, the matrix on the left-hand side can be inverted to write
\begin{equation*}
\left\lbrack
\begin{array}{c}
\vr{\ddot{q}} \\
\gvr{\lambda} 
\end{array}\right\rbrack 
=
{
\left\lbrack 
\begin{array}{cc}
\mt{M}       &      \mt{J}\trans \\
\mt{J} & \vr{0}
\end{array}
\right\rbrack
}^{-1} 
\left\lbrack 
\begin{array}{c}
\mt{E} \; \vr{u} + \left( \mt{E}_f - \mt{C} \right) \; \vr{\dot{q}} \\
-\mt{\dot{J}} \; \vr{\dot{q}} 
\end{array}
\right\rbrack,
\end{equation*}
which gives us the solution for the forward dynamics:
\begin{equation}
\vr{\ddot{q}} 
=
\left\lbrack 
\begin{array}{cc}
\mt{I}_{\mathrm{6}}  & \vr{0}
\end{array}
\right\rbrack 
{
\left\lbrack 
\begin{array}{cc}
\mt{M}       &      \mt{J}\trans \\
\mt{J} & \vr{0}
\end{array}
\right\rbrack
}^{-1} 
\left\lbrack 
\begin{array}{c}
\mt{E} \; \vr{u} + \left( \mt{E}_f - \mt{C} \right) \; \vr{\dot{q}} \\
-\mt{\dot{J}} \; \vr{\dot{q}} 
\end{array}
\right\rbrack.
\label{eq:eq_forward_dyn1}    
\end{equation}

\subsection{Cancelling the multipliers}
\label{sec:CancelLagrange}

Eqs.~\eqref{eq:eq_inverse_dyn} and~\eqref{eq:eq_forward_dyn1} require the inversion of a $6 \times 6$ and a $9 \times 9$ matrix respectively, which may be costly operations when, as in this work, we have to repeat them many times. We next see that, by exploiting two parameterizations of the feasible velocities $\vr{\dot{q}}$, we can find simpler solutions that require, in each case, no matrix inverse, and the inverse of a $6 \times 6$ matrix only. These parameterizations are
\begin{equation}
\left[\begin{array}{l}
\vr{\dot{p}} \\
\gvr{\dot{\varphi}}
\end{array}\right]=\underbrace{\left[\begin{array}{c}
\mt{I}_{\mathrm{3}} \\
\mt{M}_{IIK}
\end{array}\right]}_{\mt{\Lambda}} \vr{\dot{p}} \quad \rightarrow \quad \vr{\dot{q}} = \mt{\Lambda} \cdot \vr{\dot{p}},
\label{eq:eq_lambda}
\end{equation}
\begin{equation}
\left[\begin{array}{l}
\vr{\dot{p}} \\
\gvr{\dot{\varphi}}
\end{array}\right]=\underbrace{\left[\begin{array}{c}
\mt{M}_{FIK} \\
\mt{I}_{\mathrm{3}}
\end{array}\right]}_{\mt{\Delta}} \gvr{\dot{\varphi}} \quad \rightarrow \quad \vr{\dot{q}} = \mt{\Delta} \cdot \gvr{\dot{\varphi}},
\label{eq:eq_delta}
\end{equation}
where $\mt{M}_{FIK}$ and $\mt{M}_{IIK}$ are the matrices used to solve the forward and inverse instantaneous kinematic problems (Section~\ref{sec:IKP}).

\subsubsection{Inverse dynamics}
\label{subsec:InvDynMultFree}

Again, we wish to solve Eq.~\eqref{eq:eq_inverse_dyn} for $\vr{u}$. To do this, we can multiply it by $\mt{\Delta}\trans$ to obtain
\begin{equation}
\mt{\Delta}\trans \mt{M} \; \vr{\ddot{q}} + \mt{\Delta}\trans \mt{C} \; \vr{\dot{q}} + \mt{\Delta}\trans \mt{J}\trans \gvr{\lambda} =\mt{\Delta}\trans \mt{E} \; \vr{u} + \mt{\Delta}\trans \mt{E}_f \; \vr{\dot{q}},
\label{eq:lagrange_delta}
\end{equation}
which introduces two simplifications. On the one hand, it is not difficult to see that
\begin{equation}
\mt{\Delta}\trans \mt{J}\trans = \vr{0}.
\label{eq:eq_deltaj=0}
\end{equation}
This can be explained using the constraint in Eq.~\eqref{eq:eq_jq=0}, which is satisfied only if $\vr{\dot{q}}$ is admissible. Since Eq.~\eqref{eq:eq_delta} also provides admissible $\vr{\dot{q}}$ for any $\gvr{\dot{\varphi}}$, we have 
\begin{equation*}
    \mt{J} \, \vr{\dot{q}} = \mt{J} \, \mt{\Delta} \, \gvr{\dot{\varphi}} = \vr{0}.
\end{equation*}
As the earlier result must hold for any $\gvr{\dot{\varphi}}$, it must be $\mt{J} \, \mt{\Delta}  = \vr{0}$ and by taking the transpose we find that Eq.~\eqref{eq:eq_deltaj=0} must be true. On the other hand, we can directly simplify the right-hand side of Eq.~\eqref{eq:lagrange_delta} since we have
\begin{equation}
\mt{\Delta}\trans \mt{E} = \left[\begin{array}{cc}
\mt{M}_{FIK} \; \bm{I}_{\mathrm{3}}
\end{array}\right]\left[\begin{array}{c}
\vr{0} \\
\mt{I}_{\mathrm{3}}
\end{array}\right] =\mt{I}_{\mathrm{3}}.
\label{eq:eq_deltae=0}
\end{equation}

Therefore, by substituting Eqs.~\eqref{eq:eq_deltaj=0} and~\eqref{eq:eq_deltae=0} in Eq.~\ref{eq:lagrange_delta} we obtain a closed-form expression for the inverse dynamics in which no matrix inverse is involved: 
\begin{equation}
\vr{u}=\mt{\Delta}\trans \mt{M} \; \vr{\ddot{q}} + \mt{\Delta}\trans \left( \mt{C} - \mt{E}_f \right) \; \vr{\dot{q}}.
\label{eq:eq_simple_inverse}
\end{equation}

\subsubsection{Forward dynamics}
\label{subsec:FwdDynMultFree}

We now wish to find a simplified version of the forward dynamics. To do this we will first compute the dynamic model in task-space coordinates, which provides the time evolution of the $\vr{p}$ state variables. 
This can be done by using Eq.~\eqref{eq:eq_lambda} and its derivative
\begin{equation}
\vr{\ddot{q}}=\mt{\Lambda} \; \vr{\ddot{p}} + \mt{\dot{\Lambda}} \; \vr{\dot{p}}.
\label{eq:eq_lambda_deriv}
\end{equation}
By substituting them into Eq.~\eqref{eq:eq_simple_inverse}, we find that
\begin{align*}
\mt{\Delta}\trans \mt{M}( \mt{\Lambda} \; \vr{\ddot{p}}+\mt{\dot{\Lambda}} \; \vr{\dot{p}}) + \mt{\Delta}\trans (\mt{C} - \mt{E}_f) \mt{\Lambda} \; \vr{\dot{p}} &= \vr{u},\nonumber\\
    \mt{\Delta}\trans \mt{M} \mt{\Lambda} \; \vr{\ddot{p}} + \mt{\Delta}\trans \mt{M}\mt{\dot{\Lambda}} \; \vr{\dot{p}} + \mt{\Delta}\trans (\mt{C} - \mt{E}_f) \mt{\Lambda} \; \vr{\dot{p}} &= \vr{u},\nonumber
\end{align*}
which finally gives us
\begin{equation*}
\underbrace{\mt{\Delta}\trans \mt{M} \mt{\Lambda}}_{\textstyle \bar{\mt{M}}} \vr{\ddot{p}} + \underbrace{\mt{\Delta}\trans \left[ \mt{M}\mt{\dot{\Lambda}} + (\mt{C} - \mt{E}_f) \mt{\Lambda} \right]}_{\textstyle \bar{\mt{C}}}\vr{\dot{p}} =\vr{u},
\end{equation*}
where $\bar{\mt{M}}$ is non-singular because $\mt{\Delta}\trans$, $\mt{M}$ and $\mt{\Lambda}$ are all full rank. We now have an expression for the dynamic model in task-space coordinates:
\begin{equation}
    \bar{\mt{M}} \, \vr{\ddot{p}} + \bar{\mt{C}} \, \vr{\dot{p}} = \vr{u}.
    \label{eq:eq_dyn_task_space}
\end{equation}

The earlier expression does not allow us to solve for $\vr{\ddot{q}}$ yet, but note that Eq.~\eqref{eq:eq_lambda_deriv} can be expanded into:
\begin{align*}
    \left[\begin{array}{c}
\vr{\ddot{p}} \\
\ddot{\gvr{\varphi}}
\end{array}\right] &= \left[\begin{array}{c}
\mt{I}_{\mathrm{3}} \\
\mt{M}_{IIK}
\end{array}\right] \vr{\ddot{p}}+\left[\begin{array}{c}
\vr{0} \\
\mt{\dot{M}}_{IIK}
\end{array}\right] \vr{\dot{p}},
\end{align*}
whose second row gives us an expression of $\ddot{\gvr{\varphi}}$ in terms of $\vr{\ddot{p}}$ and $\vr{\dot{p}}$. By combining this expression with Eq.~\eqref{eq:eq_dyn_task_space} and rearranging the terms we obtain the system:
\begin{equation*}
\left\lbrace 
\begin{array}{l}
\bar{\mt{M}} \; \vr{\ddot{p}} =\vr{u} - \bar{\mt{C}} \; \vr{\dot{p}}
\\
-\mt{M}_{IIK} \; \vr{\ddot{p}} + \ddot{\gvr{\varphi}} = \mt{\dot{M}}_{IIK} \; \vr{\dot{p}}
\end{array}
\right.
\label{eq:eq_dyn_sys1}
\end{equation*}
or, in matrix block form,
\begin{equation*}
\underbrace{\left[\begin{array}{cc}
\bar{\mt{M}} & \vr{0} \\
-\mt{M}_{IIK} & \mt{I}_{\mathrm{3}}
\end{array}\right]}_{\textstyle \mt{K}(\vr{q})}\underbrace{\left[\begin{array}{l}
\vr{\ddot{p}} \\
\ddot{\gvr{\varphi}}
\end{array}\right]}_{\textstyle \vr{\ddot{q}}} = \underbrace{\left[\begin{array}{c}
\vr{u}-\bar{\mt{C}} \; \vr{\dot{p}} \\
\mt{\dot{M}}_{IIK} \; \vr{\dot{p}}
\end{array}\right]}_{\textstyle \vr{b}_{\scaleto{FD}{4pt}}(\vr{q},\vr{\dot{q}},\vr{u})},
\end{equation*}
which easily gives us the forward dynamics by inverting of the $6 \times 6$ matrix on the left-hand side:
\begin{equation}
\vr{\ddot{q}} = \mt{K}^{-1} \; \vr{b}_{\scaleto{FD}{4pt}}.
\label{eq:eq_forward_dyn2}
\end{equation}

It is now simple to write the dynamic model which expresses the derivative of the state variables $\vr{\dot{x}}$ as a function of the state variables $\vr{x}$ and the control action $\vr{u}$. To do this, we only need to augment Eq.~\eqref{eq:eq_forward_dyn2} with  the trivial equation $\vr{\dot{q}} = \vr{\dot{q}}$ to obtain
\begin{equation}
\left\lbrack
\begin{array}{c}
\vr{\dot{q}} \\
\vr{\ddot{q}} 
\end{array}
\right\rbrack
=
\left\lbrack
\begin{array}{c}
\vr{\dot{q}} \\
\mt{K}^{-1} \; \vr{b}_{\scaleto{FD}{4pt}}
\end{array}\right],
\label{eq:eq_dyn_model}
\end{equation}
which yields the robot model in the usual form 
\begin{equation*}
\vr{\dot{x}} = \vr{f}(\vr{x},\vr{u}).    
\end{equation*}
\section{Parameter identification}
\label{ch:chap4ParamIdent}

We next develop a method to identify Otbot's dynamical parameters. To obtain it we resort to nonlinear grey-box model identification, which searches for those parameters that yield the best fit between sensor signals recorded by the robot, and the values of such signals as predicted by the model. This will entail finding the minimum of a nonlinear function of the sought parameters. We start with a short introduction to system identification by prediction-error minimization, then explain the identification process we propose, and finally describe a number of tests we have done to validate our approach in simulation. Our results show that the approach is robust to the presence of noise in the signals, and that it converges fairly well in a sufficiently-large region around the actual parameter values.

\subsection{Prediction error minimization}
\label{sec:prederrormin}

Suppose that the ordinary differential equation that describes the dynamic model of our robot is given by
\begin{equation}
\vr{\dot{x}} = \vr{F}(\vr{x},\vr{u}, \mathbcal{p}),
\label{eq:chap4eq1}
\end{equation}
where $\vr{x}$ is the robot state, $\vr{u}$ are the control actions, and $\mathbcal{p}$ is the vector of system parameters. Typically, we do not have sensors to measure each one of the variables in $\vr{x}$. Instead, the robot sensors provide a vector $\vr{y}$ of outputs whose functional dependence on $\vr{x}$, $\vr{u}$, and $\mathbcal{p}$,
\begin{equation}
\vr{y} = \vr{G}(\vr{x},\vr{u}, \mathbcal{p}),
\label{eq:chap4eq2}
\end{equation}
is known in advance. In this situation we can estimate the value of $\mathbcal{p}$ as follows. Assuming the initial state \mbox{$\vr{x}_0 = \vr{x}(0)$} is known, we command the robot under some control function $\vr{u}(t)$ for $t = 0$ to $t_f$, while recording the values of $\vr{u}(t)$ and $\vr{y}(t)$ simultaneously. Since $\vr{u}(t)$ and $\vr{y}(t)$ are usually managed with digital means, they have the form of sampled-data sequences
\begin{equation*}
\begin{array}{l}
\vr{u}_0, \vr{u}_1, \ldots, \vr{u}_N,\\
\vr{y}_0, \vr{y}_1, \ldots, \vr{y}_N,
\end{array}
\label{eq:chap4eqnonumber1}
\end{equation*}
whose values correspond to the time instants \mbox{$t_0 < t_1 < \ldots < t_N$}. We can then construct the fitting function
\begin{equation}
\varepsilon(\mathbcal{p}) = \sum_{k=0}^{N} \left\Vert \vr{y}_k - \vr{y}_k^{\mathrm{pred}}  \right\Vert^{2},
\label{eq:chap4eq3}
\end{equation}
where
\begin{equation}
\vr{y}_{k}^{\mathrm{pred}} = \vr{G}(\vr{x}_k,\vr{u}_k, \mathbcal{p}),
\label{eq:chap4eq3prime}
\end{equation}
and try to find the values of $\mathbcal{p}$ that minimize $\varepsilon(\mathbcal{p})$. At first glance, it looks like the $\vr{x}_k$ values are unknown, but we can obtain them by solving a sequence of initial value problems with an appropriate numerical integration method to express $\vr{x}_k$ as a function of $\vr{x}_0$, \mbox{$\vr{u}_0, \ldots, \vr{u}_{k-1}$}, and $\mathbcal{p}$. Thus, $\varepsilon(\mathbcal{p})$ depends ultimately on $\mathbcal{p}$, as $\vr{x}_0$, \mbox{$\vr{u}_0, \ldots, \vr{u}_{N-1}$}, \mbox{$\vr{y}_0, \ldots, \vr{y}_N$} are all known a priori. In the literature, $\varepsilon(\mathbcal{p})$ is called a prediction error (or loss) function~\cite{Ljung:1999} as it provides a measure of how well the parameters $\mathbcal{p}$ explain the evolution of the outputs under the model assumed.

To minimize $\varepsilon(\mathbcal{p})$ we can proceed in a number of ways, but we will here use the Trust Region Reflective Algorithm~\cite{More:1983, Nocedal_2006}, which is a common approach in system identification~\cite{Matlab:TRRLS}. The idea of this algorithm is to iteratively replace $\varepsilon(\mathbcal{p})$ by a quadratic function $\vr{Q}(\mathbcal{p})$ that approximates $\varepsilon(\mathbcal{p})$ in a neighborhood $\mathcal{N}$ around the current point \mbox{$\mathbcal{p} = \mathbcal{p}_{curr}$}. This neighborhood is the mentioned trust region. At each iteration, an update of $\mathbcal{p}_{curr}$ is attempted by finding the point $\mathbcal{p}_{new}$ that minimizes $\vr{Q}(\mathbcal{p})$ subject to \mbox{$\mathbcal{p} \in \mathcal{N}$}. If 
\begin{equation*}
\varepsilon(\mathbcal{p}_{new}) < \varepsilon(\mathbcal{p}_{curr}),
\label{eq:chap4eqnonumber2}
\end{equation*}
we set $\mathbcal{p}_{curr} = \mathbcal{p}_{new}$ and iterate the process again. Otherwise, $\mathbcal{p}_{curr}$ remains unchanged, $\mathcal{N}$ is shrunk, and a new trial step is attempted.

The function $\vr{Q}(\mathbcal{p})$ is typically the second-order Taylor expansion of $\varepsilon(\mathbcal{p})$ around $\mathbcal{p}_{curr}$, 
\begin{equation*}
\vr{Q}(\mathbcal{p}) = \varepsilon(\mathbcal{p}_{curr}) + \vr{g}\trans \cdot (\mathbcal{p} - \mathbcal{p}_{curr}) + \frac{1}{2} (\mathbcal{p} - \mathbcal{p}_{curr})\trans \cdot \mt{H} \cdot (\mathbcal{p} - \mathbcal{p}_{curr}),
\label{eq:chap4eqnonumber3}
\end{equation*}
where $\vr{g}$ and $\mt{H}$ are the gradient and Hessian of $\varepsilon(\mathbcal{p})$ evaluated at $\mathbcal{p}_{curr}$. Since $\varepsilon(\mathbcal{p})$ typically has a complex expression, $\vr{g}$ and $\mt{H}$ are obtained by finite differences, so we only need to evaluate $\varepsilon(\mathbcal{p})$ repeatedly to obtain such derivatives. This is done by computing the $\vr{x}_k$ values from $\mathbcal{p}$, $\vr{x}_0$, \mbox{$\vr{u}_0, \ldots, \vr{u}_{k-1}$} using some integration method, then evaluating the outputs \mbox{$\vr{y}_k^{\mathrm{pred}}$} using Eq.~\eqref{eq:chap4eq3prime}, and finally computing the sum in Eq.~\eqref{eq:chap4eq3}.

In this work we have used the MATLAB implementation of the earlier algorithm, which is easily set up and run using the \emph{nlgreyest} method from the System Identification Toolbox~\cite{Matlab:nlgreyest}. To obtain the $\vr{x}_k$ values in Eq.~\eqref{eq:chap4eq3prime} we have used the \emph{ode45} method, which implements the Dormand-Prince algorithm of order 4/5~\cite{Dormand:1980}. This technique uses a Runge-Kutta method of fourth-order for time stepping, and a fifth-order method for error estimation and step adjustment. The obtained simulations are very precise therefore.

To apply the integration method, one has to decide how $\vr{u}(t)$ is to be interpolated from \mbox{$\vr{u}_0, \ldots, \vr{u}_N$}. On this regard we assume that the real robot is controlled with a zero-order-hold filter, so we also use this filter to interpolate $\vr{u}(t)$ in simulation. This means that \mbox{$\vr{u}(t) = \vr{u}_k$} for \mbox{$t_k \leq t < t_{k+1}$}, \mbox{$k=0, \ldots , N-1$}.
\begin{table}[!t]
\footnotesize
\begin{center}
\begin{tabular}{llll} 
\toprule
\textbf{Symbol} & \textbf{Meaning} & \textbf{Value} & \textbf{Unit} \\
\toprule
$l_1$ & Pivot offset relative to the wheels axis & $0.25$ & $\mathrm{m}$
\\ 
$l_2$ & One half of the wheels separation & $0.20$ & $\mathrm{m}$
\\ 
$r$ & Wheel radius & $0.10$ & $\mathrm{m}$ 
\\ 
$(x_B, y_B)$ & Coords. of c.o.m. $B$ of the chassis in chassis frame & $(-0.13, 0)$ & $\mathrm{m}$
\\
$(x_{F,0}, y_{F,0})$ & Coords. of c.o.m. $F$ of the unloaded platform in platform frame & $(0,0)$ & $\mathrm{m}$
\\ 
$m_c$ & Mass of the chassis including the wheels & $109.14$ & $\mathrm{kg}$ 
\\ 
$m_{p,0}$ & Mass of the platform without load & $21.95$ & $\mathrm{kg}$
\\ 
$I_c$ & Vertical moment of inertia of the chassis plus wheels at $B$ & $1.30$ & $\mathrm{kg} \; \mathrm{m}^2$
\\ 
$I_{p,0}$ & Vertical moment of inertia of the unloaded platform at $F$ & $2.22$ & $\mathrm{kg} \; \mathrm{m}^2$
\\ 
$I_a$ &  Axial moment of inertia of a wheel & $1.04 \times 10^{-2}$ & $\mathrm{kg} \; \mathrm{m}^2$
\\ 
$b_w$ & Viscous friction coefficient at the wheel's shaft & $0.18$ & $\mathrm{kg} \; \mathrm{m}^2 \; \mathrm{s}^{-1}$
\\
$b_p$ & Viscous friction coefficient at the pivot shaft & $0.24$ & $\mathrm{kg} \; \mathrm{m}^2 \; \mathrm{s}^{-1}$
\\
\bottomrule
\end{tabular}
\caption{Nominal parameters of the robot assumed in system identification tests.}
\label{tab:otbotparamvalues}
\end{center}
\end{table}

\subsection{Parameter identification sequence}
\label{sec:theidenprocess}

In order to use the inverse and forward dynamics equations~\eqref{eq:eq_simple_lagrange} and~\eqref{eq:eq_forward_dyn1} in our simulations, we need to determine the parameters that appear in the matrices $\mt{M}$, $\mt{C}$, $\mt{J}$ and $\mt{E}_f$. All parameters are assumed to be constant except those of the platform: its mass $m_p$, moment of inertia $I_p$, and coordinates of the c.o.m. $(x_F, y_F)$, which may be different from one task to another depending on the load carried by the robot. For convenience we denote the values of the unloaded platform parameters by $m_{p,0}$,  $I_{p,0}$, $x_{F,0}$ and  $y_{F,0}$, respectively, and we will refer to the varying parameters $m_p, I_p, x_F, y_F$, as the working platform parameters. 

Table~\ref{tab:otbotparamvalues} summarizes all the parameters with their nominal values. We assume that some of them are already known or that they can be directly measured, so we will not specify any identification procedure for them. This applies to the parameters $l_1$, $l_2$, $r$, $m_{p,0}$, $x_{F,0}$, and $y_{F,0}$, whose nominal values will be taken for granted.

The identification process is divided into three consecutive steps (Table~\ref{tab:threeidensteps}), as described next:
\begin{enumerate}
\renewcommand\labelenumi{\bfseries\theenumi}
\item \textbf{Identification of basic parameters:} This step involves parameters of individual parts of the robot that can be identified by the activation of a single motor. This includes the vertical moment of inertia of the unloaded platform, $I_{p,0}$, the moment of inertia of a wheel with respect to its rotation axis, $I_a$, and their corresponding friction coefficients, $b_p$, $b_w$. We assume the center of mass of the platform is at the pivot point, and that both wheels are identical and have the same friction coefficient $b_w$.
\item \textbf{Identification of the chassis parameters:} In his step, we try to determine the four parameters of the chassis: $m_c, I_c, x_B, y_B$. For this, we assume that the platform is unloaded, so that its parameters are all previously known or obtained from step 1.
\item \textbf{Identification of the working platform parameters:} In this step we will identify the parameters corresponding to the platform together with the unknown load placed on it: $m_p, I_p, x_F, y_F$, which may be different for each experiment. In this case we make use of the parameter values already identified in steps 1 and 2. This step must be repeated each time the load carried by the robot changes.
\end{enumerate}

\begin{table}[h!]
	\begin{center}
		\small
		\begin{tabular}{lll} 
			\toprule
			\textbf{Step 1} & \textbf{Step 2} & \textbf{Step 3} \\
			\textbf{Basic parameters} & \textbf{Chassis parameters} & \textbf{Working-platform parameters} \\
			\toprule
			$I_{p,0}$ & $m_c$ & $m_p$\\
			$I_a$     & $I_c$ & $I_p$\\
			$b_w$     & $x_B$ & $x_F$\\
			$b_p$     & $y_B$ & $y_F$\\
			\bottomrule
		\end{tabular}
		\caption{The three identification steps.}
		\label{tab:threeidensteps}
	\end{center}
\end{table}

\subsection{Identification results}
\label{sec:identresults}

The identification process requires the determination of the $\vr{u}_k$ and $\vr{y}_k$ sequences defining the fitting function~\eqref{eq:chap4eq3}. In a real robot, this would involve executing some control sequence $\vr{u}_k$ in the real world while recording the outputs $\vr{y}_k$ provided by the sensor readings. Since in our case the physical robot is not available, we will obtain the outputs $\vr{y}_k$ by an accurate simulation of the dynamics using the nominal parameter values. To be more realistic, we will add some Gaussian noise to the output of each sensor, with standard deviations that we consider plausible for each kind of sensor. Then, in practice, for each identification experiment we perform multiple simulations with the same control action sequence: a first one that plays the role of the real robot using the nominal parameters to obtain the corresponding $\vr{y}_k$ sequence, and one more for each set of parameters to test to obtain the $\vr{y}_k^{\mathrm{pred}}$ sequence, all simulations using the same dynamic equations and integration algorithm.

\subsubsection{Identification of basic parameters}
\label{subsec:basparident}

The parameters to be identified in this step are the moment of inertia $I$ and friction coefficient $b$ of two robot parts: the unloaded platform and a wheel. In both cases there is a motor acting directly on the corresponding axis, and we assume that both motors are equipped with angular encoders able to provide an accurate measure of their angular velocity. The identification procedure is analogous in both cases. It consists in applying a constant torque to the corresponding axis and registering the readings of the encoder. Thus, in this case, the output function in Eq.~\eqref{eq:chap4eq2} simply provides the angular velocity. The dynamic equation governing the motion is
\begin{equation}
u - b \cdot \dot{\varphi} = I \cdot \ddot{\varphi}
\label{eq:chap4BPIeq1}
\end{equation}
where $(u, b, \varphi, I) = (\tau_p, b_p, \varphi_p, I_{p,0})$ in the case of the platform, and $(\tau_r, b_w, \varphi_r, I_a)$ in the case of the (right) wheel. In each experiment, two parameters must be identified simultaneously, $I$ and $b$.

\begin{figure}[t!]
    \centering
    \begin{subfigure}[b]{\textwidth}
    	\centering
      \includegraphics[width=\textwidth]{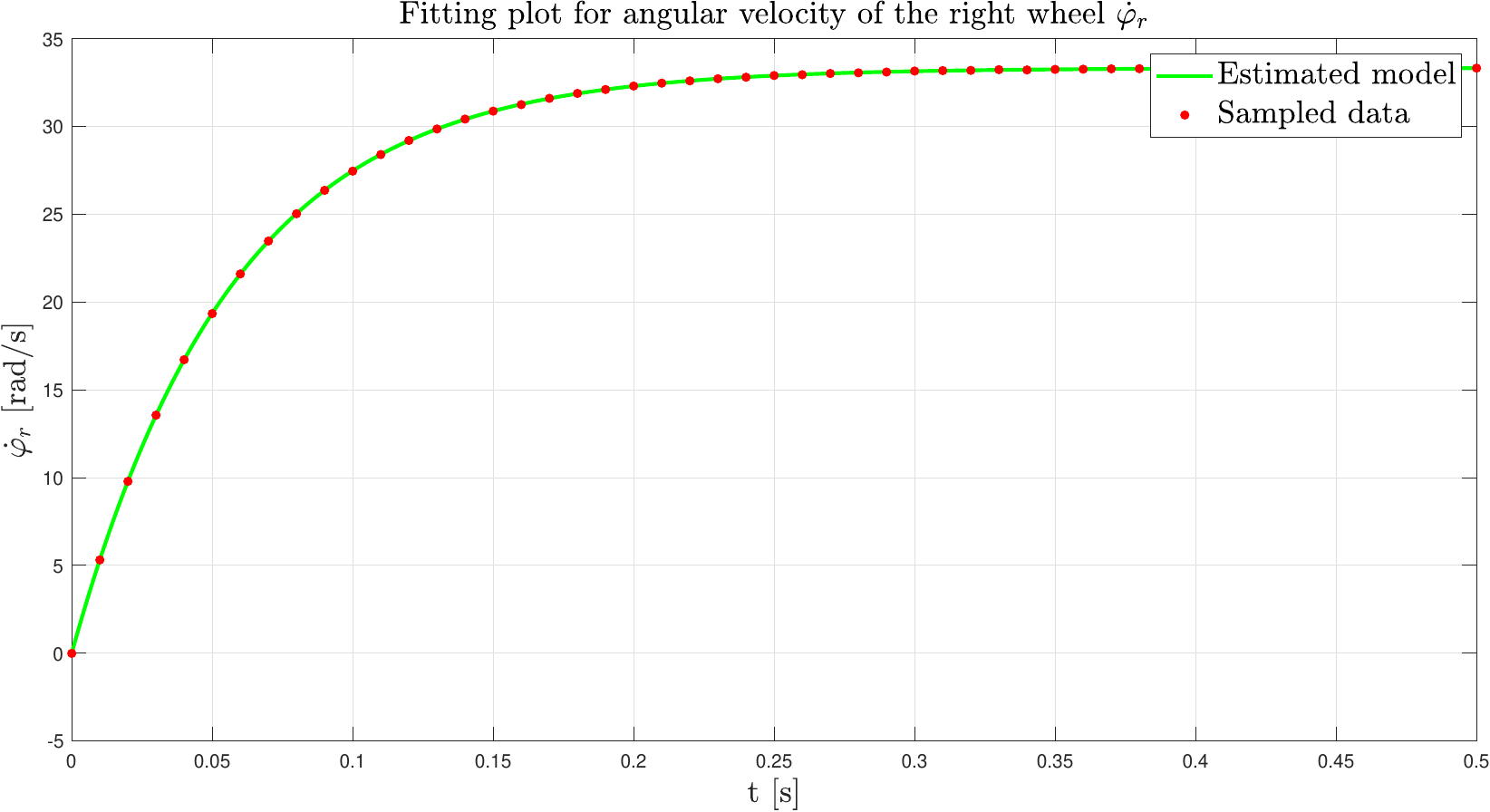}
      \label{fig:fittingplotwheelshaft}
    \end{subfigure}

	  \vspace{10mm}

    \begin{subfigure}[b]{\textwidth}
    	\centering
    	\includegraphics[width=\textwidth]{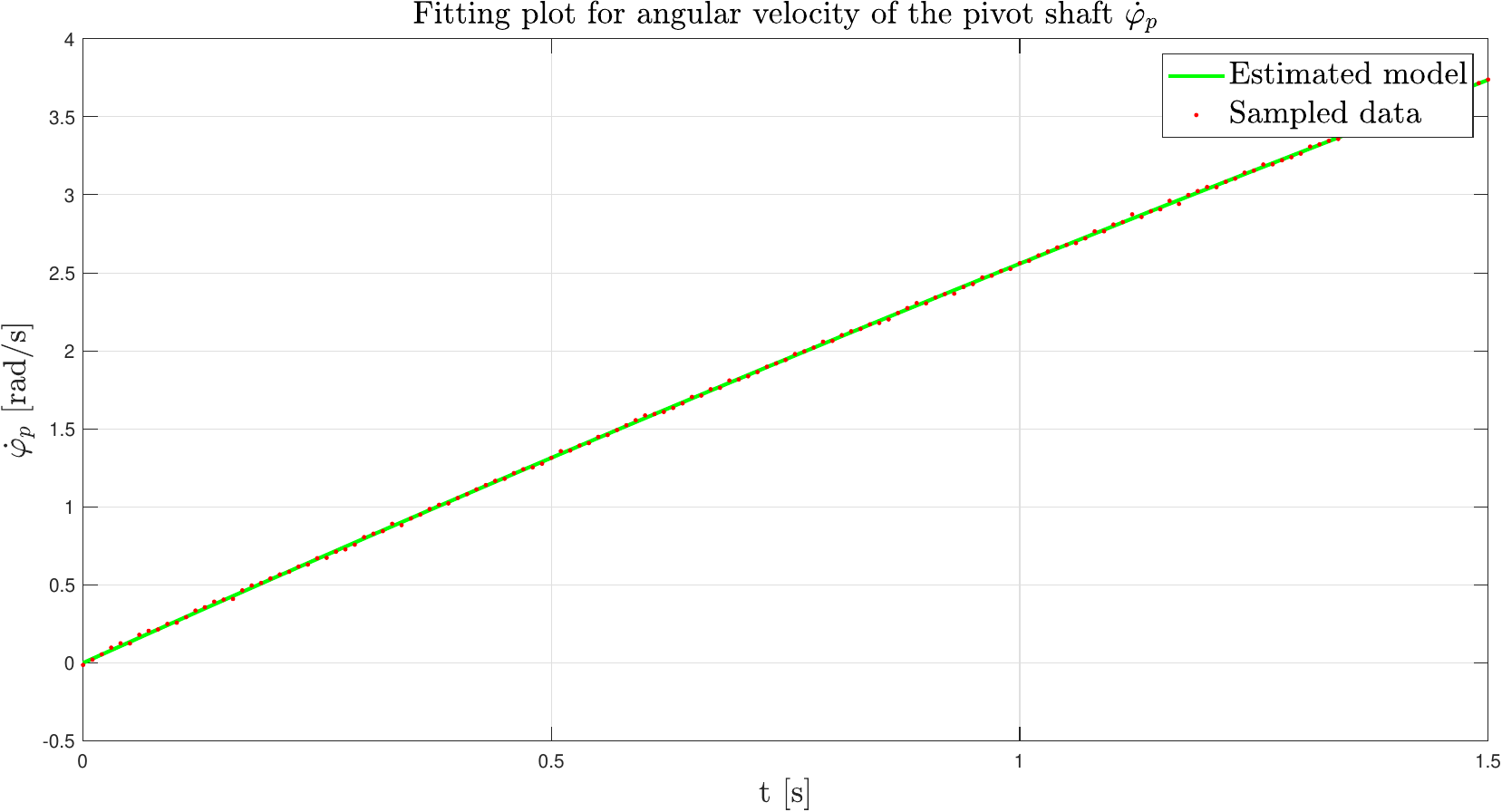}
      \label{fig:fittingplotpivotshaft}
    \end{subfigure}

    \caption{Fitting plots for the wheel and platform parameter identification.}
    \label{fig:FittPlotsStep1}
\end{figure}

To perform the experiments, we applied a constant torque of $6\mathrm{N} \mathrm{m}$ during $1.5\mathrm{s}$ for the platform and $0.5\mathrm{s}$ for the wheel, sampling the angular velocity at a frequency of $100\mathrm{Hz}$. The simulated encoder readings are corrupted by Gaussian noise with a standard deviation of $\sigma = 0.01 \mathrm{rad}/\mathrm{s}$, which reflects the good precision usually provided by digital encoders.
The initial guess (the initial $\mathbcal{p}_{curr}$ of the Trust Region Reflective Algorithm) and the final estimated value for each parameter are shown in Table~\ref{tab:initialguesstabStep1}.
\begin{table}[t!]
\begin{center}
\small
\begin{tabular}{llllll} 
\toprule
\textbf{Parameter} & \textbf{Nominal value} & \textbf{Initial guess} & \textbf{Estimated value} & \textbf{Absolute error} & \textbf{Unit}\\
\toprule
$b_w$     &  $0.18$    &  $0.09$   &  $0.18$   &  $8.61 \times 10^{-6}$   & $\mathrm{kg} \; \mathrm{m}^2 \; \mathrm{s}^{-1}$\\
$I_a$     &  $0.01$    &  $0.01$   &  $0.01$   &  $7.40 \times 10^{-6}$   & $\mathrm{kg} \; \mathrm{m}^2$\\
$b_p$     &  $0.24$    &  $0.12$   &  $0.24$   &  $4.90 \times 10^{-4}$   & $\mathrm{kg} \; \mathrm{m}^2 \; \mathrm{s}^{-1}$\\
$I_{p,0}$ &  $2.22$    & $1.11$    &  $2.22$      &  $2.77 \times 10^{-4}$   & $\mathrm{kg} \; \mathrm{m}^2$\\
\bottomrule
\end{tabular}
\caption{Results of the basic parameters identification.}
\label{tab:initialguesstabStep1}
\end{center}
\end{table}
The results of the experiments are shown in Fig.~\ref{fig:FittPlotsStep1}.

\subsubsection{Identification of the chassis parameters}
\label{subsec:chasparamident}

In this step, the four parameters of the chassis must be identified simultaneously by driving the robot with some control action sequence. To obtain a precise identification of all four parameters, it is necessary that the trajectory followed in the experiment provides sufficient information to reveal the influence of each parameter in the trajectory, otherwise, some parameter could become irrelevant, making it unidentifiable, or maybe the observed outputs could be obtained with different combinations of parameters. Several actuation patterns involving the three motors have been tried, using equal or different torque profiles, combining them in different ways, putting them in and out of synchrony, and so on. Our purpose was to find an action sequence able to provide the required information in a short time and using limited torques, so that the robot displacement is confined in a reduced area. The final choice that produced good results was to apply constant torques to each motor during $3\mathrm{s}$ with the following values: 
\begin{align*}
	&\tau_r = 6 \, \text{Nm}   \\
	&\tau_l = -10 \, \text{Nm} \\
	&\tau_p = 6 \, \text{Nm}
\end{align*}

\begin{figure}[b!]
	\begin{subfigure}{.5\textwidth}
		\centering
		\includegraphics[width=\textwidth]{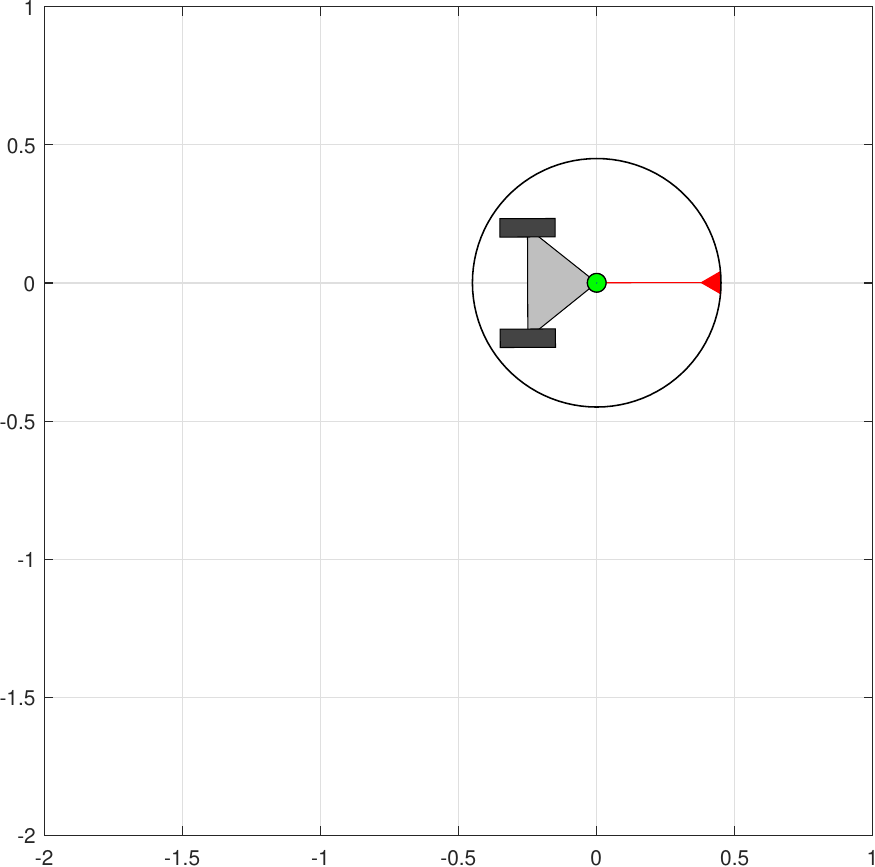}
		\subcaption{Otbot at the initial configuration.}
		\label{fig:Setp2Motiont0}
	\end{subfigure}		
	\begin{subfigure}{.5\textwidth}
		\centering
		\includegraphics[width=\textwidth]{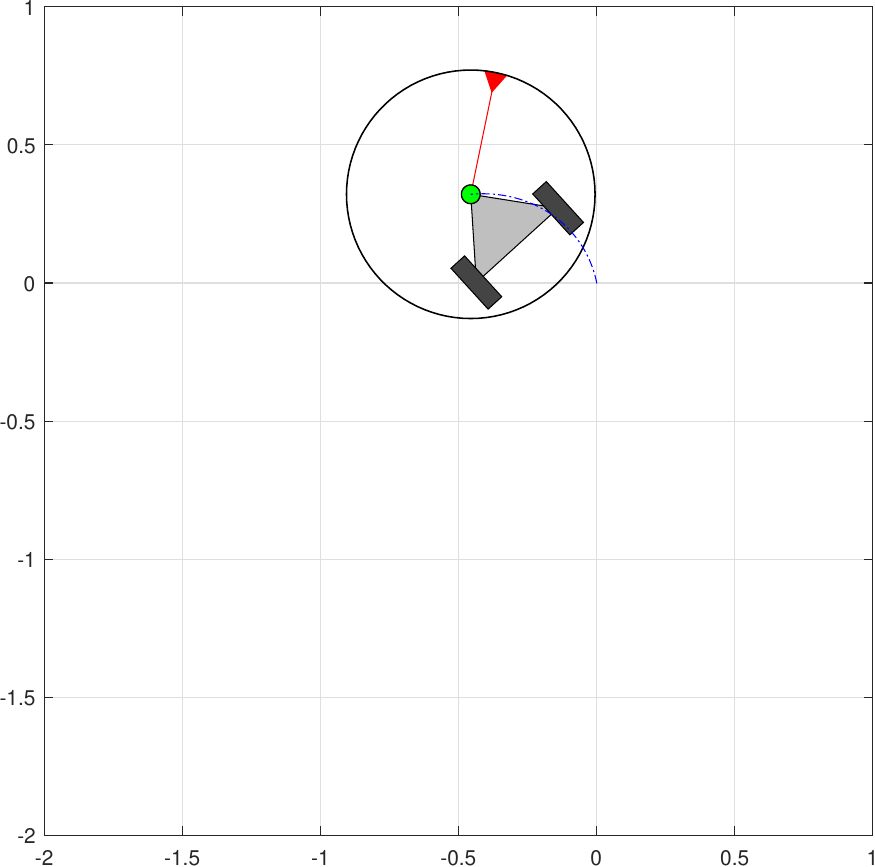}
		\subcaption{Otbot configuration after $1 \mathrm{s}$.}
		\label{fig:Setp2Motiont1}
	\end{subfigure}
	
	\vspace{10mm}
	
	\begin{subfigure}{.5\textwidth}
		\centering
		\includegraphics[width=\textwidth]{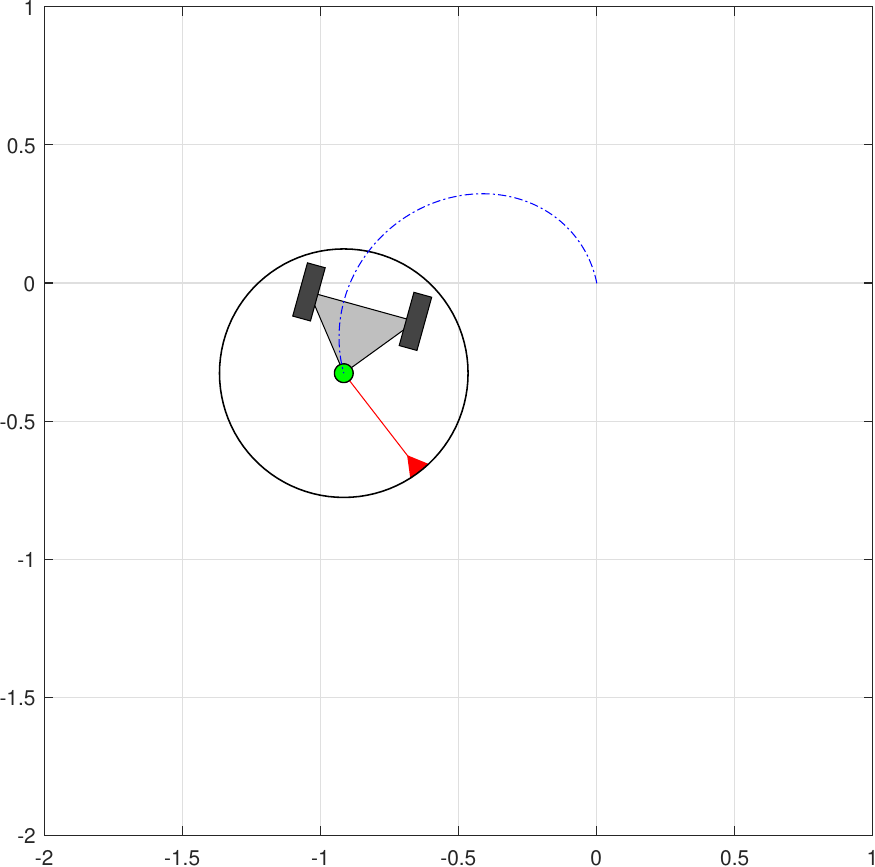}
		\subcaption{Otbot configuration after $2 \mathrm{s}$.}
		\label{fig:Setp2Motiont2}
	\end{subfigure}		
	\begin{subfigure}{.5\textwidth}
		\centering
		\includegraphics[width=\textwidth]{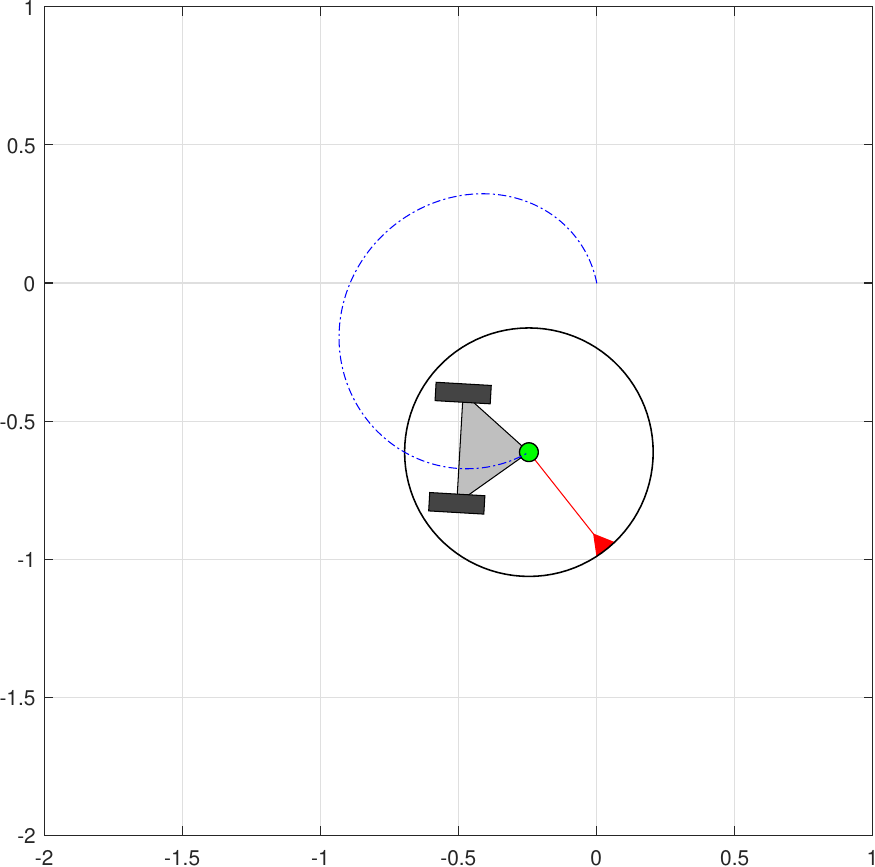}
		\subcaption{Otbot configuration after $3 \mathrm{s}$.}
		\label{fig:Setp2Motiont3}
	\end{subfigure}
	
	\vspace{2mm}
	
	\caption{Experiment for the identification of the chassis parameters. The path followed by the pivot joint is shown in blue. See \url{https://youtu.be/SChkwezm7MM} for an animation.}
	\label{fig:Setp2Motion}
\end{figure}

\begin{figure}[b!]
	\centering
	\begin{subfigure}[b]{\textwidth}
		\centering
		\includegraphics[width=0.8\textwidth]{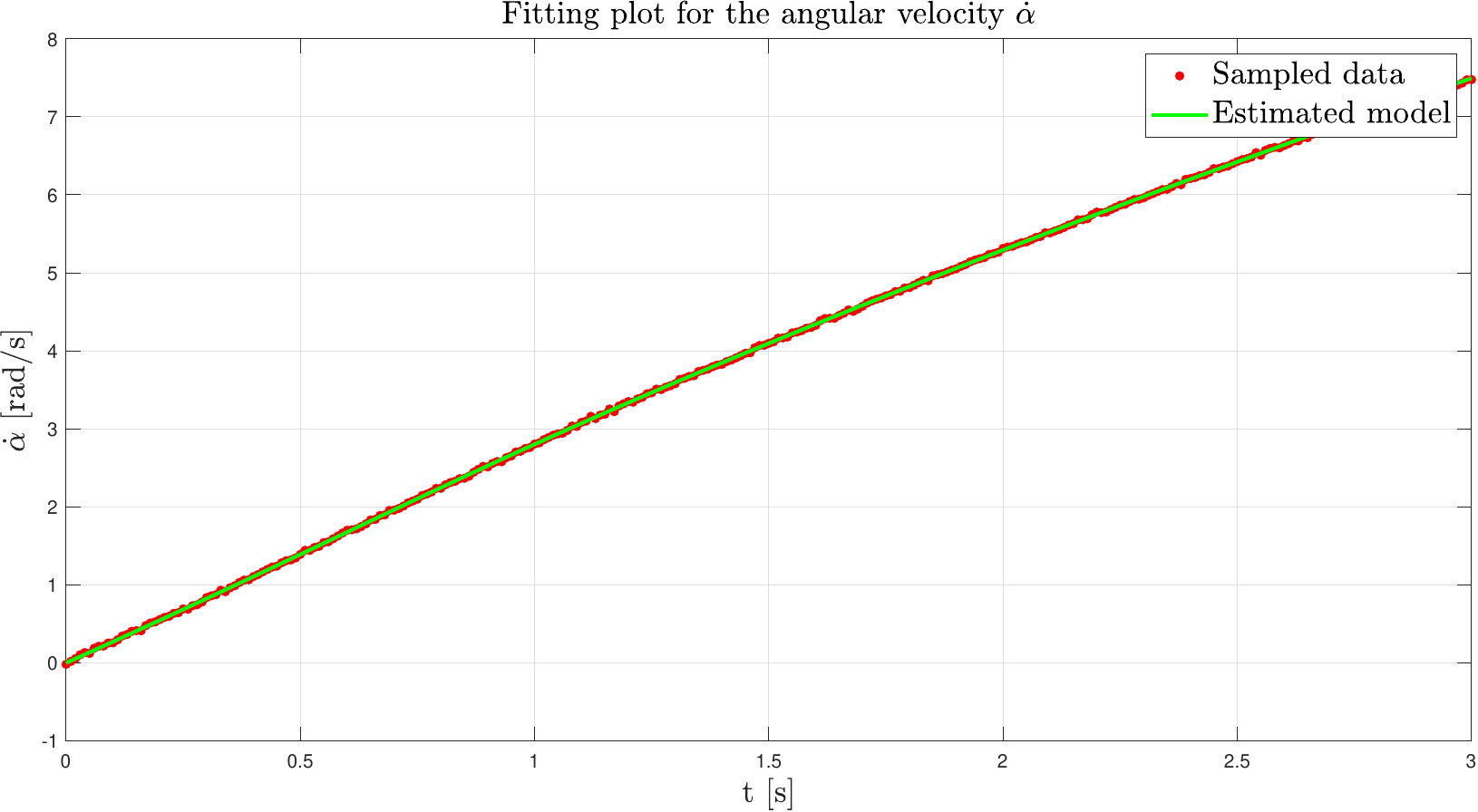}
		\label{fig:fittingplotdalphaStep2}
	\end{subfigure}
	
	\vspace{2mm}
	
	\begin{subfigure}[b]{\textwidth}
		\centering
		\includegraphics[width=0.8\textwidth]{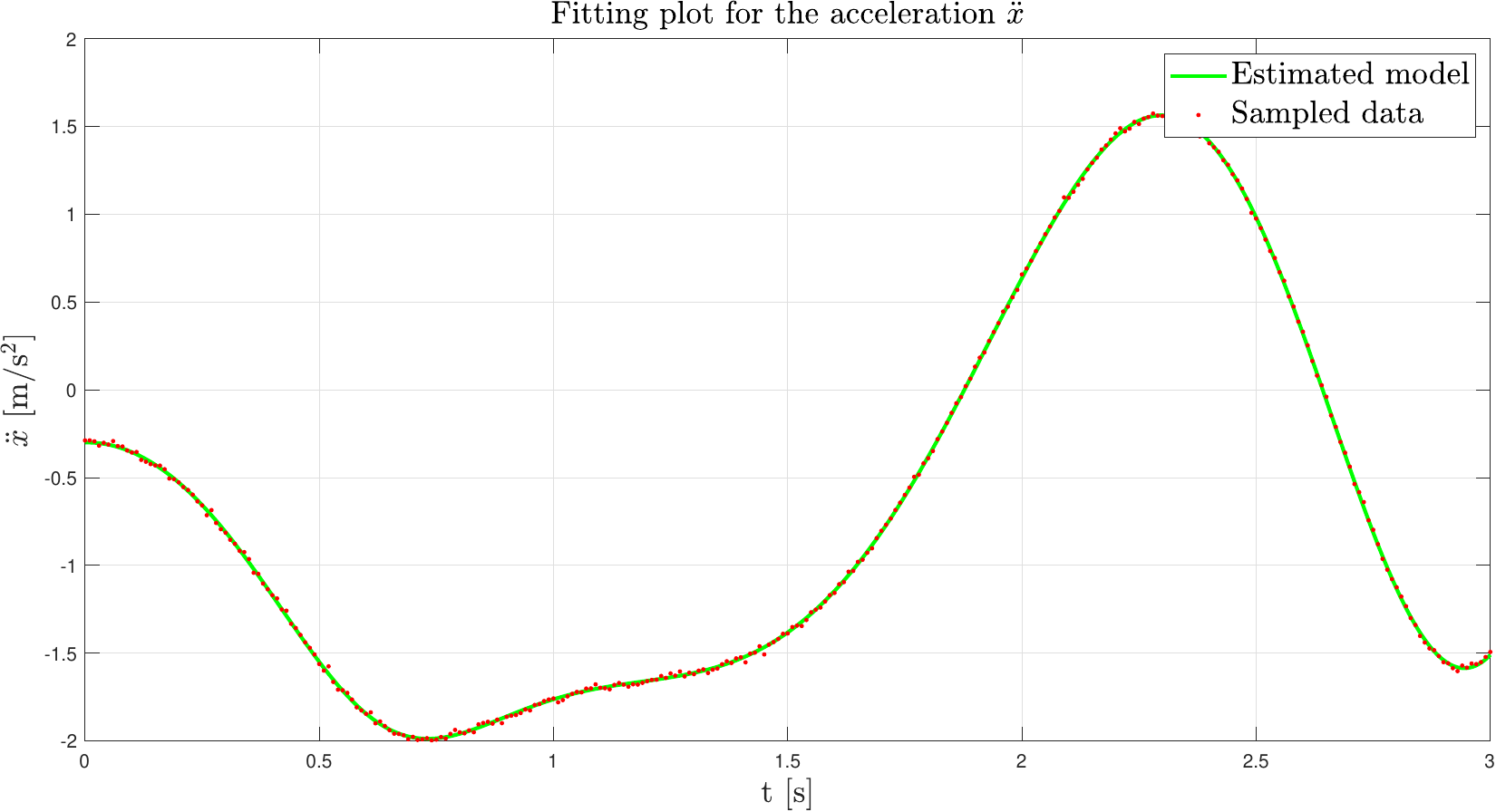}
		\label{fig:fittingplotddxStep2}
	\end{subfigure}
	
	\vspace{2mm}
	
	\begin{subfigure}[b]{\textwidth}
		\centering
		\includegraphics[width=0.8\textwidth]{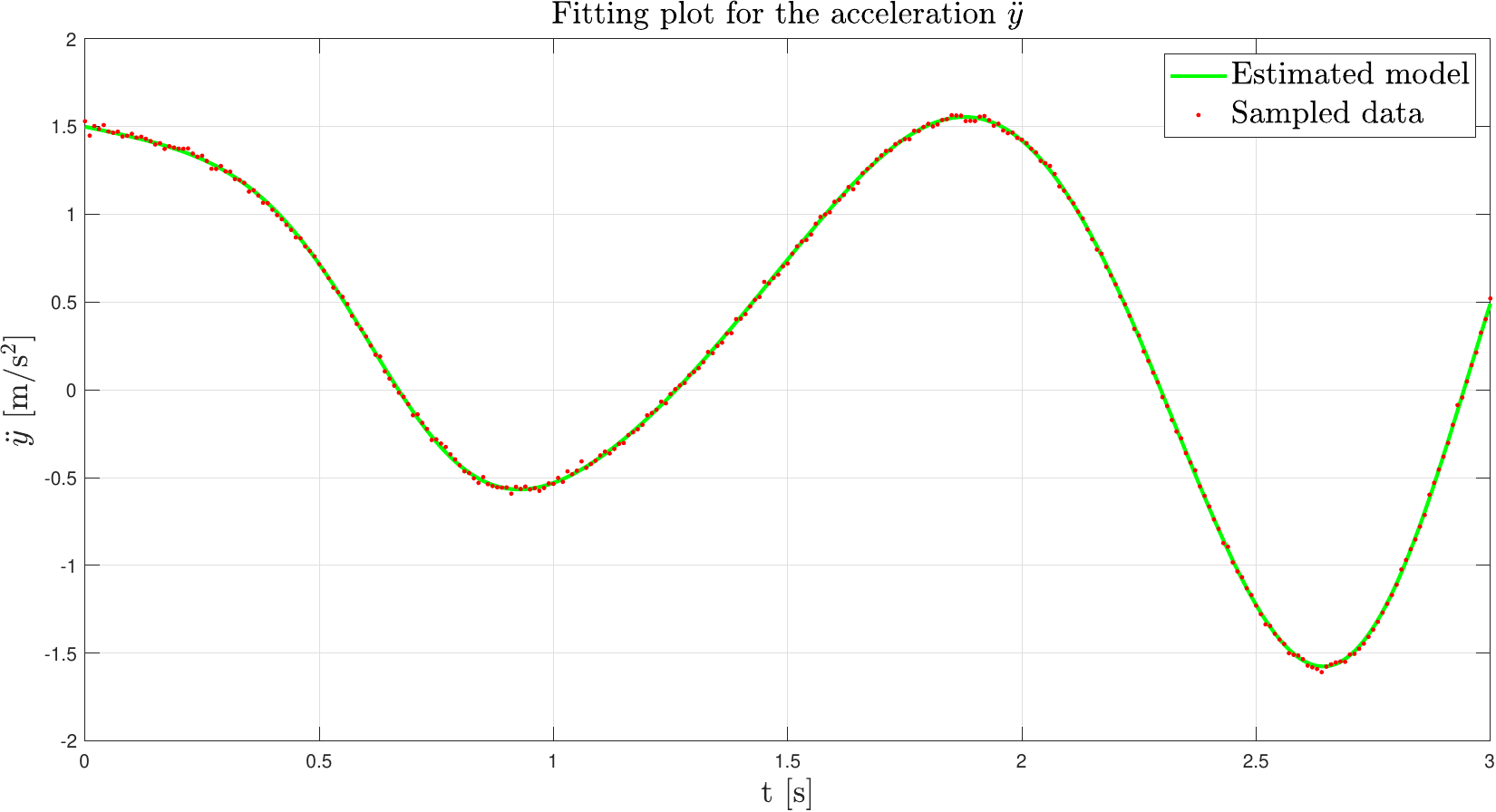}
		\label{fig:fittingplotddyStep2}
	\end{subfigure}
	
	\caption{Fitting plots relative to the identification of the chassis parameters.}
	\label{fig:FittPlotsStep2}
\end{figure}

The output signals $\vr{y}_k$ are obtained assuming that the robot is equipped with an IMU attached to the platform that provides the instantaneous values of the acceleration of the pivot point \mbox{$(\ddot{x},\ddot{y})$} in basis $B''$, and the angular velocity of the platform $\dot{\alpha}$. As in step 1, we sampled the sensor at a frequency of $100\mathrm{Hz}$. The standard deviation of the noise applied to the sensor readings has been computed according to the usual formula for an IMU
\begin{equation}
\sigma_{IMU} = ND\sqrt{SR},
\label{eq:chap4IMUnoise}
\end{equation}
where $ND$ is the noise density, a parameter provided by the manufacturer that we assume to have a value of $1.37 \times 10^{-3} \mathrm{m} \; \mathrm{s}^{-2} \; \mathrm{Hz}^{-1/2}$, and $SR$ is the sample rate, that in our case is $100 \mathrm{Hz}$. By replacing these values into Eq.~\eqref{eq:chap4IMUnoise} we end up with a standard deviation of $\sigma_{IMU} = 13.73 \times 10^{-3}$. The initial guess and the final estimated value for each parameter are shown in Table~\ref{tab:initialguesstabStep2}.
\begin{table}[t!]
\begin{center}
	\small
\begin{tabular}{llllll} 
\toprule
\textbf{Parameter} & \textbf{Nominal value} & \textbf{Initial guess} & \textbf{Estimated value} & \textbf{Absolute error} & \textbf{Unit}\\
\toprule
$m_c$     &  $109.14$  &  $54.57$   &  $109.12$                &  $0.02$                 & $\mathrm{kg}$\\
$I_c$     &  $1.30$    &  $0.65$    &  $1.31$                  &  $8.87 \times 10^{-4}$  & $\mathrm{kg} \; \mathrm{m}^2$\\
$x_B$     &  $-0.13$   &  $-0.07$   &  $-0.13$                 &  $1.72 \times 10^{-5}$  & $\mathrm{m}$\\
$y_B$     &  $0.00$       &  $0.25$    &  $-4.31 \times 10^{-5}$  &  $4.31 \times 10^{-5}$  & $\mathrm{m}$\\
\bottomrule
\end{tabular}
\caption{Results of the chassis parameters identification.}
\label{tab:initialguesstabStep2}
\end{center}
\end{table} 
The result of the experiment is shown in Fig.~\ref{fig:FittPlotsStep2}.

\subsubsection{Identification of the working platform parameters}
\label{subsec:loadplatfidentif}

This step is conceptually very similar to the previous one, since the four parameters we have to identify for the working platform correspond to the four parameters identified in step 1 for the chassis. The difference is that, instead of fixing the platform parameters and estimating those of the chassis, we will fix the parameters of the chassis and try to estimate those of the platform, which now we assume to be unknown. A further difference with step 1 is the larger deviation from the nominal parameter values imposed to the initial guess. This is required because we assume that the robot should be able to carry loads several times heavier than the platform itself, so that we may have only a very rough idea of its possible mass, moment of inertia and position of the c.o.m., and it is important to check that the identification process is reliable in the wide range of working conditions that can be expected.

In order to assess the robustness of the identification in front of the deviation of the initial guess, we performed a series of tests with increasing amounts of deviation. For convenience reasons, instead of using different weights and positions of the load and perform the identification with the same initial guess, we proceed in the opposite way, i.e., we increase the deviation of the initial guess while keeping the platform parameters unmodified. This allows us to use the same experiment as in step 1, with the same simulated sensor data for all tests, thus avoiding the need to simulate a new trajectory for each load configuration. This strategy permits to test many instances of parameter deviations of different magnitudes in a fast way.

\begin{figure}[p]
    \centering
    \begin{subfigure}[b]{\textwidth}
    	\centering
      \includegraphics[width=\textwidth,height=90mm]{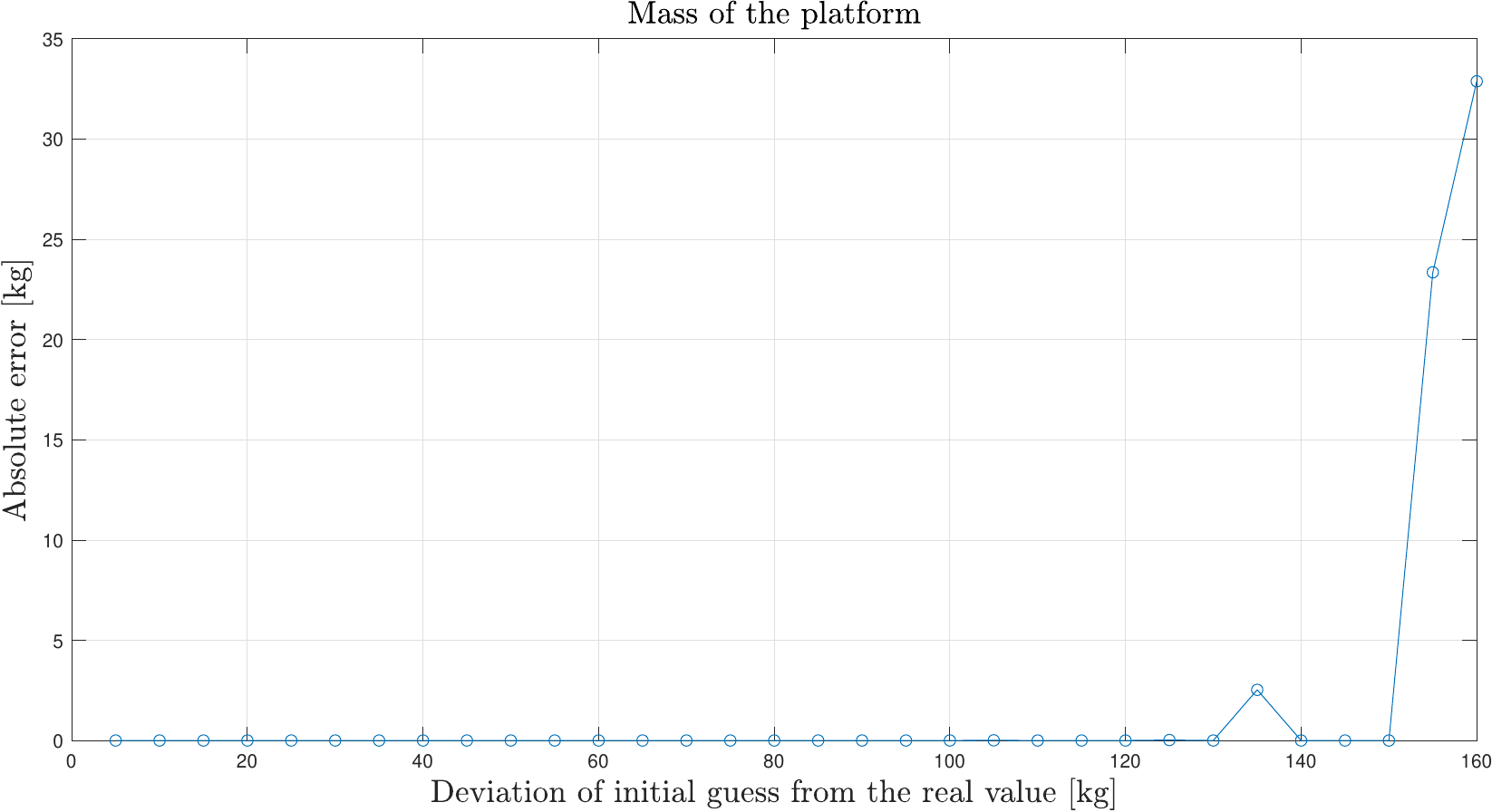}
      \label{fig:sensibplotStep3mp}
    \end{subfigure}

	  \vspace{10mm}

    \begin{subfigure}[b]{\textwidth}
    	\centering
    	\includegraphics[width=\textwidth,height=90mm]{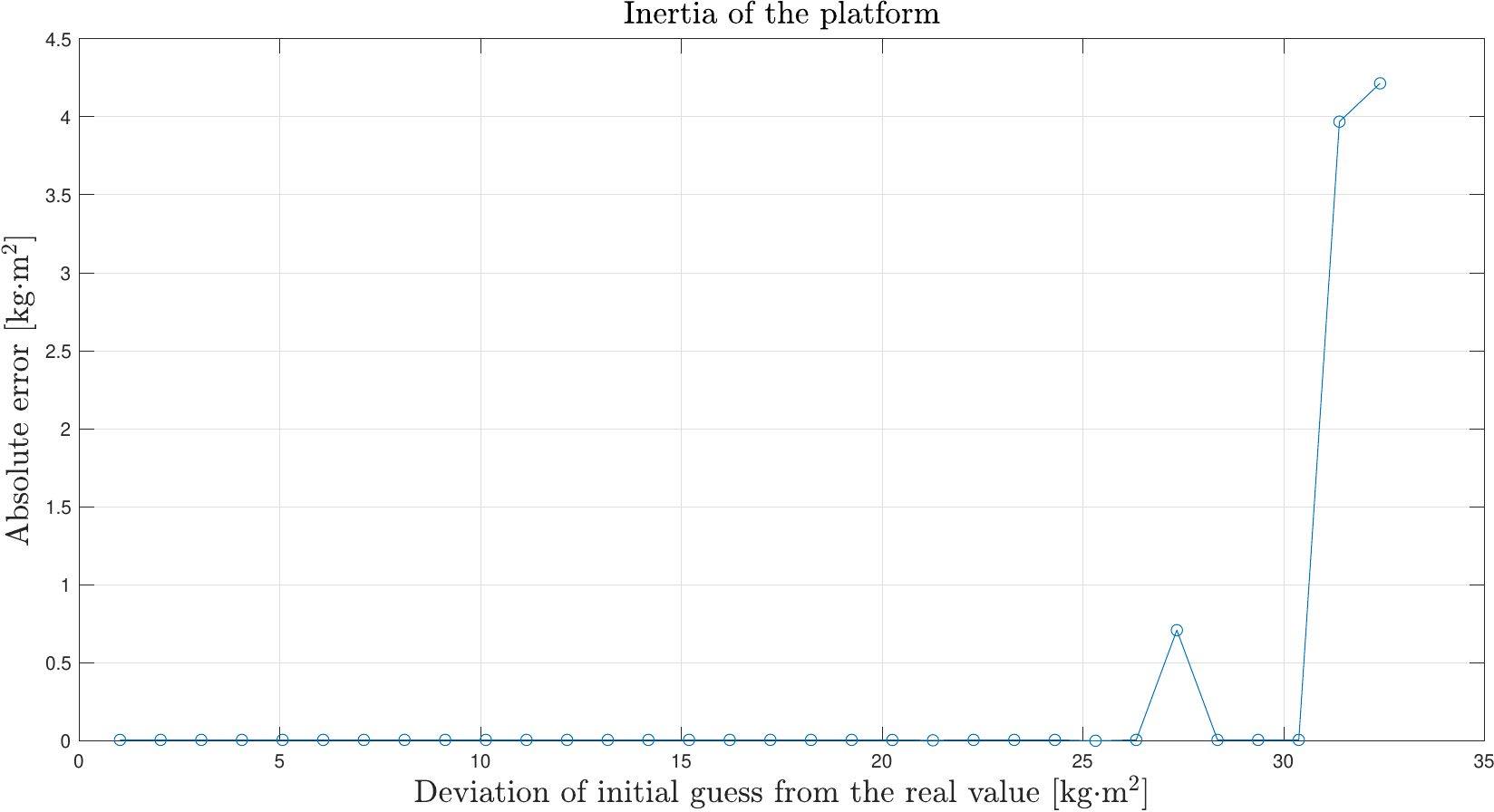}
      \label{fig:sensibplotStep3Ip}
    \end{subfigure}

    \caption{Sensitivity plots for the inertial parameters of the working platform. The curves show the absolute error in the mass and moment of inertia of the platform as a function of the deviation of their initial guesses from the real values.}
    \label{fig:sensibplotStep3part1}
\end{figure}

The range of variations of the initial guess to test has been determined from the assumption that the mass of the load can reach a maximum of $500\mathrm{kg}$, and its position on the platform may be arbitrary within a circle with a radius of $0.45\mathrm{m}$ around the pivot. To avoid a proliferation of tests, we build each initial guess by altering the load mass and distance to the pivot with the same percentage of their maximum allowed values, which is a rather unfavorable choice. For the moment of inertia, we take a value compatible with the selected mass and distance according to the Steiner theorem. Figure~\ref{fig:sensibplotStep3part1} shows the sensitivity plots relating the absolute error on each parameter as a function of the amount of deviation. It can be seen that except for a sporadic outlier, all parameters are identified with great precision when the amount of deviation is as large as more than $25\%$ of the maximum allowed. It has to be noted that, despite the trajectory of the experiment has a duration of $3\mathrm{s}$, good  results are already obtained after the first second, and thus, all the tests have been made using only $1\mathrm{s}$ of simulation.

\begin{figure}[p]
    \centering
    \begin{subfigure}[b]{\textwidth}
    	\centering
      \includegraphics[width=\textwidth,height=90mm]{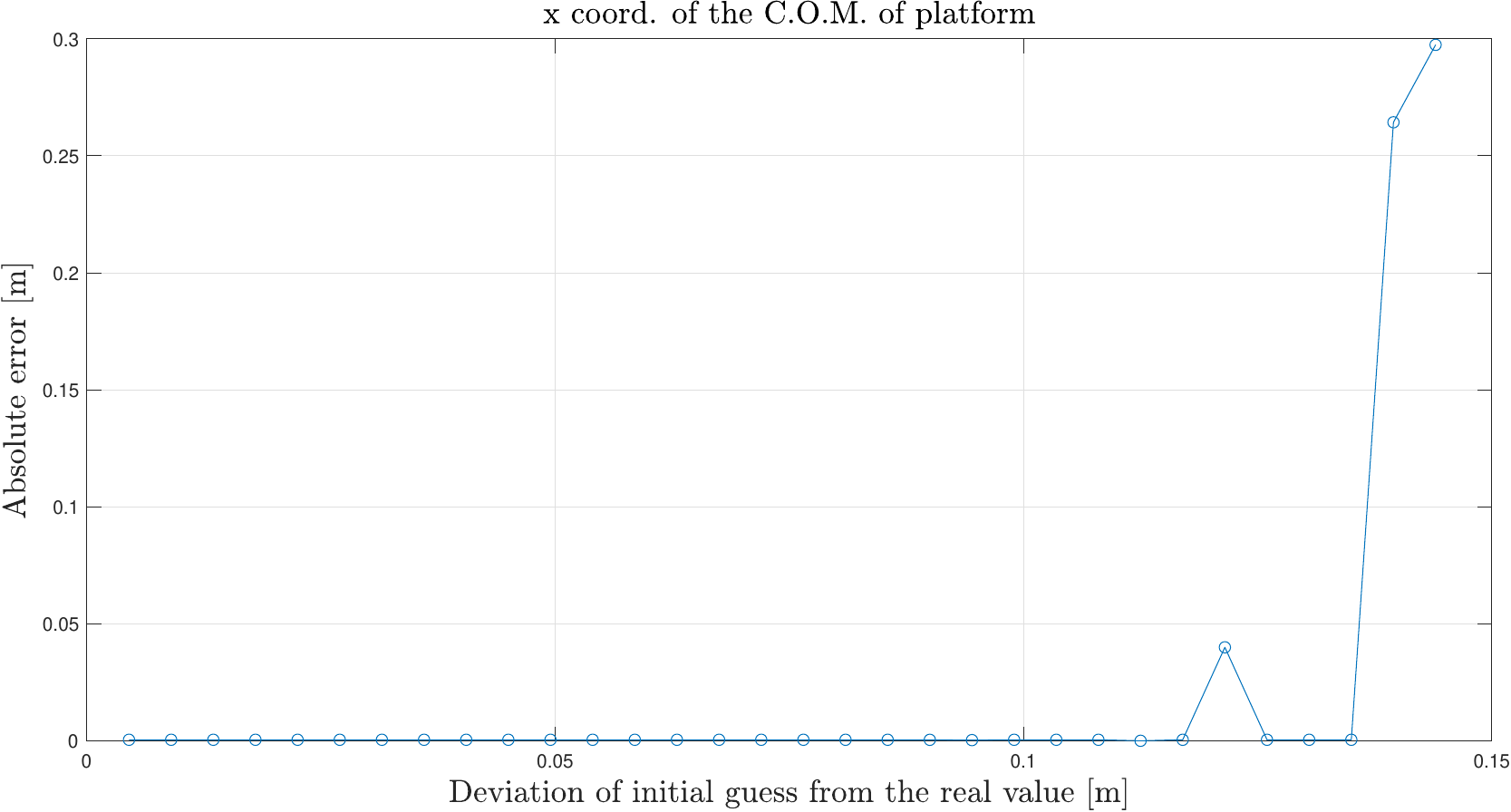}
      \label{fig:sensibplotStep3xF}
    \end{subfigure}

	  \vspace{10mm}

    \begin{subfigure}[b]{\textwidth}
    	\centering
    	\includegraphics[width=\textwidth,height=90mm]{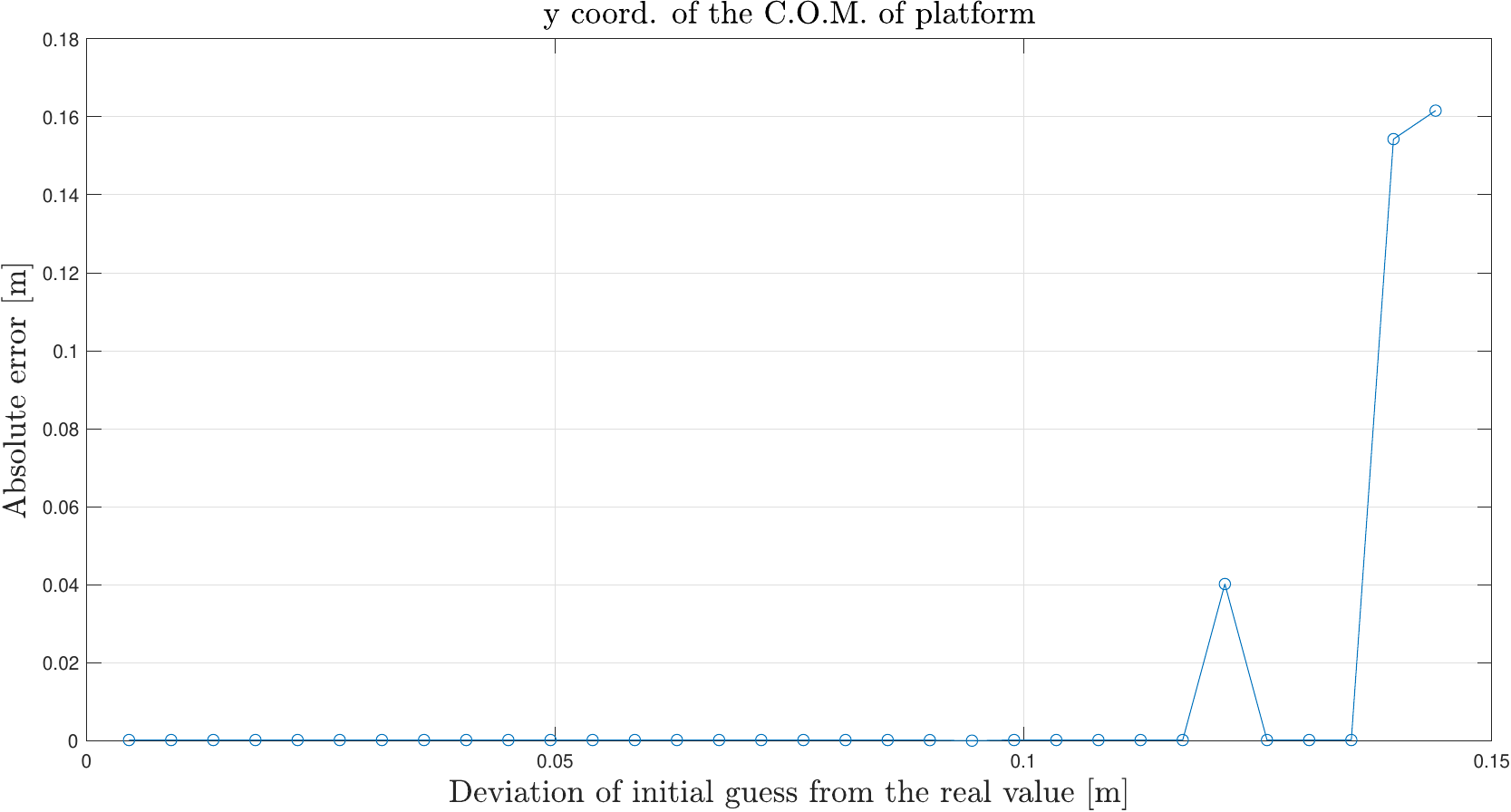}
      \label{fig:sensibplotStep3yF}
    \end{subfigure}

    \caption{Sensitivity plots for the coordinates of the center of mass of the working platform. The curves show the absolute error in the coordinates of the center of mass of the platform as a function of the deviation of their initial guesses from the real values.}
    \label{fig:sensibplotStep3part2}
\end{figure}

To illustrate the accuracy of the estimation, Table~\ref{tab:initialguesstabStep3} shows the results obtained for a $25\%$ deviation corresponding to a mass of the load of $120\mathrm{kg}$ (which corresponds to a parameter value $m_p = 146.95\mathrm{kg}$ when added to that of the unloaded platform), and a position of its center of mass \mbox{$(x_F, y_F) = (0.11, 0.11)\mathrm{m}$}.
\begin{table}[t!]
\begin{center}
\small
\begin{tabular}{llllll} 
\toprule
\textbf{Parameter} & \textbf{Nominal value} & \textbf{Initial guess} & \textbf{Estimated value} & \textbf{Absolute error} & \textbf{Unit}\\
\toprule
$m_p$     &  $21.95$ &  $146.95$  &  $21.99$                   &  $0.05$                & $\mathrm{kg}$\\
$I_p$     &  $2.22$  &  $5.94$    &  $2.22$                    &  $1.15 \times 10^{-3}$ & $\mathrm{kg} \; \mathrm{m}^2$\\
$x_F$     &  $0.00$     &  $0.11$    &  $7.50 \times 10^{-5}$     &  $7.50 \times 10^{-5}$ & $\mathrm{m}$\\
$y_F$     &  $0.00$     &  $0.11$    &  $-2.32 \times 10^{-4}$    &  $2.32 \times 10^{-4}$ & $\mathrm{m}$\\
\bottomrule
\end{tabular}
\caption{Identification results of the working platform parameters.}
\label{tab:initialguesstabStep3}
\end{center}
\end{table}
\newpage

\section{Tracking control}
\label{ch:trackingcontrl}

We next move on to designing a controller to allow the robot to track a desired trajectory. For this purpose we have implemented a computed-torque controller that is able to converge to the trajectory from any starting state. This controller will also allow the robot to counteract unforeseen force perturbations or deviations from the trajectory.

\subsection{A computed-torque controller}
\label{sec:CTC}

This type of controller can only stabilize as many variables as the number of actuated degrees of freedom of the robot. Since there are only three actuated joints in Otbot (the wheel and pivot joints), we need to base this controller on an equation of motion that uses only three configuration variables and their derivatives. Fortunately, we have already computed the equation of motion in the task-space coordinates \mbox{$\vr{p} = (x,y,\alpha)$} in Eq.~\eqref{eq:eq_dyn_task_space}, which we here recall:
\begin{equation}
    \bar{\mt{M}} \; \vr{\ddot{p}} + \bar{\mt{C}} \; \vr{\dot{p}} =\vr{u}.
    \label{eq:eq_dyn_task_space2}
\end{equation}
This equation is ideal since it will let us control the \vr{p} variables, which are those typically used to specify a task or mission for the robot. Thus, our goal is to find a control law that makes a robot trajectory $\vr{p}(t)$ asymptotically convergent to a desired trajectory $\vr{p}_d(t)$. To this end, we first convert the ODE in Eq.~\eqref{eq:eq_dyn_task_space2} into a linear one using the feedback law
\begin{equation}
\vr{u}= \bar{\mt{M}} \; \vr{v} + \bar{\mt{C}} \; \vr{\dot{p}},
\label{eq:eq_feedback_law}    
\end{equation}
where $\vr{v} \in \mathbb{R}^3$ is a new control input. Note that if we substitute Eq.~\eqref{eq:eq_feedback_law} into~\eqref{eq:eq_dyn_task_space2} we obtain
\begin{equation*}
\bar{\mt{M}} \; \vr{\ddot{p}} = \bar{\mt{M}} \; \vr{v}
\label{eq:eq_u2withMbar}
\end{equation*}
and since $\bar{\mt{M}}$ is nonsingular this implies that 
\begin{equation}
\vr{\ddot{p}} = \vr{v},
\label{eq:eq_u2}
\end{equation}
so the system now exhibits double-integrator dynamics.

The system in Eq.~\eqref{eq:eq_u2} is easy to stabilize along $\vr{p}_d(t)$. Since its three scalar equations are uncoupled, it suffices to design a controller for each scalar subsystem independently. To do so, let us write the $i$-th component of Eq.~\eqref{eq:eq_u2} as
\begin{equation}
\ddot{p} = v
\label{eq:eq_u2_scalar}
\end{equation}
and let $p_d(t)$ be the desired trajectory for $p$. The first-order form of Eq.~\eqref{eq:eq_u2_scalar} is
\begin{equation*}
\underbrace{
\left[
\begin{array}{c}
\dot{p}\\
\ddot{p}
\end{array}
\right]}_{\vr{\dot{z}}} = \underbrace{\left[
\begin{array}{cc}
0 & 1 \\
0 & 0
\end{array}
\right]}_{\mt{A}}
\underbrace{\left[
\begin{array}{c}
p\\
\dot{p}
\end{array}
\right]}_{\vr{z}} + \underbrace{\left[
\begin{array}{c}
0\\
1
\end{array}
\right]}_{\mt{B}} v,
\label{eq:double_int_SS_scalar}
\end{equation*}
which we compactly write as
\begin{equation}
\vr{\dot{z}} = \mt{A} \; \vr{z} + \mt{B} \; v
\label{eq:double_int_SS_scalar_compact}
\end{equation}
In \vr{z}-space, the trajectory to be followed, and its open-loop controls, are
\begin{equation*}
\vr{z}_d(t) = \left( p_d(t), \dot{p}_d(t) \right),
\label{eq:desired_traj_z}
\end{equation*}
\begin{equation*}
v_d(t) = \ddot{p}_d(t),
\label{eq:desired_traj_v}
\end{equation*}
and note that they satisfy
\begin{equation}
\vr{\dot{z}}_d = \mt{A} \; \vr{z}_d + \mt{B} \; v_d.
\label{eq:difeq_desired}
\end{equation}
Then, the trajectory error is given by
\begin{equation*}
\vr{e}(t) = \vr{z}(t) - \vr{z}_d(t) = \left[
\begin{array}{c}
p(t) - p_d(t)\\
\dot{p}(t) - \dot{p}_d(t)
\end{array}
\right],
\label{eq:eq_error_inz}
\end{equation*}
so its time derivative is
\begin{equation}
\vr{\dot{e}}(t) = \vr{\dot{z}}(t) - \vr{\dot{z}}_d(t).
\label{eq:eq_derror_inz}
\end{equation}
By substituting Eqs.~\eqref{eq:double_int_SS_scalar_compact} and~\eqref{eq:difeq_desired} into Eq.~\eqref{eq:eq_derror_inz} we get
\begin{equation*}
\vr{\dot{e}} = \mt{A}\underbrace{(\vr{z} - \vr{z}_d)}_{\vr{e}} + \mt{B}\underbrace{(v - v_d)}_{\bar{v}},
\label{eq:difeq_error_inz}
\end{equation*}
which shows that the error dynamics is given by
\begin{equation}
\vr{\dot{e}} = \mt{A} \; \vr{e} + \mt{B} \; \bar{v}.
\label{eq:difeq_error_inz_compact}
\end{equation}
Thus, the stabilization of the system in Eq.~\eqref{eq:double_int_SS_scalar_compact} along $\vr{z}_d(t)$ reduces to stabilizing $\vr{e}(t)$ to the origin of $\mathbb{R}^2$. This can be achieved by using the feedback law
\begin{equation}
\bar{v} = - \mt{K} \; \vr{e},
\label{eq:feedback_law1}
\end{equation}
where
\begin{equation*}
\mt{K} = \left[ k_p, k_v \right]
\label{eq:feedback_law2}
\end{equation*}
In this law, $k_p$ and $k_v$ are called the position and velocity gains, and it is well-known that, if they are positive,
\begin{equation*}
\lim_{t\to\infty} \vr{e}(t) = 0
\label{eq:error_limit}
\end{equation*}
irrespective of the initial condition $\vr{e}(0)$. Therefore, this controller ensures a globally-stable tracking of $p_d(t)$, which is quite a desirable property. 

Observe that since \mbox{$\bar{v} = v - v_d = v - \ddot{p}_d$}, Eq.~\eqref{eq:feedback_law1} can be expressed as
\begin{equation}
v = \ddot{p}_d - k_p \left(p - p_d \right) - k_v \left(\dot{p} - \dot{p}_d \right),
\label{eq:controller_law}
\end{equation}
which is the usual form assumed for a PD controller in trajectory tracking. By particularizing Eq.~\eqref{eq:controller_law} for each one of the three coordinates in \mbox{$\vr{p} = (x, y, \alpha)$}, we see that the control law needed to stabilize the system in Eq.~\eqref{eq:eq_u2} along $\vr{p}_d(t)$ is
\begin{align}
\begin{split}		
\vr{v} = \underbrace{\left[
\begin{array}{c}
\ddot{x}_d\\
\ddot{y}_d\\
\ddot{\alpha}_d
\end{array}
\right]}_{\vr{\ddot{p}}_d}
& - 
\underbrace{\left[
\begin{array}{ccc}
k_{p,x} & 0 & 0\\
0 & k_{p,y} & 0\\
0 & 0 & k_{p,\alpha} 
\end{array}
\right]}_{\mt{K}_p} \underbrace{\left[
\begin{array}{c}
x - x_d\\
y - y_d\\
\alpha - \alpha_d
\end{array}
\right]}_{\vr{p} - \vr{p}_d} \\
& - 
\underbrace{\left[
\begin{array}{ccc}
k_{v,x} & 0 & 0\\
0 & k_{v,y} & 0\\
0 & 0 & k_{v,\alpha} 
\end{array}
\right]}_{\mt{K}_v} \underbrace{\left[
\begin{array}{c}
\dot{x} - \dot{x}_d\\
\dot{y} - \dot{y}_d\\
\dot{\alpha} - \dot{\alpha}_d
\end{array}
\right]}_{\vr{\dot{p}} - \vr{\dot{p}}_d}
\label{eq:controller_law_forfullp}
\end{split}
\end{align}
where $k_{p,x}$, $k_{p,y}$, $k_{p,\alpha}$ and $k_{v,x}$, $k_{v,y}$, $k_{v,\alpha}$ are the position and velocity gains for $x$, $y$ and $\alpha$.

If we write Eq.~\eqref{eq:controller_law_forfullp} as
\begin{equation*}
\vr{v} = \vr{\ddot{p}}_d - \mt{K}_p \left(\vr{p} - \vr{p}_d \right) - \mt{K}_v \left(\vr{\dot{p}} - \vr{\dot{p}}_d \right)
\label{eq:controller_law_forfullp_compact}
\end{equation*}
and substitute it in Eq.~\eqref{eq:eq_feedback_law}, we finally obtain
\begin{equation}
\vr{u} = \bar{\mt{M}} \left[ \vr{\ddot{p}}_d - \mt{K}_p \left( \vr{p} - \vr{p}_d \right) - \mt{K}_v \left( \vr{\dot{p}} - \vr{\dot{p}}_d \right) \right] + \bar{\mt{C}} \; \vr{\dot{p}}
\label{eq:chap5full_CTC}
\end{equation}
which is the desired computed-torque law for Otbot. This law can also be written as
\begin{equation}
\vr{u} = \underbrace{\bar{\mt{M}} \; \vr{\ddot{p}}_d + \bar{\mt{C}} \; \vr{\dot{p}}}_{\vr{u}_{traj}} + \underbrace{\bar{\mt{M}} \left[ - \mt{K}_p \left( \vr{p} - \vr{p}_d \right) - \mt{K}_v \left( \vr{\dot{p}} - \vr{\dot{p}}_d \right) \right]}_{\vr{u}_{corr}},
\label{eq:full_CTC_brokendown}
\end{equation}
where $\vr{u}_{traj}$ is the nominal torque needed to follow $\vr{p}_d(t)$ when we are on track along this trajectory, and $\mt{u}_{corr}(t)$ is the correction torque used to compensate position and velocity errors.

Note that to control the robot using this law we need feedback of the $\vr{p}$ variables $x$, $y$, and $\alpha$, as well as the angle $\varphi_p$ of the pivot joint. The $\vr{p}$ variables are needed to compute the position and velocity errors, and the pivot joint angle is used together with $\alpha$ to evaluate $\bar{\mt{M}}$ and $\bar{\mt{C}}$. Such a feedback can be obtained from the onboard IMU and the angular encoder of the pivot joint.

\subsection{Tuning of the control law}
\label{sec:TunControlLaw}

Once the controller has been defined, we must set appropiate values for $k_p$ and $k_v$ in each instance of Eq.~\eqref{eq:controller_law}. To this end we substitute Eq.~\eqref{eq:feedback_law1} into Eq.~\eqref{eq:difeq_error_inz_compact} to get
\begin{equation*}
\vr{\dot{e}} = \underbrace{\left( \mt{A} - \mt{B} \mt{K} \right)}_{\mt{C}} \vr{e},
\label{eq:eq_CTC_24}
\end{equation*}
which is the autonomous ODE describing the error of the closed-loop system for the $p$ variable in consideration. From the theory of linear systems we know that the solution of this ODE is
\begin{equation}
\vr{e}(t) = \underbrace{e^{s_1 t} c_1 \vr{v}_1}_{\vr{e}_1(t)} + \underbrace{e^{s_2 t} c_2 \vr{v}_2}_{\vr{e}_2(t)},
\label{eq:eq_CTC_25}
\end{equation}
where $s_1$ and $s_2$ are the eigenvalues of \mbox{$\mt{C} = \mt{A} - \mt{B}\mt{K}$}, $\vr{v}_1$ and $\vr{v}_2$ are their associate eigenvectors, and $c_1$ and $c_2$ are the constants that satisfy
\begin{equation*}
\left[
\begin{array}{cc}
\vr{v}_1 & \vr{v}_2
\end{array}
\right]
\left[
\begin{array}{c}
c_1\\
c_2
\end{array}
\right] = \vr{e}(0).
\label{eq:eq_CTC_26}
\end{equation*}
To tune the controller, we will select values for $s_1$ and $s_2$ that ensure an overdamped exponential decay of $\vr{e}(t)$, and then we will find the gains $k_p$ and $k_v$ that correspond to such values. These gains can be written as a function of $s_1$ and $s_2$ by noting that $s_1$ and $s_2$ are the roots of
\begin{equation}
\mathrm{det} \left( \mt{C} - \lambda \mt{I} \right) = \lambda^2 + k_v \lambda + k_p,
\label{eq:eq_CTC_27}
\end{equation}
or equivalently of
\begin{equation}
\left( \lambda - s_1 \right) \left( \lambda - s_2 \right) = \lambda^2 - \lambda \left( s_1 + s_2 \right) + s_1 \; s_2,
\label{eq:eq_CTC_28}
\end{equation}
so identifying the right-hand sides of~\eqref{eq:eq_CTC_27} and~\eqref{eq:eq_CTC_28} we obtain
\begin{align*}
k_p &= s_1 \; s_2,\\
k_v &= - \left( s_1 + s_2 \right).
\end{align*}
To force an overdamped decay of $\vr{e}(t)$ towards zero, $s_1$ and $s_2$ must be negative real numbers. At this point it is customary to choose
\begin{equation*}
s_1 = - \frac{4}{T_{stab}},
\label{eq:eq_CTC_31}
\end{equation*}
where $T_{stab}$ is the time we allow for $\vr{e}(t)$ to be smaller than $2\% \; \vr{e}(0)$, and
\begin{equation*}
s_2 = 10 \; s_1,
\label{eq:eq_CTC_32}
\end{equation*}
so in Eq.~\eqref{eq:eq_CTC_25} $\vr{e}_2(t)$ decays much more rapidly than $\vr{e}_1(t)$. In doing so we have
\begin{equation*}
\vr{e}(t) = e^{-\frac{4}{T_{stab}} t} c_1 \vr{v}_1 + e^{-\frac{40}{T_{stab}} t} c_2 \vr{v}_2,
\label{eq:eq_CTC_33}
\end{equation*}
so for $t = T_{stab}$ we obtain
\begin{equation*}
\vr{e}(T_{stab}) = \underbrace{e^{-4}}_{0,018} c_1 \vr{v}_1 + \underbrace{e^{-40}}_{\simeq 0} c_2 \vr{v}_2,
\label{eq:eq_CTC_34}
\end{equation*}
which ensures that \mbox{$\vr{e}(T_{stab}) \leqslant 2\% \; \vr{e}(0)$} as desired. Note that $\vr{e}(t)$ contains both the position and velocity errors, so both errors will almost disappear after $T_{stab}$ seconds.

In Otbot we have set $T_{stab} = 3 \mathrm{s}$ for all variables \mbox{$p \in \{ x, y, \alpha \}$}, which yields
\begin{align*}
s_1 &= - \frac{4}{3} = -1.333, \\
s_2 &= 10 \; s_1 = -13.333,
\end{align*}
and,
\begin{align*}
k_p &= s_1 \; s_2 = 17.778, \\
k_v &= - \left( s_1 + s_2 \right) = 14.667. 
\end{align*}
Therefore, the gains we have used in Eq.~\eqref{eq:controller_law_forfullp} are
\begin{equation*}
\mt{K}_p = \left[
\begin{array}{ccc}
17.778 & 0 & 0\\
0 & 17.778 & 0\\
0 & 0 & 17.778
\end{array}
\right], \; \; \mt{K}_v = \left[
\begin{array}{ccc}
14.667 & 0 & 0 \\
0 & 14.667 & 0 \\
0 & 0 & 14.667
\end{array}
\right].
\label{eq:eq_CTC_37}
\end{equation*}
Note however that it is possible to choose different stabilitzation times $T_{stab}$ for each variable if desired, which would yield different gains along the diagonals of $\mt{K}_p$ and $\mt{K}_v$.

\subsection{Checking the feasibility of the control torques}
\label{sec:checkfeastorque}

Once the gains $\mt{K}_p$ and $\mt{K}_v$ have been decided, it is important to check that the control inputs given by Eq.~\eqref{eq:full_CTC_brokendown},
\begin{equation*}
\vr{u} = \underbrace{\bar{\mt{M}} \; \vr{\ddot{p}}_d + \bar{\mt{C}} \; \vr{\dot{p}}}_{\vr{u}_{traj}} + \underbrace{\bar{\mt{M}} \left[ - \mt{K}_p \left( \vr{p} - \vr{p}_d \right) - \mt{K}_v \left( \vr{\dot{p}} - \vr{\dot{p}}_d \right) \right]}_{\vr{u}_{corr}},
\label{eq:full_CTC_brokendown2}
\end{equation*}
stay within the allowable torque limits of the motors along $\vr{p}_d(t)$, otherwise the global stability properties of the controller may be compromised.

The exact value of $\vr{u}_{traj}(t)$ and $\vr{u}_{corr}(t)$ cannot be known a priori, as they depend on unmodelled perturbations that occur during the actual operation of the robot. However, we may find reasonable bounds for them if we know the intervals of variation for \mbox{$\vr{e}_p = \vr{p} - \vr{p}_d$} and \mbox{$\vr{e}_v = \vr{\dot{p}} - \vr{\dot{p}}_d$}. These bounds can be computed as follows.

We first simulate the desired trajectory $\vr{p}_d(t)$ in open loop, using \mbox{$\vr{u} = \vr{u}_{traj}(t)$} as the control input. This will provide the robot state \mbox{$\vr{x}(t) = \left( \vr{q}(t), \vr{\dot{q}}(t) \right)$} for each time $t$ along $\vr{p}_d(t)$, so $\vr{\dot{p}}(t)$, $\bar{\mt{M}}(t)$ and $\bar{\mt{C}}(t)$ will also be known. The real trajectory will be slightly different, but we can assume that the actual values $\vr{\dot{p}}(t)$, $\bar{\mt{M}}(t)$ and $\bar{\mt{C}}(t)$ stay close to those on $\vr{p}_d(t)$.

For each time $t$ we then evaluate the range of possible values taken by $\vr{u}_{corr}(t)$ and $\vr{u}_{traj}(t)$ using interval arithmetics~\cite{Jaulin_2001}. Since $\bar{\mt{M}}$, $\mt{K}_p$ and $\mt{K}_v$ are all known at time $t$, the expression of $\vr{u}_{corr}(t)$ is linear in $\vr{e}_p$ and $\vr{e}_v$, and it is easy to obtain an interval bound for $\vr{u}_{corr}(t)$ using sums of intervals
\begin{equation*}
[x_{min},x_{max}] + [y_{min}, y_{max}] = [x_{min} + y_{min}, x_{max} + y_{max}] , 
\label{sumofint}
\end{equation*}
and multiplications of an interval by a constant
\begin{equation*}
k \cdot [x_{min}, x_{max}] = 
\left\lbrace
\begin{array}{l}
[k \cdot x_{max}, k \cdot x_{min}] \; \; for \; k < 0 \\
\left[k \cdot x_{min}, k \cdot x_{max}\right] \; \; for \; k \geq 0 
\end{array}
\right. 
.
\label{eq:multofint}  
\end{equation*}
Regarding $\vr{u}_{traj}(t)$, since $\vr{e}_v$ takes values within a prescribed interval and $\vr{\dot{p}}_d(t)$ is known, the interval of variation of $\vr{\dot{p}}(t)$ will also be known, which allows the interval evaluation of $\bar{\mt{C}} \; \vr{\dot{p}}$ using the previous operations. The resulting intervals must be shifted using the scalar quantities of \mbox{$\bar{\mt{M}}(t) \; \vr{\ddot{p}}_d(t)$} to obtain $\vr{u}_{traj}(t)$.

In the end, we only have to add the intervals obtained for $\vr{u}_{traj}(t)$ and $\vr{u}_{corr}(t)$ for each time $t$, and check whether the result stays within the allowable torque limits of the robot.
\begin{figure}[b!]
	\centering
	\includegraphics[width=\linewidth]{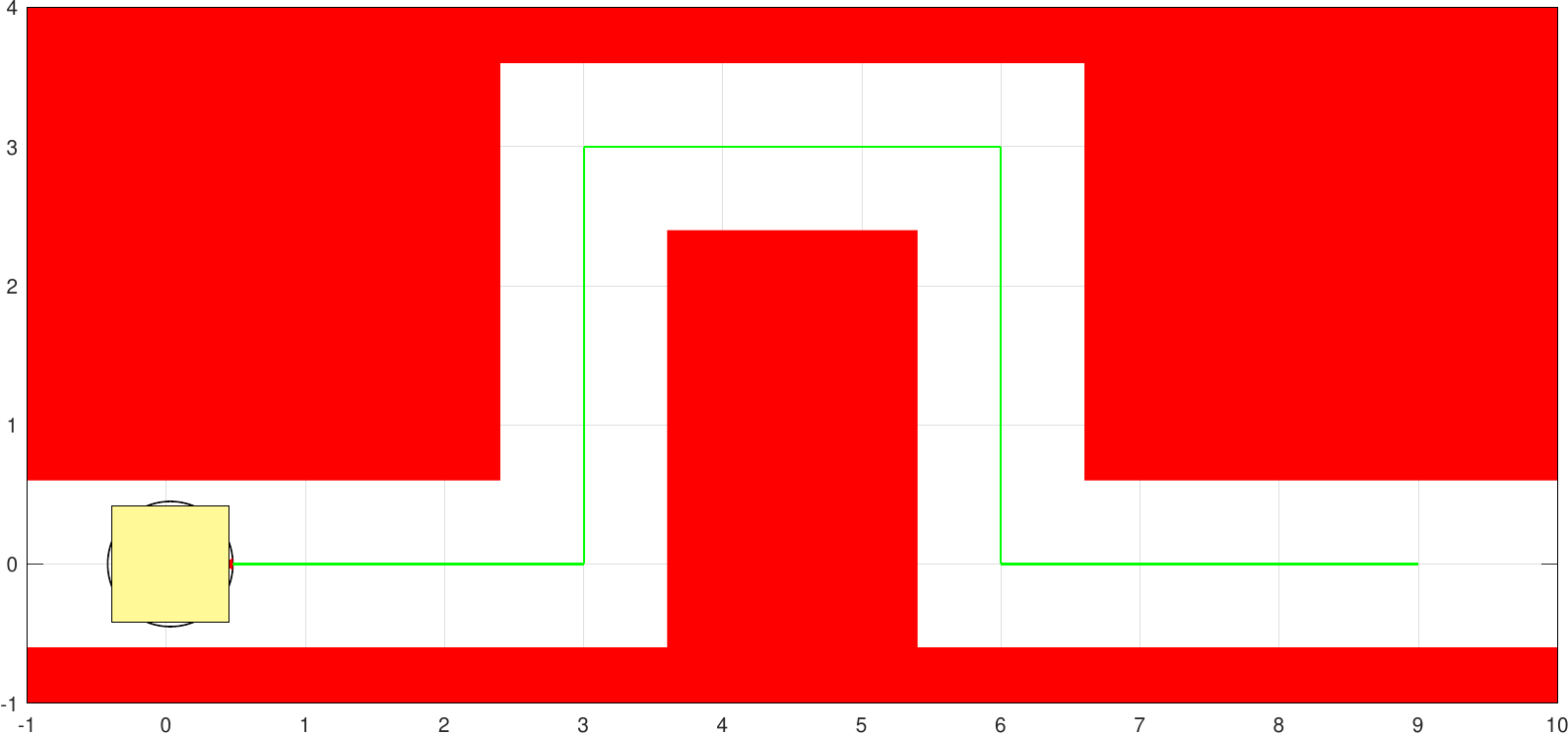}
	\caption{\label{fig:NarrCorrExperiment} Narrow corridor experiment. The vehicle has to transport the yellow load at a constant orientation with a constant speed of $0.6\mathrm{m/s}$ along the green path. See \url{https://youtu.be/dhKDoDd0wZo} for an animated version of the motion. The movements underlying the chassis can be appreciated in \url{https://youtu.be/c62ks498Z2U}.} 
\end{figure}

\subsection{Test cases}
\label{sec:testcases}

We next explain several simulations that we have done to verify the obtained control law. We start by doing some experiments in order to test the omnidirectionality of the robot (Section~\ref{subsec:omnitest}). Some experiments are then performed to verify that the controller has the ability to stabilize trajectories that are the result of a motion planner (Section~\ref{subsec:trajtrack}). Finally, we verify the effectiveness of the control law by simulating the system under unmodelled force perturbations (Section~\ref{subsec:distreject}). In all experiments we have used the same robot parameters assumed in Section~\ref{ch:chap4ParamIdent}. Viscous friction has been neglected, as it does not bring anything illustrative to our tests.

\begin{figure}[tb]
	\centering
	\begin{subfigure}[b]{\textwidth}
		\centering
		\includegraphics[width=\textwidth]{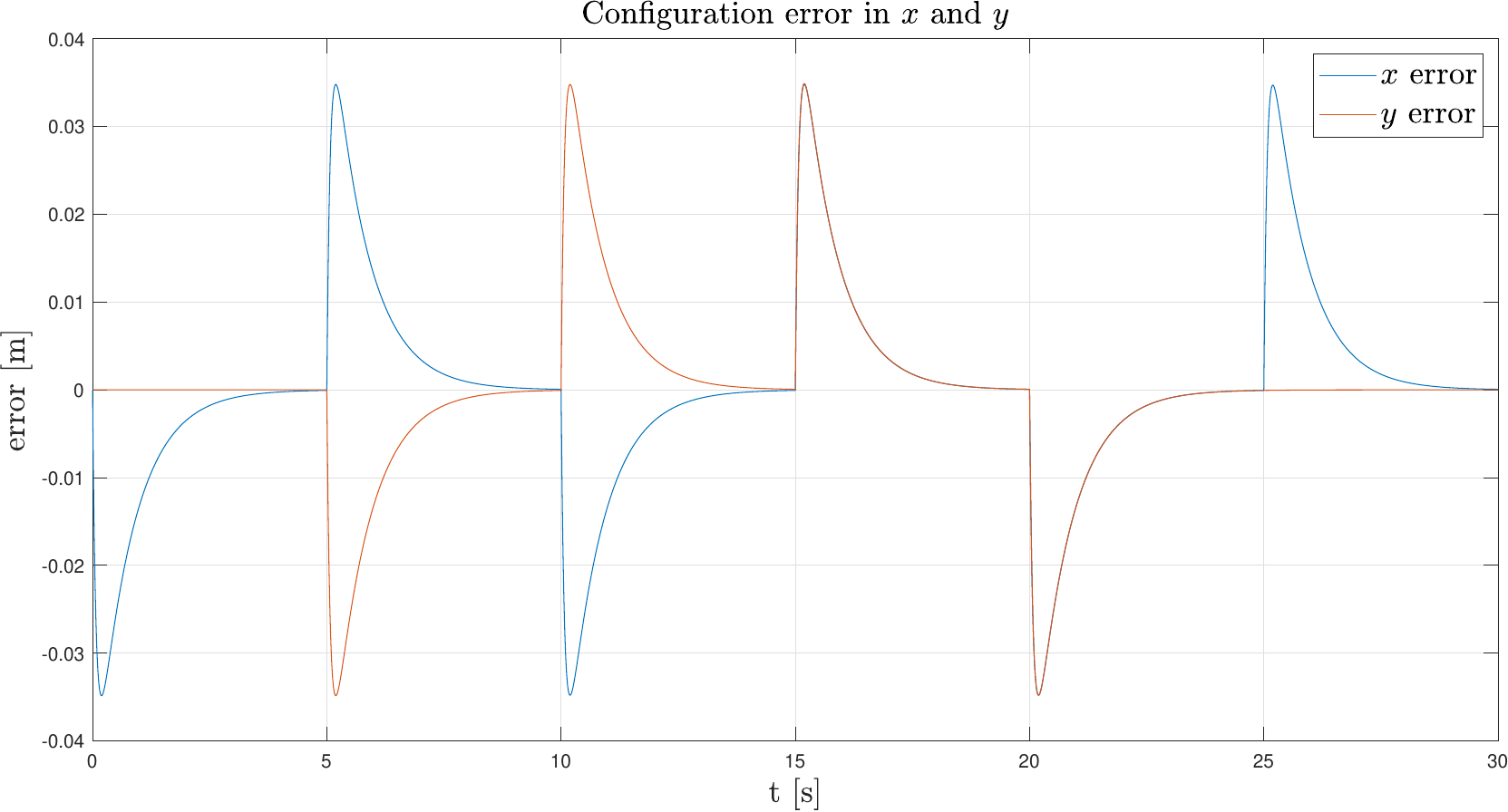}
		\label{fig:XYerrorovertime}
	\end{subfigure}
	
	
	\begin{subfigure}[b]{\textwidth}
		\centering
		\includegraphics[width=\textwidth]{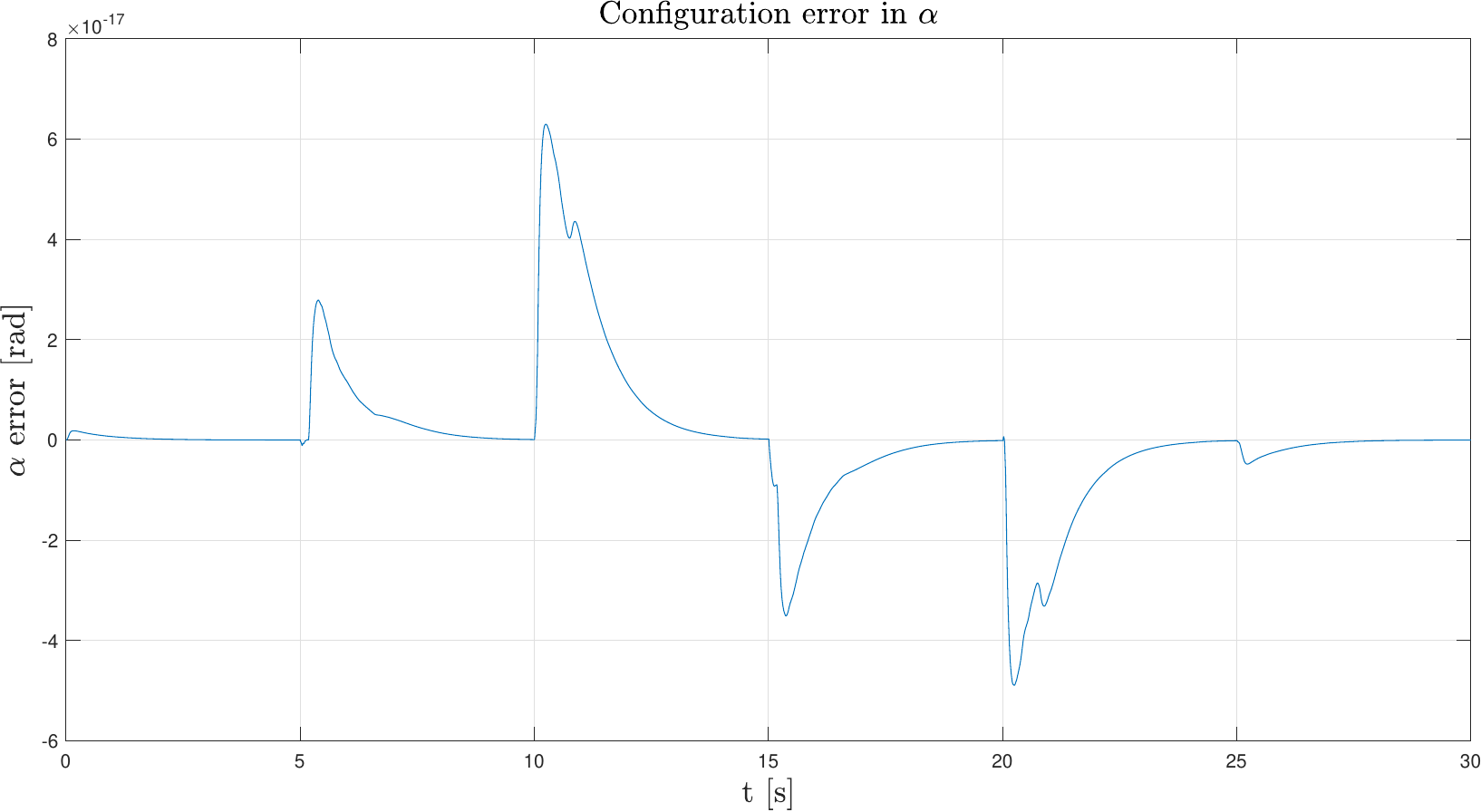}
		\label{fig:Alphaerrorovertime}
	\end{subfigure}
	
	\caption{Configuration errors in $x$, $y$ and $\alpha$ in the narrow corridor example.}
	\label{fig:NarrCorr10}
\end{figure}

\begin{figure}[tb]
	\centering
	\begin{subfigure}[b]{\textwidth}
		\centering
		\includegraphics[width=\textwidth]{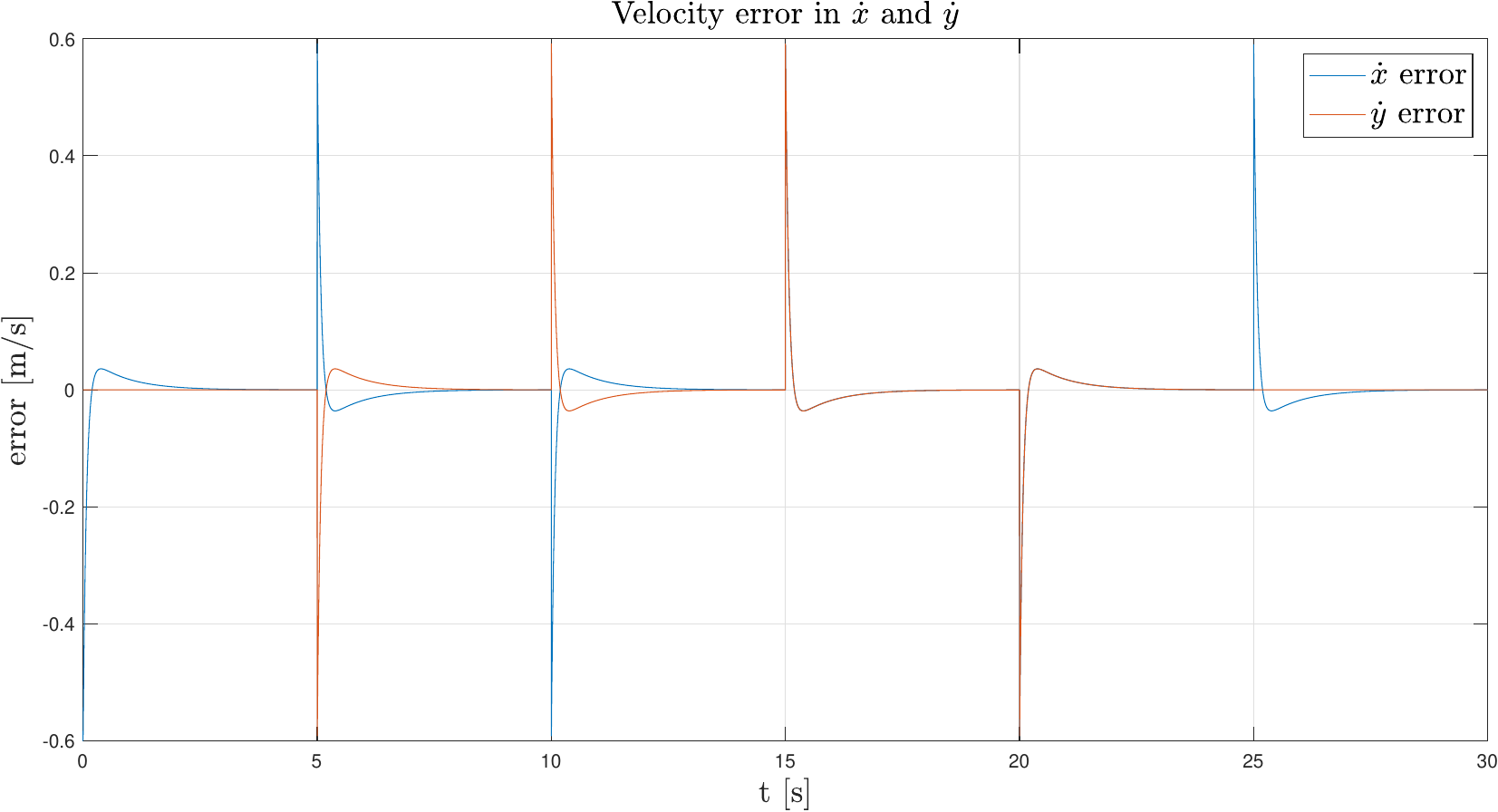}
		\label{fig:dXdYerrorovertime}
	\end{subfigure}
	
	
	\begin{subfigure}[b]{\textwidth}
		\centering
		\includegraphics[width=\textwidth]{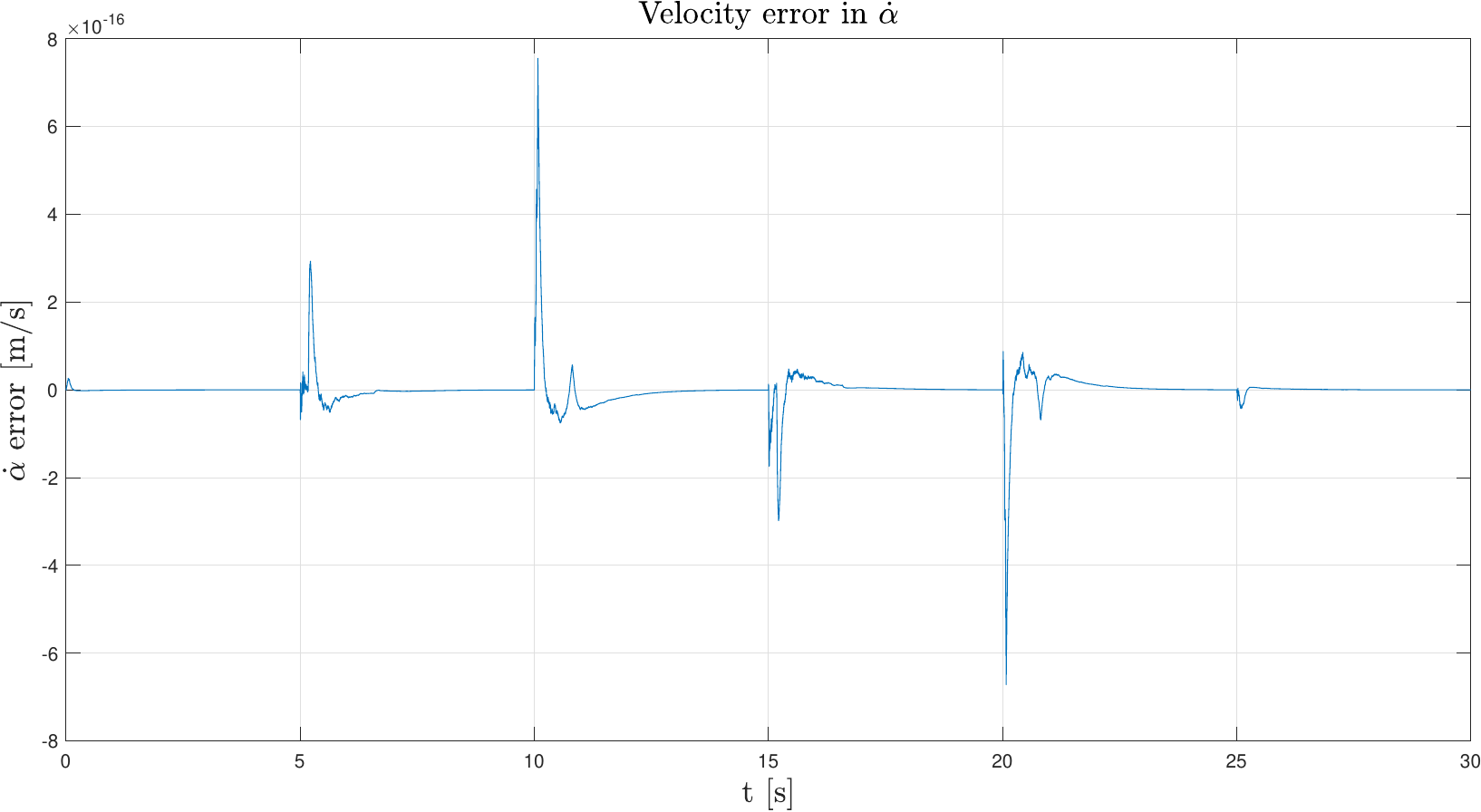}
		\label{fig:dAlphaerrorovertime}
	\end{subfigure}
	
	\caption{Velocity errors in $\dot{x}$, $\dot{y}$ and $\dot{\alpha}$ in the narrow corridor example.}
	\label{fig:NarrCorr20}
\end{figure}

\subsubsection{Narrow corridor test}
\label{subsec:omnitest}

As we explained in Section~\ref{ch:introduction}, one of the main advantages of Otbot is that its platform has omnidirectional mobility in the plane. This is a very desirable property that allows a simpler negotiation of obstacles and narrow corridors, or the correction of small errors without having to maneuver. To test our controller in one of these situations we have set Otbot to transport a load along a narrow corridor with $90^{\circ}$ turns while keeping $\alpha = 0^{\circ}$ (see Fig.~\ref{fig:NarrCorrExperiment} and its video animations). Notice that such a movement would be impossible for a vehicle with, e.g., Ackerman kinematics, or for a standard differential drive robot. Even if the load were allowed to rotate along the path, the amount of maneuvering would be considerable at the corners of the corridor.

In this test, the path to be followed consists of five rectilinear segments of $3\mathrm{m}$ in length, which have to be traversed at a constant velocity of $0.6 \mathrm{m}/\mathrm{s}$. Thus, each segment has to be covered in $5$ seconds by the robot, which is assumed to be at rest at the beginning. The tracking errors obtained in this example are given in Figs.~\ref{fig:NarrCorr10} and~\ref{fig:NarrCorr20}.

Notice that, because the robot is initially at rest, errors in both $x$ and $\dot{x}$ appear initially. However, the robot is able to counteract them in $T_{stab} = 3\mathrm{s}$ as we assumed in Section~\ref{sec:TunControlLaw}. This point is easier to check in the error plot for $\dot{x}$ (Fig.~\ref{fig:NarrCorr20}, top).

Note also that errors in $x$, $y$, $\dot{x}$, and $\dot{y}$ arise at each $90^{\circ}$ corner of the path because of the sudden change in direction of the reference velocity $(x_d,y_d)$. Again, the robot takes about $3\mathrm{s}$ to eliminate this error.

From the plots we see that the error in the platform angle $\alpha$ is negligible, so the platform stays with $\alpha = 0^{\circ}$ during the motion. This is in agreement with the fact that the robot departs from $\alpha = 0^{\circ}$ for $t=0\mathrm{s}$, and the chassis cannot transmit any torque to the platform because viscous friction is assumed to be negligible.

In the last $5$ trajectory seconds the reference signal for $x(t)$ and $y(t)$ is kept at
\begin{align*}
x_d(t) &= 9\mathrm{m}\\
y_d(t) &= 0\mathrm{m}
\end{align*}
which explains the positive error in $x$ that we observe, and its progressive elimination. Thus, the robot slightly overshoots the goal position $(9,0)\mathrm{m}$, and has to return backwards to reach it. The overshoot can be eliminated by simply designing a trajectory that smoothly achieves \mbox{$(\dot{x}(t),\dot{y}(t)) = (0,0)$} for \mbox{$t=25\mathrm{s}$}.


\subsubsection{Tracking the trajectory of a motion planner}
\label{subsec:trajtrack}

In this section we show another practical example of the usage of the controller. It is well known that one way to design robot trajectories is by means of a motion planner. Such a planner allows us to obtain trajectories that satisfy multiple constraints on the states and actions of the robot, like collision-avoidance, or velocity or torque-limit constraints. In this particular example we wish to stabilize a trajectory that was obtained with the motion planner in~\cite{gautier2021mscthesis}, which generates the trajectory by discretizing $x(t)$ and $u(t)$, and solving a constrained optimization problem. Because of the discretization, the planner returns a sequence of actions \mbox{$\vr{u}_0, \ldots, \vr{u}_N$}, and states \mbox{$\vr{x}_0, \ldots, \vr{x}_N$} that are not exactly compatible. That is, if we use \mbox{$\vr{u}_0, \ldots, \vr{u}_N$} to perform open-loop control, the robot will end up deviating from the desired trajectory, possibly colliding with some obstacle. Instead, if we use the controller to directly track the states \mbox{$\vr{x}_0, \ldots, \vr{x}_N$} in closed loop, we will see that the robot is able to accurately follow the desired trajectory.

\begin{figure}[t!]

	\begin{center}

		\includegraphics[width=\textwidth]{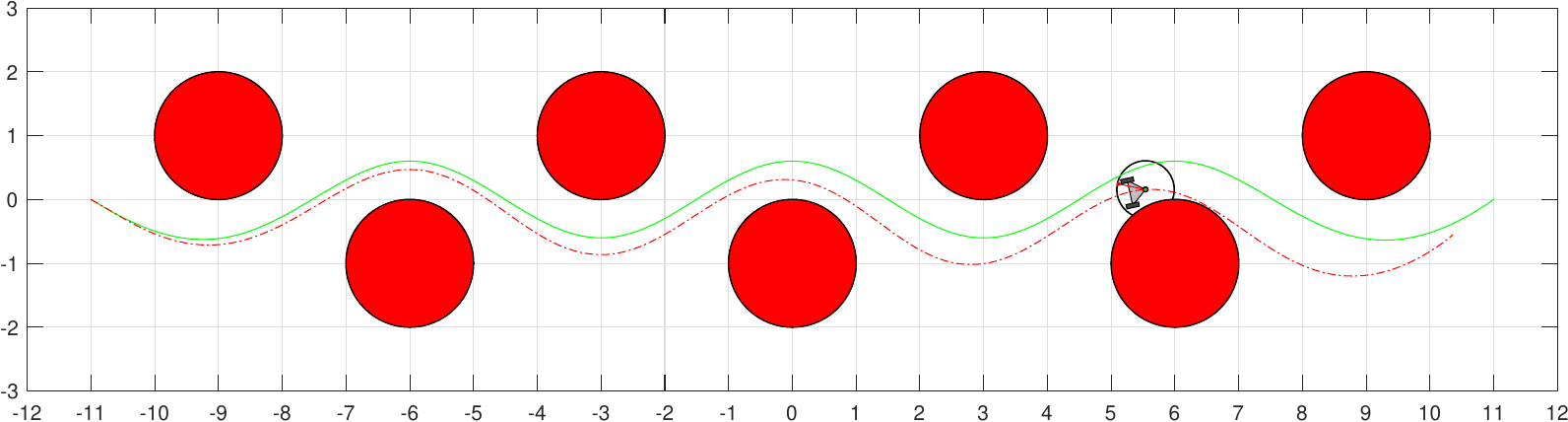}
			
		\vspace{4mm}

		\includegraphics[width=.49\textwidth]{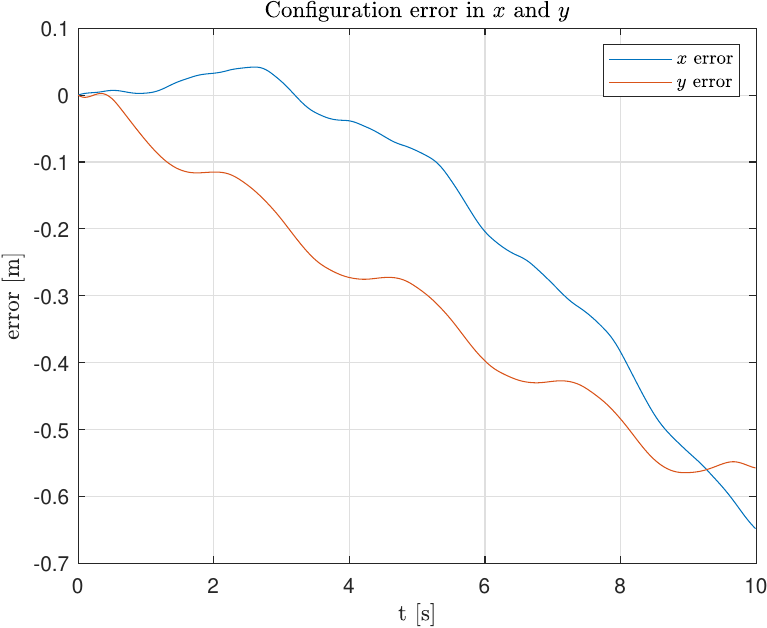}
		\includegraphics[width=.49\textwidth]{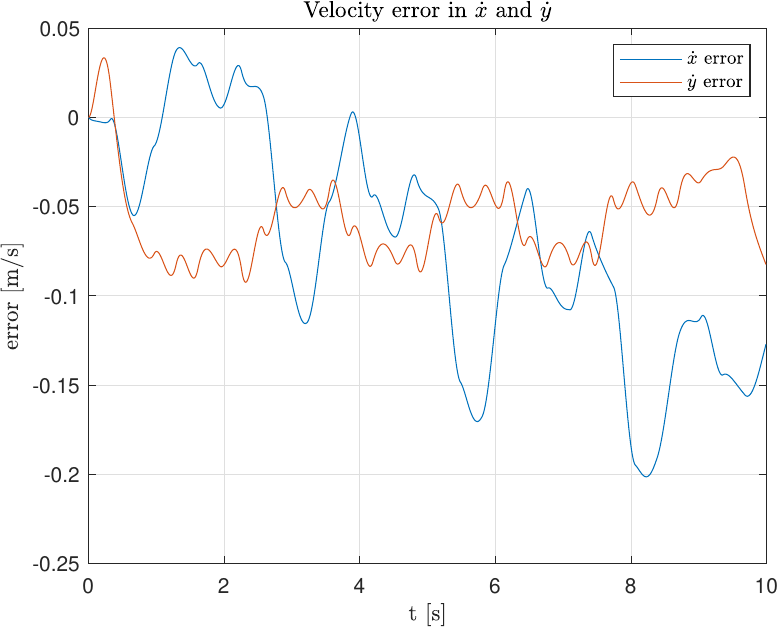}

		\vspace{4mm}
			
		\includegraphics[width=.49\textwidth]{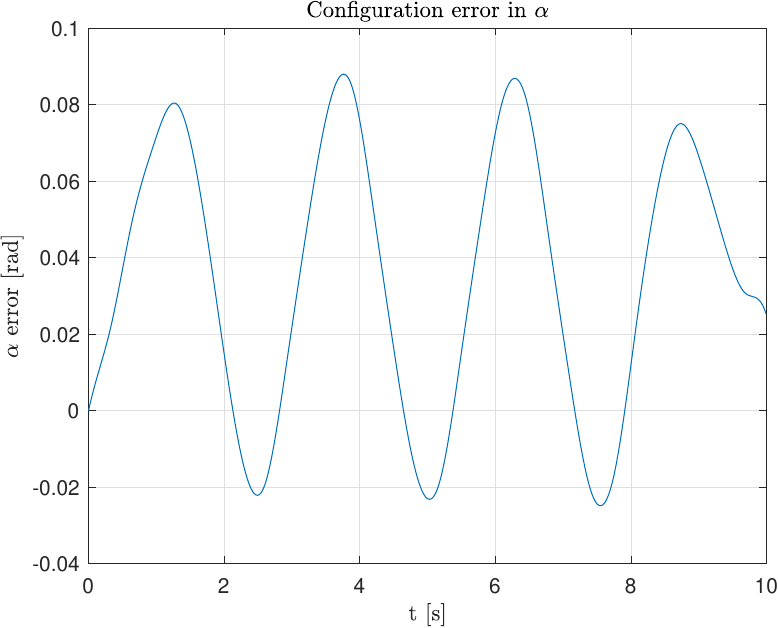}
		\includegraphics[width=.49\textwidth]{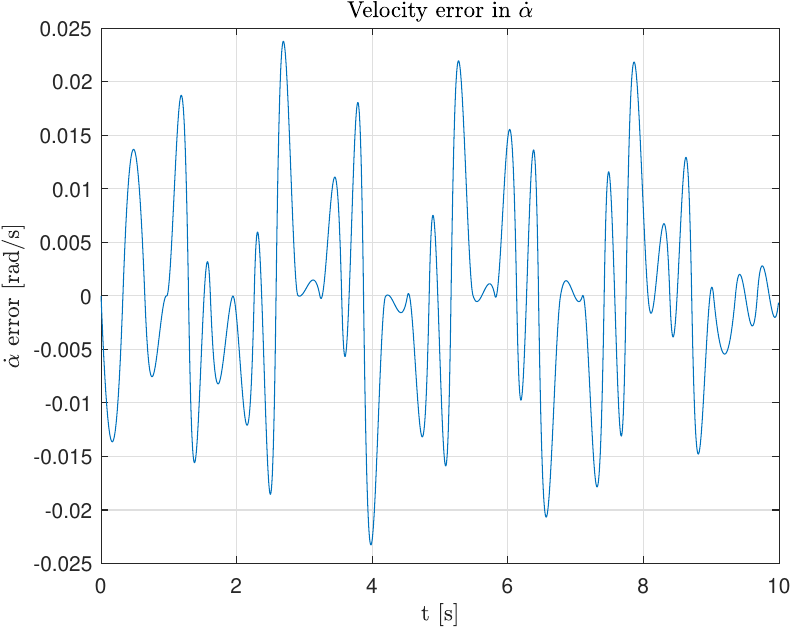}
	
	\end{center}
	
	\caption{Top: Simulation of the crowded corridor experiment in open-loop. The vehicle has to pass through the corridor without colliding with the circular obstacles (the axes' units are in $\mathrm{m}$). The desired and obtained trajectories are shown in green and red respectively. Bottom: The error plots of the motion. See \url{https://youtu.be/J3dnIHSabnA} for an animated version of the experiment.}
	\label{fig:CrowdCorrOL10}
\end{figure}

\begin{figure}[t!]
	
	\begin{center}
		
		\includegraphics[width=\textwidth]{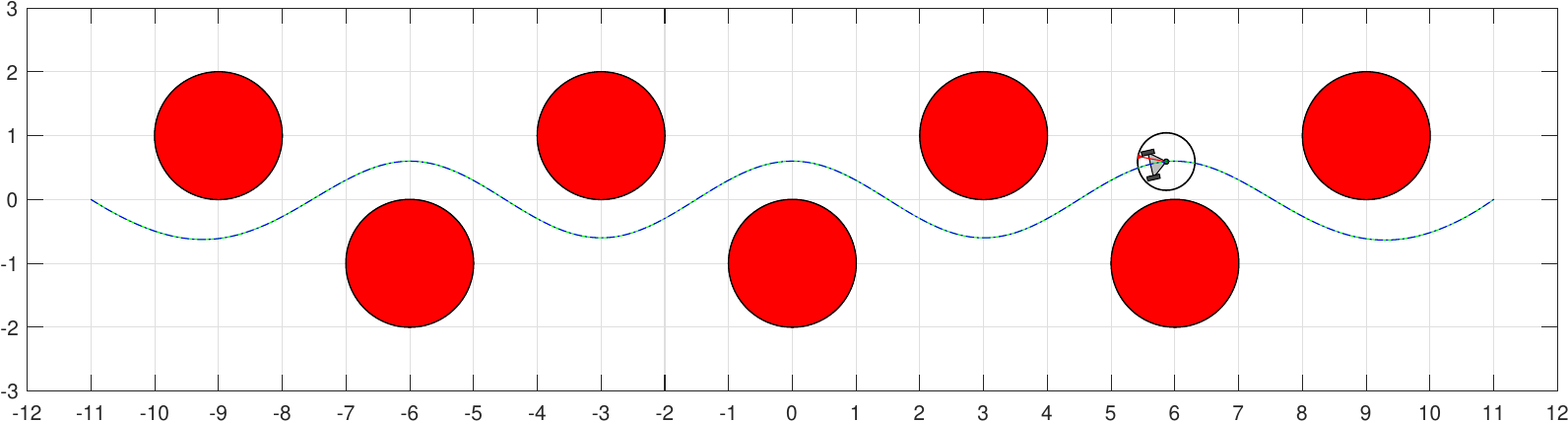}
		
		\vspace{4mm}
					
		\includegraphics[width=.49\textwidth]{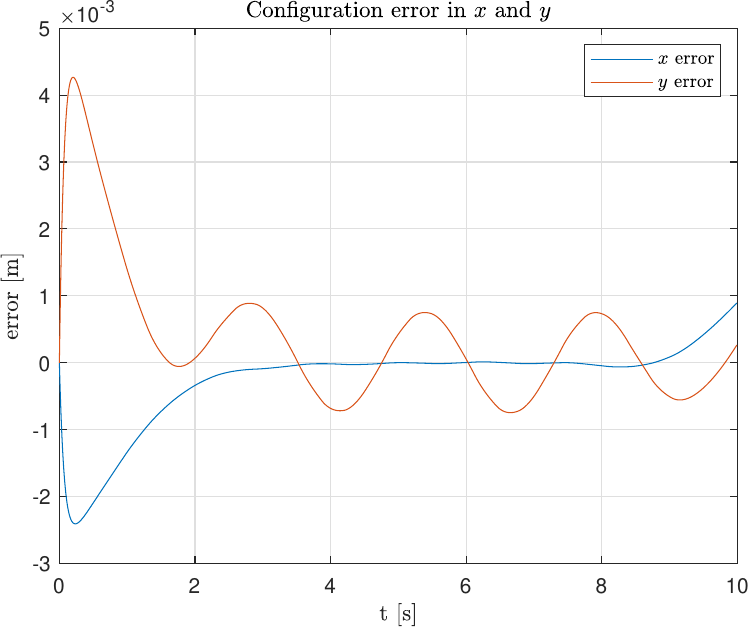}
		\includegraphics[width=.49\textwidth]{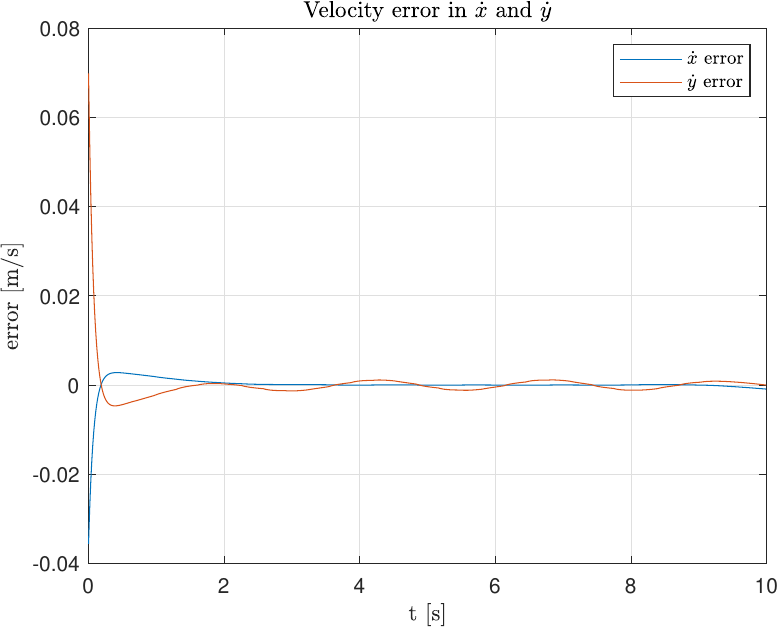}
		
	    \vspace{4mm}
		
		\includegraphics[width=.49\textwidth]{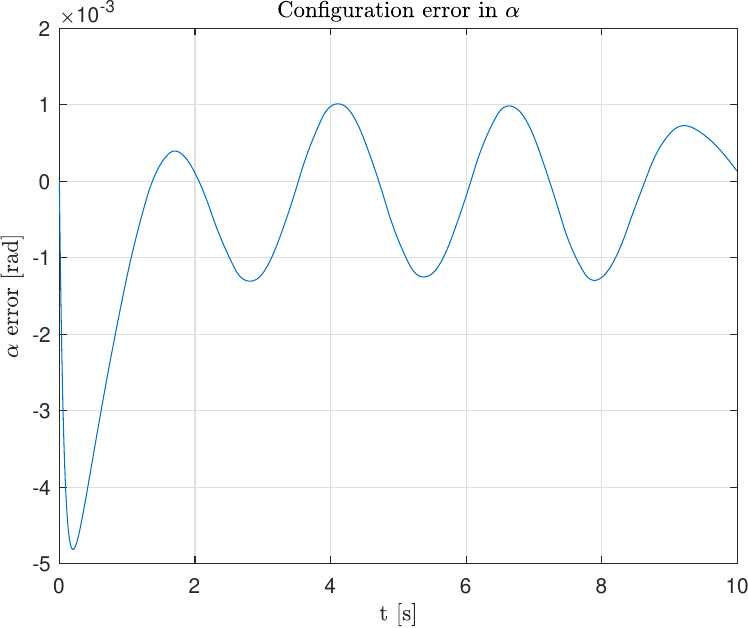}
		\includegraphics[width=.49\textwidth]{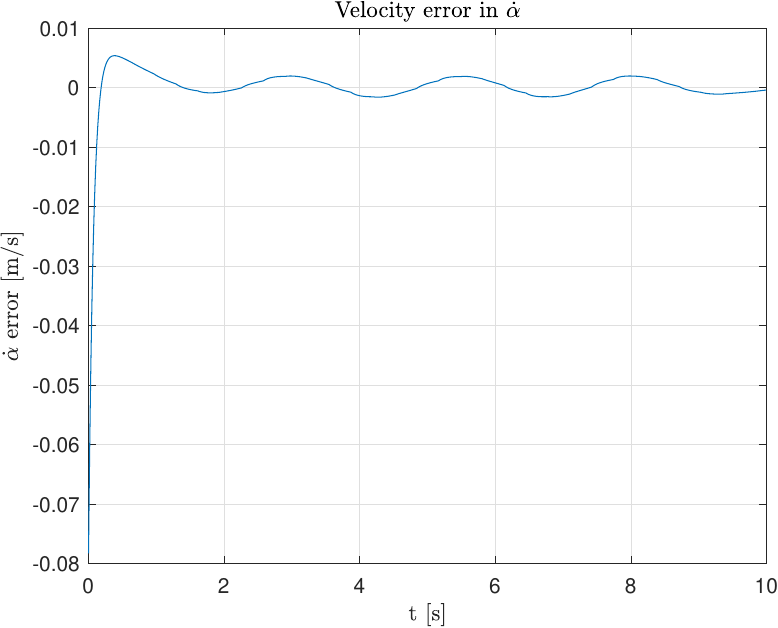}
		
	\end{center}
	
	\caption{Top: Simulation of the crowded corridor experiment in closed-loop. The desired and obtained trajectories now coincide, and the robot is able to avoid the obstacles. Bottom: The error plots of the motion. See \url{https://youtu.be/x5mmnUQ9x50} for an animated version of the experiment.}
	\label{fig:CrowdCorrCL10}
\end{figure}

The planned trajectory we wish to stabilize is shown in green in Fig.~\ref{fig:CrowdCorrOL10} top (units in $\mathrm{m}$). Here, the robot is set to traverse a narrow corridor that is crowded with circular obstacles. The figure also shows as a dashed line the trajectory that results from controlling the robot in open loop using \mbox{$\vr{u}_0, \ldots, \vr{u}_N$}. From the associate video we note that the robot deviates from the green trajectory and collides with two obstacles. The rest of Fig.~\ref{fig:CrowdCorrOL10} provides the position and velocity errors obtained during the motion, which are significant, and even cumulative in $x$ and $y$. In contrast, Fig.~\ref{fig:CrowdCorrCL10} shows how, when tracking \mbox{$\vr{x}_0, \ldots, \vr{x}_N$} in closed loop, the robot can accurately follow the desired trajectory. Position and velocity errors are smaller than $1\mathrm{mm}$ and $1\mathrm{cm}/\mathrm{s}$ most of the time, while those in $\alpha$ and $\dot{\alpha}$ are below $0.001 \mathrm{rad}$ and $0.005 \mathrm{rad}/\mathrm{s}$ in general. The errors we obtain are not zero exactly because the reference trajectories for $\dot{x}$, $\dot{y}$, and $\dot{\alpha}$ are obtained by computing finite differences on the discrete trajectories for $x$, $y$, and $\alpha$. Thus, the position and velocity reference signals have small mutual inconsistencies.


\subsubsection{Disturbance rejection}
\label{subsec:distreject}

We next want to test the robustness of the control law to external force perturbations not included in the model. To do so, we will simulate the effect of applying a small external force $$\vr{f}_p = (f_{p,x}, f_{p,y})$$ on the pivot joint $P$ of the platform, at various time instants during the simulation horizon.

To simulate the effect of these forces, the generalized force corresponding to $\vr{f}_p$ must be added to the right-hand side of Eq.~\eqref{eq:eq_lagrange} This force, which we denote by $\mt{Q}_p$, is obtained analogously to how the generalized force of actuation was obtained (Section~\ref{subsec:GenFofaction}). Clearly, the virtual power generated by $\vr{f}_p$ is 
\begin{equation}
P_p = \left[
\begin{array}{cc}
f_{p,x} & f_{p,y}
\end{array}
\right] \; \left[
\begin{array}{c}
\dot{x}\\
\dot{y}
\end{array}
\right]
\label{eq:ch5powereq1}
\end{equation}
which, to identify $\vr{Q}_p$, must be written in the form
\begin{equation}
P_p = \mt{Q}_p\trans \; \vr{\dot{q}}.
\label{eq:ch5powereq2}
\end{equation} 
However, since
\begin{equation}
	\vr{\dot{q}} = \left[
	\begin{array}{c}
		\dot{x}\\
		\dot{y}\\
		\dot{\alpha}\\
		\dot{\varphi}_r\\
		\dot{\varphi}_l\\
		\dot{\varphi}_p\\
	\end{array}
	\right].
\end{equation}
we see that
\begin{equation*}
\mt{Q}_p = \left( f_{p,x}, f_{p,y}, 0, 0, 0, 0 \right).
\label{eq:ch5qp}
\end{equation*}

\begin{figure}[b!]
	\begin{center}
	\begin{subfigure}{.48\textwidth}
		\centering
		\includegraphics[width=\textwidth]{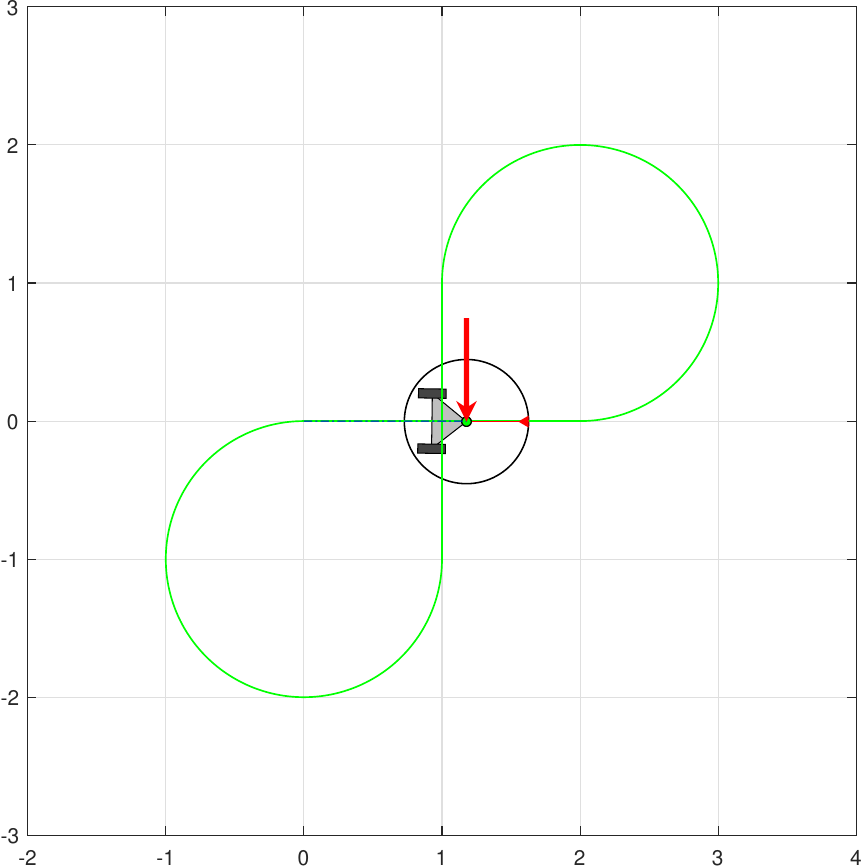}
		\subcaption{First disturbance applied at $3 \mathrm{s}$.}
		\label{fig:InfinityShapeDist1}
	\end{subfigure}		
	\begin{subfigure}{.48\textwidth}
		\centering
		\includegraphics[width=\textwidth]{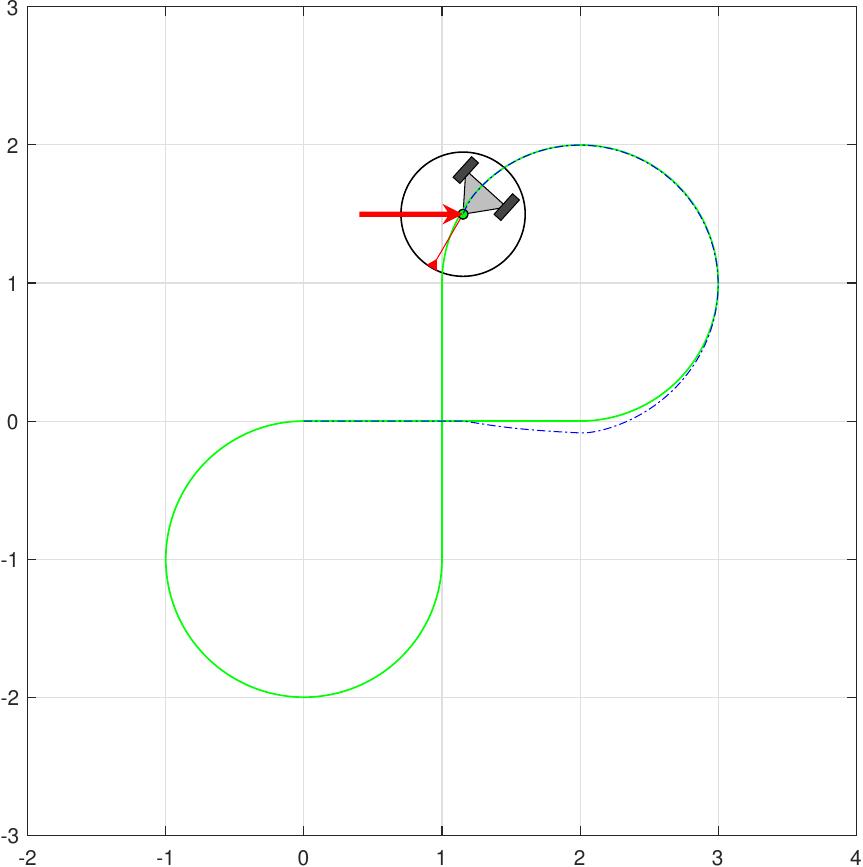}
		\subcaption{Second disturbance applied at $8 \mathrm{s}$.}
		\label{fig:InfinityShapeDist2}
	\end{subfigure}
	
	\vspace{5mm}
	
	\begin{subfigure}{.48\textwidth}
		\centering
		\includegraphics[width=\textwidth]{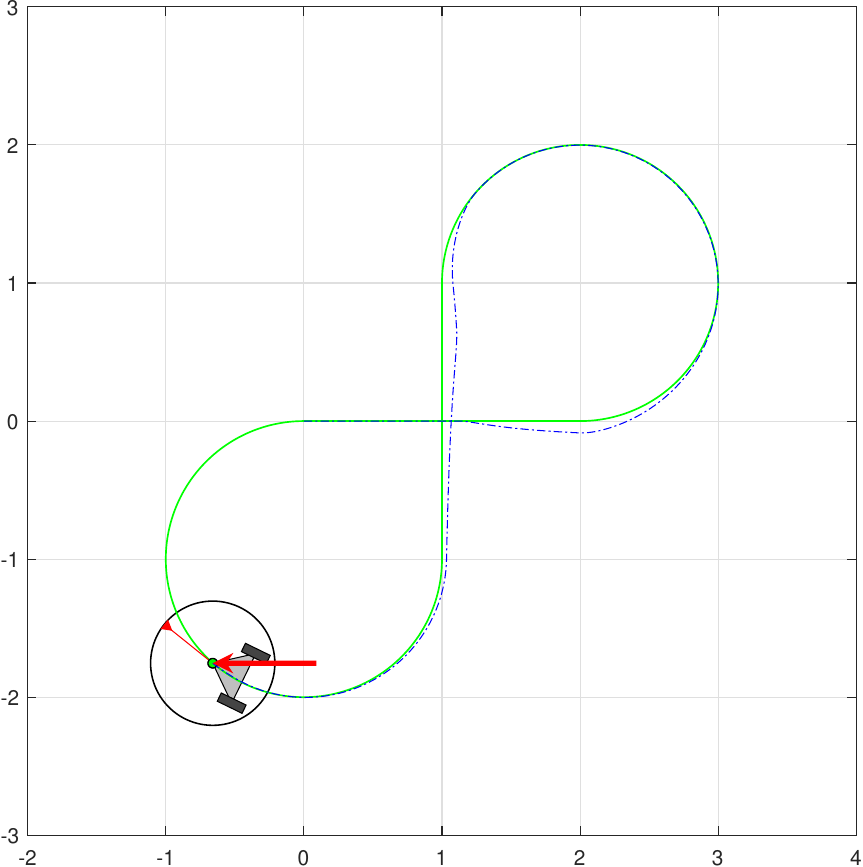}
		\subcaption{Third disturbance applied at $11 \mathrm{s}$.}
		\label{fig:InfinityShapeDist3}
	\end{subfigure}		
	\begin{subfigure}{.48\textwidth}
		\centering
		\includegraphics[width=\textwidth]{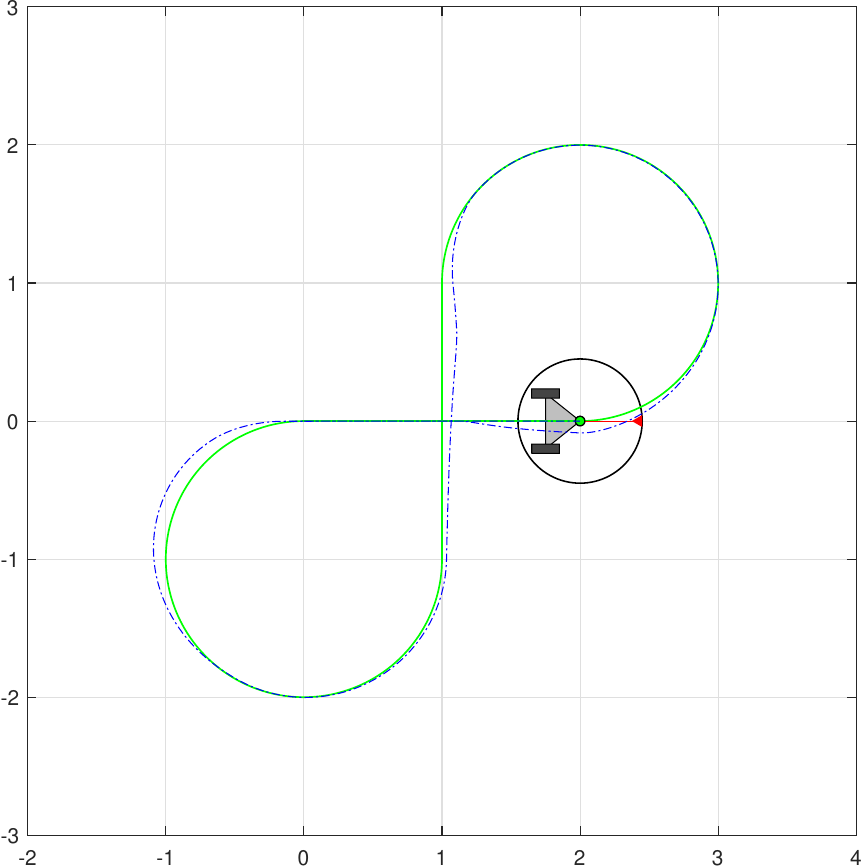}
		\subcaption{Otbot at the goal position.}
		\label{fig:InfinityShapeDistFp}
	\end{subfigure}
	\end{center}
		
	\caption{Disturbance rejection test. The vehicle has to follow the green path, but at certain time intervals we apply force disturbances which deviate the vehicle from the desired trajectory. The actual trajectory path is marked in blue. See \url{https://youtu.be/6bkYEmZqX_o} for an animated version of the motion.}
	\label{fig:InfShapeCLRute}
\end{figure}

\begin{figure}[t!]
	\begin{center}
		\includegraphics[width=\textwidth]{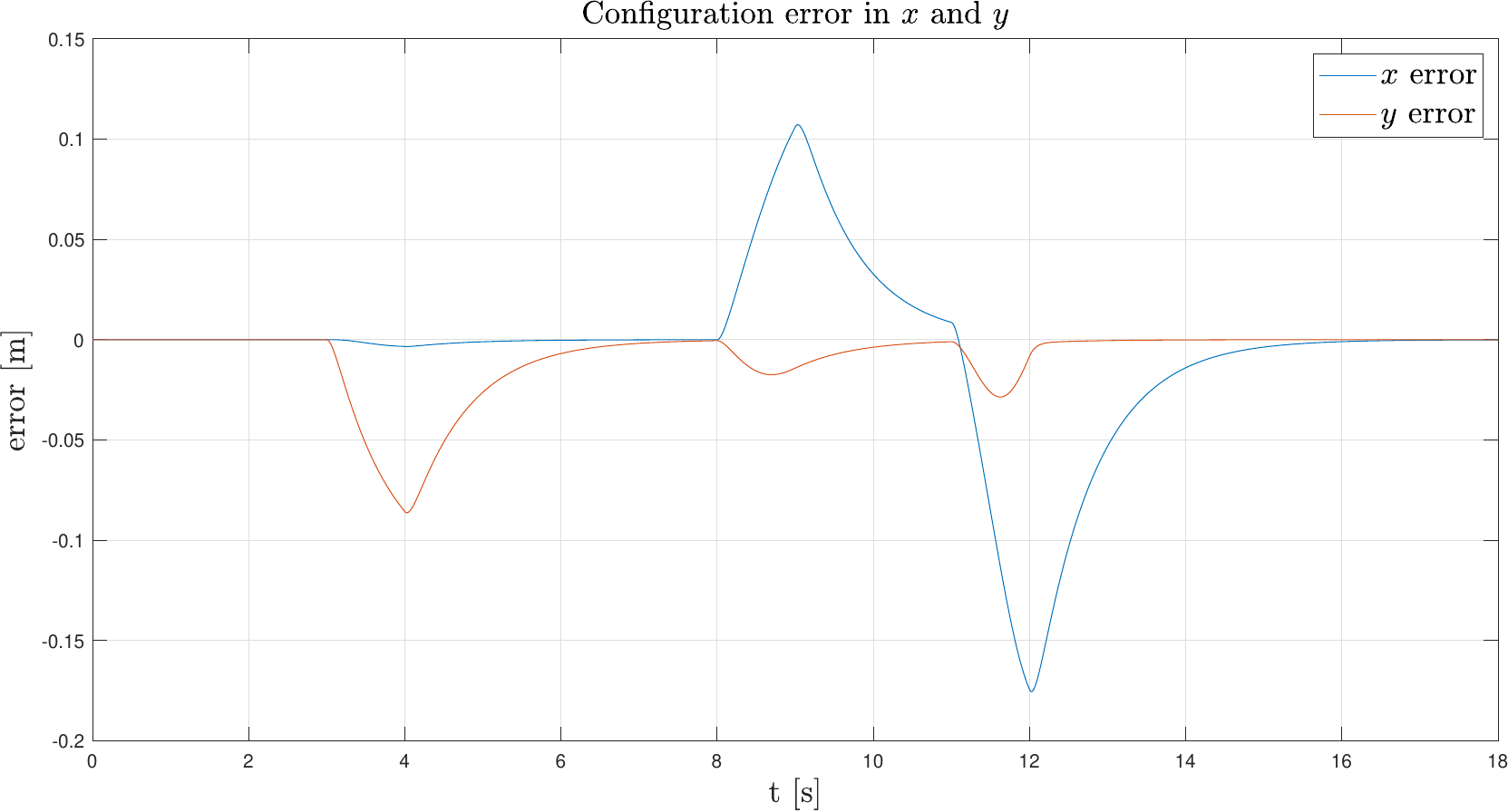}
		
		\vspace{5mm}
		
		\includegraphics[width=\textwidth]{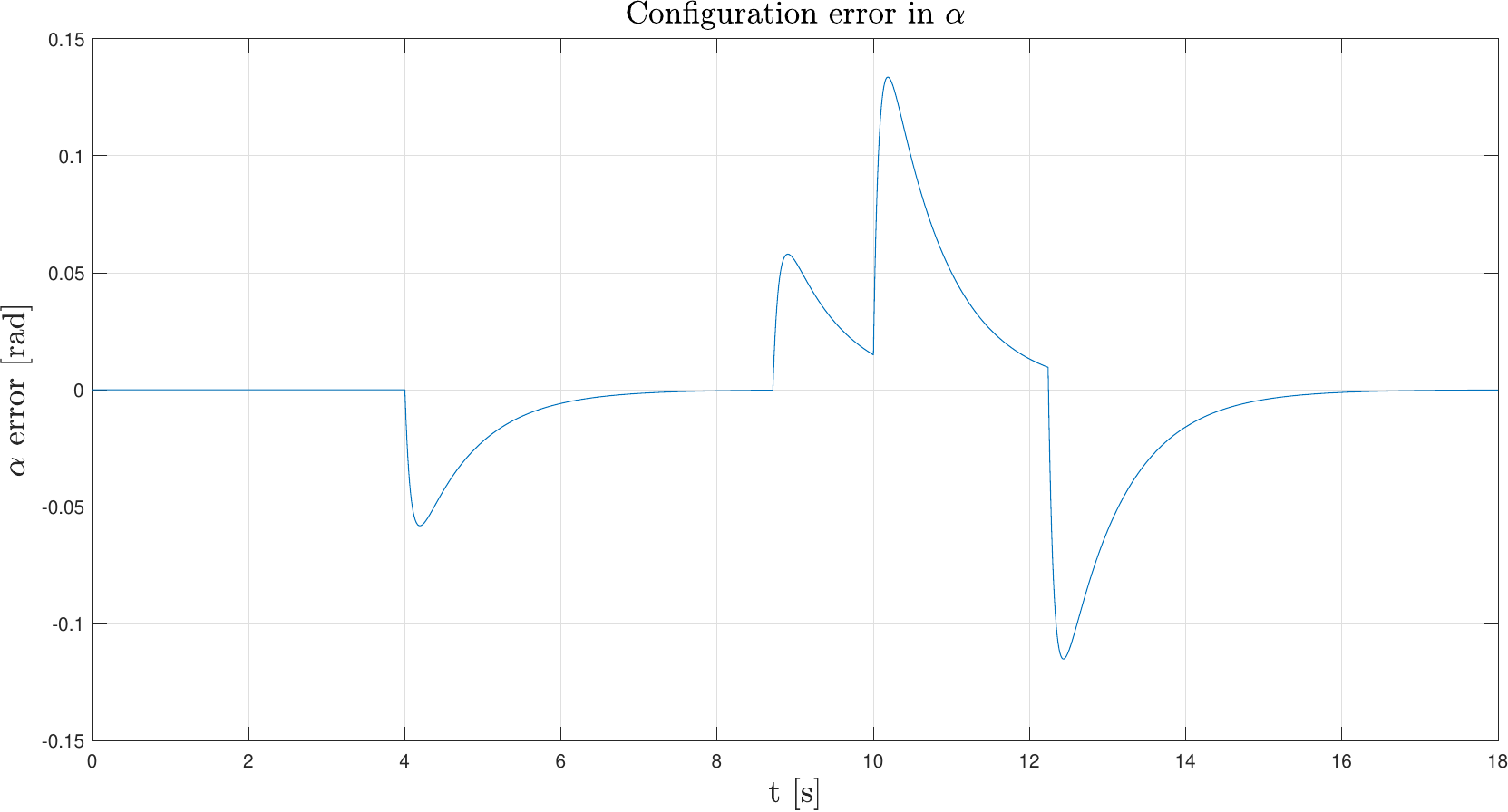}
    \end{center}
    \caption{Configuration errors in $x$, $y$, $\alpha$ in the disturbance rejection test.}
    \label{fig:DRCL10}
\end{figure}

\begin{figure}[t!]		
	\begin{center}
		\includegraphics[width=\textwidth]{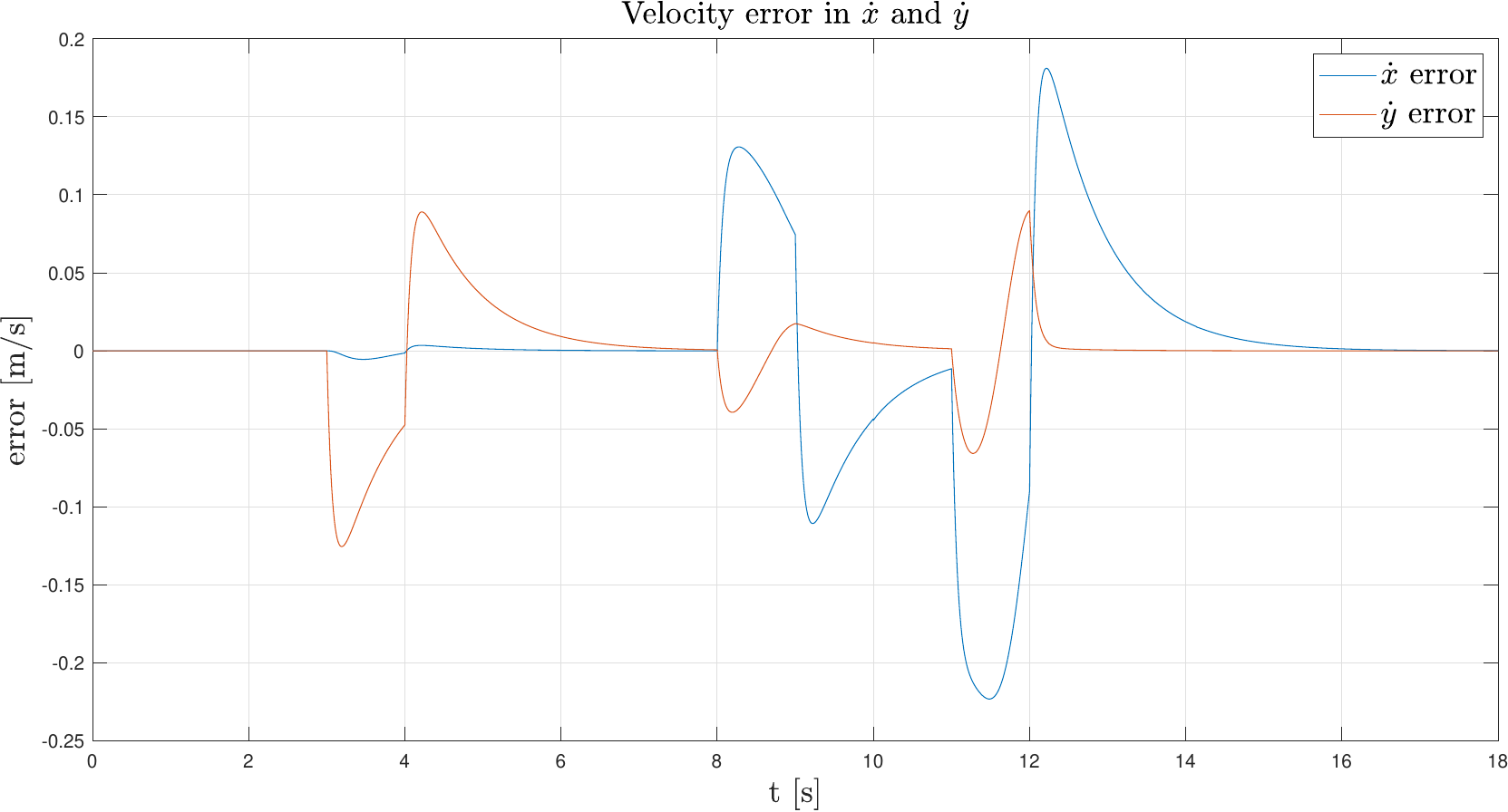}
		
		\vspace{5mm}
		
		\includegraphics[width=\textwidth]{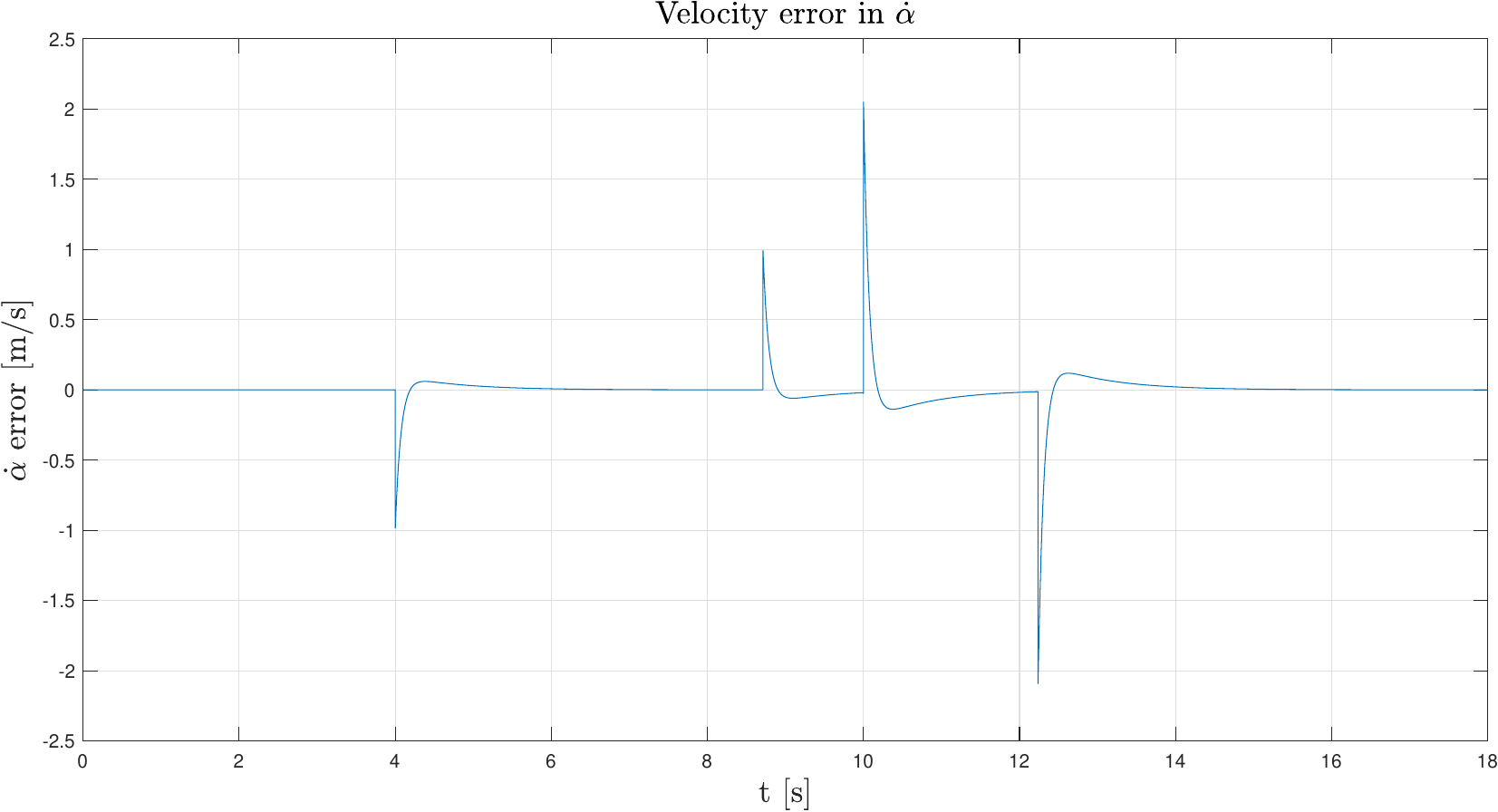}
	\end{center}
    \caption{Velocity errors in $\dot{x}$, $\dot{y}$, $\dot{\alpha}$ in the disturbance rejection test.}
    \label{fig:DRCL20}
\end{figure}

To illustrate the effect of these forces, the results of their application during a simulation of $18 \mathrm{s}$ will be shown below (Fig.~\ref{fig:InfShapeCLRute}). During this simulation, we will see Otbot following a ``figure 8'' trajectory composed of straight and circular sections. During this tracking three perturbation forces are applied on $F$:

\begin{itemize}
	\setlength{\itemsep}{0\baselineskip}
	\item A force of $150 \mathrm{N}$ in the negative $y$ direction at $t=4 \mathrm{s}$.
	\item A force of $200 \mathrm{N}$ in the positive $x$ direction at $t=8 \mathrm{s}$.
	\item A force of $350 \mathrm{N}$ in the negative $x$ direction at $t = 11 \mathrm{s}$.
\end{itemize}
The three forces are applied individually during $1 \mathrm{s}$ from the times indicated. From the plots in Figs.~\ref{fig:DRCL10} and~\ref{fig:DRCL20} and the video in \url{https://youtu.be/6bkYEmZqX_o} we can see how Otbot corrects all the deviations caused by the three perturbation forces in about $3\mathrm{s}$ as expected.

\section{Conclusions}
\label{ch:conclusions}


\subsection{Summary of contributions}
\label{sec:contributions}

In this work we have obtained a dynamic model for Otbot, we have proposed a method to identify its parameters, and we have designed a control law that allows the tracking of predefined trajectories in a robust and accurate way.

The dynamic model is very generic. It takes into account the mass and moment of inertia of all robot bodies, the viscous friction on the axes, and assumes arbitrary positions for the centers of mass of the chassis and the platform. The model has also been manipulated to obtain compact solutions of the direct and inverse dynamic problems, which are necessary to simulate the robot motions or to obtain the torques that generate a certain trajectory. In addition, we have obtained the dynamic model in task space coordinates, which is key to design the control law we propose.

The identification method is based on exciting the robot with constant torques for a few seconds, recording the signals of certain sensors onboard, and determining the values of the parameters that minimize the difference between the predicted signals and the recorded ones. This minimization must be started from a good estimate of the actual parameters, but the process has fairly large basins of attraction towards them. The identification experiment, moreover, uses robot sensors exclusively, and its learning motion can be performed in small spaces of only about $4\mathrm{m}^2$. 


The controller we obtain is based on the computed torque technique, which ensures global stability. This means that the robot will converge to the desired trajectory regardless of its initial conditions, as long as the applied torques do not exceed the limits of the motors themselves. The controller's gains can also be easily adjusted to guarantee that the tracking errors are eliminated within a predefined time. Finally, the controller has been validated in simulation by applying it to track (1) a narrow corridor where the robot's omnidirectional capability must be used to make sudden changes of direction, (2) a motion with small inconsistencies between position and velocity references, and (3) a trajectory during which unmodelled force disturbances appear. In all cases the controller eliminates the errors that arise in the desired stabilization time. 

\subsection{Future work}
\label{sec:futurework}

Otbot is a little known mobile platform with interesting properties. The fact that it achieves omnidirectionality with conventional wheels makes it advantageous compared to vehicles using Mecanum or omni wheels, which are more complex to manufacture and maintain, and support lower loads in general. Otbot also does not possess actuation singularities, which makes it preferable over other omnidirectional platforms with conventional wheels that present such critical configurations~\cite{batlle2009holonomy}. For all these reasons, we believe that it would be good to continue studying this robot in the future, to address the following aspects among others:
\begin{itemize}
	\setlength{\itemsep}{0\baselineskip}
	\item A detailed study should be made of the advantages that the Otbot may have compared to other non-omnidirectional platforms that are widely used in industry, such as Amazon's Kiva robots~\cite{d2008future}. These robots resemble Otbot, but their pivot joint is on the wheels' axis, so they do not have omnidirectional capability. While Otbot can move continuously through narrow corridors with sudden changes of direction, the Kiva robots may need to stop, reorient, and continue in order to make progress. It would be good to compare the minimum-time trajectories of the two robots and determine how faster is Otbot in comparison.
	\item An experimental prototype of Otbot should be built, equipped with the necessary sensors in order to experimentally validate the parameter identification and control methods we propose.
	\item To validate the identification process one should identify the parameters using the learning trajectory we propose, and then see how well these parameters predict other movements of the robot. This check could be done by recording new sensory signals with movements generated by known control actions, and seeing whether the identified model anticipates these movements from the mentioned actions.
	\item To validate the control law, we should see whether the robot is able to follow the desired trajectory despite small disturbances or wheel slippages that could affect it. This means that good measurements of the position and orientation of the platform would have to be available, which could be achieved by means global positioning systems, or inertial navigation systems complemented by positioning based on calibrated landmarks.
	\item The robot could be equipped with a path planner to provide it with autonomy to generate navigation plans for obstacle environments. The work in~\cite{gautier2021mscthesis} is an important step forward in this direction.
	\item Finally, the proposed controller could be made adaptive, so the system could autonomously adjust the dynamic parameters whenever they changed on the fly, such as when depositing a load of unknown mass and moment of inertia on an arbitrary point on the platform. 
\end{itemize}


\renewcommand{\refname}{\spacedlowsmallcaps{References}} 

\bibliographystyle{unsrt}

\bibliography{references} 


\end{document}